\documentclass[journal]{IEEEtran}
\usepackage[numbers,sort&compress]{natbib}
\usepackage{footnote}
\usepackage{url}

\usepackage{soul}
\soulregister\cite7
\soulregister\citep7
\soulregister\citet7
\soulregister\ref7
\soulregister\pageref7

\ifCLASSINFOpdf
   \usepackage[pdftex]{graphicx}
   \graphicspath{{image}}
   \DeclareGraphicsExtensions{.pdf,.jpeg,.png}
\else
   \usepackage[dvips]{graphicx}
   \graphicspath{{../eps/}}
   \DeclareGraphicsExtensions{.eps}
\fi
\usepackage{amsmath}
\usepackage{amsfonts}
\usepackage{color}
\usepackage{multirow}
\usepackage{booktabs} 

\usepackage{algpseudocode}
\usepackage{algorithm}
\ifCLASSOPTIONcompsoc
  \usepackage[caption=false,font=normalsize,labelfont=sf,textfont=sf]{subfig}
\else
  \usepackage[caption=false,font=footnotesize]{subfig}
\fi
\usepackage{fixltx2e}

\usepackage{stfloats}
\fnbelowfloat
\begin{document}
	
%
\title{Low-rank Meets Sparseness: An Integrated
	Spatial-Spectral Total Variation Approach to Hyperspectral
	Denoising}
%
%

\author{Haijin~Zeng,  Shaoguang Huang, Yongyong Chen,
	Hiep Luong,
	and Wilfried Philips
	\thanks{Haijin Zeng,  Hiep Luong and W. Philips are with Image Processing and Interpretation, imec research group
		at Ghent University, Belgium (e-mail: Haijin.Zeng@UGent.be; hiep.luong@UGent.be;
		Wilfried.Philips@ugent.be)}
	\thanks{Shaoguang Huang is with the Group for Artificial Intelligence and
		Sparse Modeling (GAIM), Department of Telecommunications and Information Processing (TELIN), Ghent University, 9000 Ghent, Belgium (e-mail:
		shaoguang.huang@ugent.be).}
	\thanks{Yongyong Chen is with the School of Computer Science and Technology,
		Harbin Institute of Technology (Shenzhen), Shenzhen 518055, China ( Email:YongyongChen.cn@gmail.com)}
	\thanks{Preprint, 24 November, 2020.}}

%
%

\markboth{Preprint, November, 2020}%
{Shell \MakeLowercase{\textit{et al.}}: Bare Demo of IEEEtran.cls for IEEE Journals}
%



\maketitle

\begin{abstract}
Spatial-Spectral Total Variation (SSTV) can quantify
local smoothness of image structures, so it is widely used in
hyperspectral image (HSI) processing tasks. 
Essentially, SSTV assumes a sparse structure of gradient maps calculated along the spatial and spectral directions. 
In fact, these gradient tensors are not only sparse, but also (approximately) low-rank under FFT, which we have verified by numerical tests and theoretical analysis. 
Based on this fact, we propose a novel TV regularization to simultaneously characterize the sparsity and low-rank priors of the gradient map (LRSTV). 
The new regularization not only imposes sparsity on the gradient map itself, but also penalize the rank on the gradient map after Fourier transform along the spectral dimension. 
It naturally encodes the sparsity and low-rank priors of the gradient map, and thus is expected to reflect the inherent structure of the original image more faithfully. Further, we use LRSTV to replace conventional SSTV and embed it in the HSI processing model to improve its performance. 
Experimental results on multiple public data-sets with heavy mixed noise show that the proposed model can get 1.5dB improvement of PSNR. 
\end{abstract}


%
\IEEEpeerreviewmaketitle

\section{Introduction}
%
%
%
%


Three-dimensional (3D) hyperspectral images (HSIs) have precise spectral accuracy. 
Compared with traditional imaging systems, HSI images with many spectral bands can reveal material properties, which helps many image analysis tasks.
Such as face recognition, mineral exploration, target detection and quality control \cite{xu_target_2016_5_3}. 
In real life applications, imaging conditions, i.e., weather, sensor sensitivity, photon effects, lighting conditions and cali-bration errors, cause many types of degradation in the  HSI images \cite{goetz2009three}. 
These degradations including noise seriously affect the subsequent processing of HSI. 
Therefore, an effective recovery algorithm is a necessary requirement for improving the accuracy of subsequent processing.

Hyperspectral image denoising, i.e., estimating a clean HSI from a noisy one, is essentially an inverse problem. 
Regularization is one of the most effective and widely used methods to solve this kind of inverse problem. 
One of the key components of this method is to explore and encode the prior structure and constrain the solution space accordingly,
in the form of regularization term.
Regularization technology based on spatial-spectral total variation (SSTV) is a widely used and most effective powerful image restoration method. 
SSTV explores the local smoothness (in most pixels) along the HSI spatial and spectral directions. 
The spatial local smoothness means that similar objects/shapes are usually distributed adjacently with similar spectral waves. 
Moreover, sensor data from  adjacent spectral bands, tends to have some correlation,
e.g., because the spectral filters in the sensors have overlapping bandpass characteristics similarity  \cite{E3DTV}. 
It should be noted however, that spectra also contain important narrow
peaks and valleys. These properties lead to piecewise smoothness, which translates in
a low total variation. 
The local smooth prior structure possessed by HSI can be equivalently characterized as the sparsity of the gradient map calculated along the spatial and spectral modes of the HSI, and then naturally embedded as the total variation of the different modes of the HSI.

Although it is widely used in various image processing tasks and has achieved success, 
SSTV based on the $L_1$ norm only describes the sparsity, and does not further explore some more structures behind the HSI gradient map. 
Specifically, in SSTV regularization, the $L_1 $ norm is utilized to measure the sparsity in the gradient domain, that is, the same and independently distributed Laplacian prior distribution is used to measure the sparsity of all bands of the gradient map in the three directions (spectrum, spatial width and height). 
This shows that SSTV implicitly assumes that the sparsity in all bands of these gradient maps is independent. 
However, this is always different from the real scene. 
The original HSI has obvious spectral band correlation, the gradient map also inherits the low-rankness in the image domain, therefore, the different bands of the gradient map are not independent \cite{E3DTV}.

To represent the low-rank priors of the gradient map in addition to well-known sparsity, we proposed a novel TV regularization term. 
The contributions of this paper are listed as follows:

\begin{itemize}
	\item[ 1)]
	 		We propose a novel total variation regularization term integrated sparsity and low-rankness of the gradient map (LRSTV). 
	 		The new TV more faithfully represents the sparseness and additional low-rank prior features of the gradient map along all HSI bands, and can replace the conventional SSTV to improve the performance of general HSI processing model.
	
	\item[ 2)] We have fully verified the fact that the gradient map of the image is not only sparse, but also (approximately) low-rank from both numerical testing and theoretical analysis.
	
	\item[ 3)] With LRSTV as the regularization term, we propose a LRSTV regularized tensor low-rank decomposition model for HSI denoising, and an ADMM based algorithm to solve the proposed model. All update equations in the algorithm are in closed form.
	
	\item[ 4)] Both simulation data and real data experiments are implemented. 
	The comprehensive experimental results confirmed the superiority of the proposed model over the existing technology, and also proved that the easy replacement of traditional SSTV with LRSTV can get promising performance improvement.

\end{itemize}

The rest of this paper is organized as follows:
Section \ref{related work} reviews studies relevant to low-rank based methods and total variation regularized low-rank models.
Section \ref{key model} proposes the LRSTV regularization term, a low-rank denoising model based on it,
and an ADMM based optimization algorithm is designed.
Simulated and real HSI data experiments verify the performance of the proposed model in Section \ref{results}.
Finally, Section \ref{conclusion} summarizes this paper.


\begin{figure*}[!t]
	\centering
	\includegraphics[width=0.9\linewidth]{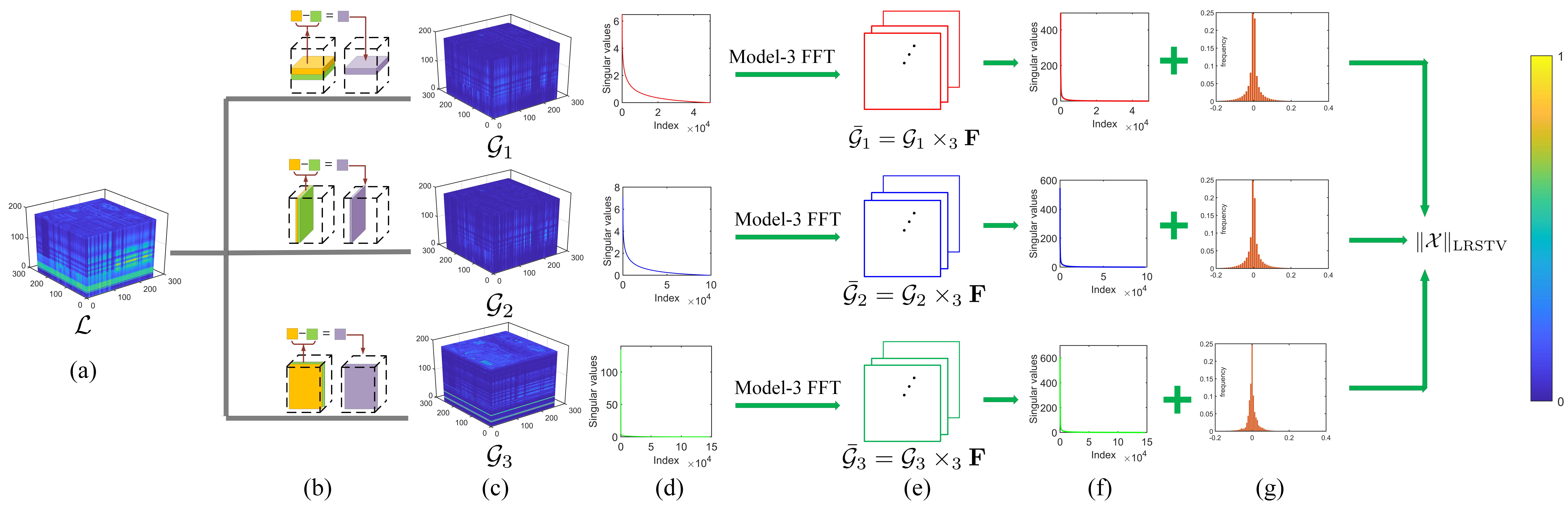}
	\caption{Illustration of the the proposed LRSTV regularization terms. 
		(a) An example of real HSI data \emph{HYDICE Washington DC Mall}, represented as a 	tensor $\mathcal{L}$;
		(b) Illustrations of the difference operators along the spatial height, width and
		spectral modes, respectively; 
		(c) The gradient maps of $\mathcal{L}$ in spatial height, width and spectrum, represented as $\mathcal{G}_n, n = 1, 2, 3$, respectively. Each of these tensors is stacked by 160 slices of gradient maps; 
		(d) The singular value curves of the gradient maps; 
		(e) FFT transformed gradient maps, $\mathbf{F}$ denotes a Fourier transform matrix; 
		(f) The singular value curves of FFT transformed gradient maps;
		(g) The frequency distributions of the gradient maps of \emph{HYDICE Washington DC Mall} data.}
	\label{fig:lrsstv}	
\end{figure*}

\section{Related Work} \label{related work}

A simple method for hyperspectral image denoising is to denoise each band independently. 
Under this denoising mechanism, traditional 1-D or 2-D denoising methods designed for natural image can be directly applied to HSI Denoising. 
Each pixel of HSI can be regarded as a 1-D signal, which can be denoised. 
Also each spectral band of HSI can also be regarded as a grayscale image, which can also be denoised. 
However, these noise reduction methods ignore the strong correlation in the spectral or spatial direction in HSI, so there is still a large room for improvement in model performance.

To alleviate the problems in the band-by-band denoising technology, lots of work has focused on extending the 1-D or 2-D denoising model to the 3-D model. 
These methods have achieved simultaneous exploration of the spatial and Spectral information, also have achieved better results compared to band-by-band denoising technology. 
Zhang et al. \cite {zhang_bayesian_2012_23_30} propose a recovery model on the basis of Bayesian theory, which averages HSI spectral bands to generate potential multi-spectral images. 
Xue et al. \cite{XueMLSTD} propose a new multilayer sparsity-based 3-D decomposition method, which effectively represents the structured sparsity of a tensor and results in promising results.

%

Non-local self-similarity is a widely used prior information in HSI processing \cite{xie_Nonlocal}. 
It means that for any image block of HSI in space, a similar texture or structure can always be found in the entire image. 
For the hyperspectral image of each band, the non-local similarity block reveals the low rank of the spatial dimension. 
Therefoe, by exploring this low rankness, the spatial noise can be significantly alleviated.
BM4D \cite{BM4D} is a classic method of encoding this non-local prior. 
Similar to the block technique in BM4D, Zhang et al. \cite {LRMR} propose the LRMR model for HSI denoising, which divides the HSI into overlapping small blocks, and then ranks them into a matrix band by band, and finally uses a low matrix to approximate the restoration of the potential image. 
To increase the approximate accuracy of low-rank matrices, Chen et al. \cite {nonLRMA} proposed a non-convex $ \gamma $ norm instead of the nuclear norm in LRMR, and proposed the NonLRMA model. 
He et al. \cite{NAILRMA} propose a noise-adjusted iterative denosiing model based on LRMA (NAILRMA). 
In addition, based on the plug-and-play framework, Zeng et al. \cite{zengAccess} proposed the NLRPnP model, which can simultaneously represent the local low-rank structure and non-local self-similarity of HSI.

%
%

Although the matrix-based technology has achieved success, many recent studies have proved that the tensor-based restoration method has certain advantages over the matrix-based method \cite{chang2020weighted}. 
The key point is that the tensor method directly models the 3-D HSI, while the matrix technology needs to unfold the 3-D HSI to 2-D data. This dimensionality reduction process will lose or destroy some inherent structural information. Under the tensor framework, the low-rank structure of HSI can be represented by tensor decomposition of tensor approximation \cite{XueSSLRR}. 
The most widely used methods are Tucker decomposition \& Tucker rank, CP decomposition \& CP rank, T-SVD decomposition \& tubal rank and average rank, tensor train, and tensor ring  \cite{zengCVIU, Xile_tensor_train, Xile_tensor_ring} etc. 
\cite{xue2019nonlocal} combines nonlocal low-rank with CP tensor decomposition, which can represent the global spectral correlation and spatial nonlocal self-similarity.
In addition, Zeng et al.\cite{zengLRDPR} introduce the deep network with Tucker-based low-rank prior into the plug-and-play (PnP) framework and propose a deep PnP regularized HSI restoration model, which achieves promising results.

TV regularization technology is another powerful image restoration method, which can effectively describe the local smooth structure of the image. A variety of HSI restoration models have been derived based on TV \cite{E3DTV, zengSP}, and they have been proven to effectively improve the performance of the model.
Mathematically, TV represents the sparsity of the gradient map of image by utilizing $L_1$ norm (anisotropic TV) or $L_2$ norm (isotropic TV).
To better represent the sparsity of the gradient map, \cite{zengTGRS} proposes a non-convex 3-D TV, which designs a $L_1-\alpha L_2$ metric to encode the sparsity.
In addition, due to the convexity and simplicity of TV, it is easy to integrate it and other prior regularization terms into a unified model to enhance the performance of HSI restoration algorithm. 
See, e.g., \cite{LRTDTV, LRTV,zengTGRS,liao2013two} among others.

\section{HSI restoration via a LRSTV regularized tensor decomposition} \label{key model}

\subsection{LRSTV Regularization} \label{LRSTV_section}

The regularization technology based on the total variation (TV) has been regarded as a powerful image restoration method. It can effectively preserve the spatial sparsity in addition to protecting the boundary information of the image. 
TV has two mathematical definitions, i.e.,

\begin{equation}
\label{eq:TV}
\begin{split}
\|\mathbf{X}\|_{\mathrm{TV}}^{\mathrm{iso}} &:=\sum_{i=1}^{mn} \sqrt{\left(\text{D}_{1}^{i} \mathbf{X}\right)^{2}+\left(\text{D}_{2}^{i} \mathbf{X}\right)^{2}}\\
\|\mathbf{X}\|_{\mathrm{TV}}^{\mathrm{ani}} &:=\sum_{i=1}^{mn}\left|\text{D}_{1}^{i} \mathbf{X}\right|+\left|\text{D}_{2}^{i} \mathbf{X}\right|
\end{split}
\end{equation}
where $\mathbf{X}\in\mathbb{R}^{m \times n}$;
$\text{D}^{i}_{1}, \text{D}^{i}_{2}$ denote the gradient operator along horizontal and vertical direction at pixel $i$. 
Subsequently, 2-D TV was expanded to 3-D and widely used in HSI processing. 
The widely used and worked sparsity measure designed for gradient map is the SSTV, i.e., using $L_1$-norm metric to measure sparsity \cite{LRTDTV},
\begin{equation}
\label{eq:SSTV}
\|\mathcal{L}\|_{\mathrm{SSTV}}^{\mathrm{ani}}=\tau_{1}\left\|\mathcal{D}_{1} \mathcal{L}\right\|_{1}+\tau_{2}\left\|\mathcal{D}_{2} \mathcal{L}\right\|_{1}+\tau_{3}\left\|\mathcal{D}_{3} \mathcal{L}\right\|_{1},
\end{equation}
where $\mathcal{L} \in \mathbb{R}^{m \times n \times p}$ denotes an HSI, $\mathcal{D}_{3}$ is the additional gradient operator along the spectral direction and $\tau_{1}, \tau_{2}, \tau_{3}$ are nonnegative regularization parameters.

Mathematically, SSTV regularization, i.e., 3-D anisotropic TV, measures the sparsity of gradient map by utilizing $L_1$ norm as the convex approximation of $L_0$ norm. This metric has achieved great success in many tasks of image processing. However, we noticed one thing that in SSTV only sparsity is explored. 
Here, we can't help but want to ask a question: is the gradient map only sparse, is there no other prior information that can be explored?
\begin{figure}[H]
	\centering
	\subfloat[A HSI]{\includegraphics[width=0.3\linewidth]{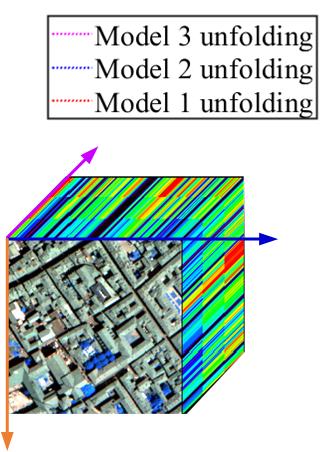}}%
	\hfil
	\subfloat[Singular Values]{\includegraphics[width=0.4\linewidth]{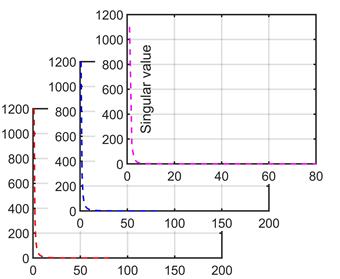}}%
	\caption{(a) A real HSI cube of Pavia Centre, (b) the singular values of its model-$n(n=1,2,3)$ unfolding matrix. From the figure, one can see that the hird-order real HSI is obviously correlated along its three modes.}
	\label{fig:HSI_Low-rank}
\end{figure}
According to the linear mixture model of hyperspectral image, one can represent each spectral feature by using a linear combination of a small number of pure spectral end members. 
Specially, let $\mathbf{L}_{(3)}$ denotes the mode-$3$ matrix of HSI $\mathcal{L}$, $\mathbf{H} \in \mathbb{R}^{p \times r}$ represents the endmember matrix, and $\mathbf{P} \in \mathbb{R}^{mn \times r}$ denotes the abundance matrix. 
Then we have $\mathbf{L}_{(3)} = \mathbf{P}\mathbf{H}^{T}$, and $r \ll p $ or $r \ll mn$ \citep{LRTDTV, zengCVIU}, which shows that the number of end members $r$ is relatively small than $mn$ and $p$.
This means that only a few of the singular values of $\mathbf{L}_{(3)}$ are non-zero.
Taking Pavia as an example, as shown in Fig. \ref{fig:HSI_Low-rank}, the singular value curve of $\mathbf{L}_{(3)}$ decays rapidly.
Furthermore, in Fig. \ref{fig:HSI_Low-rank}, one can also see that both the singular value curves of $\mathbf{L}_{(1)}$ and $\mathbf{L}_{(2)}$ also decay rapidly.
In summary, all the three modes of HSI have similar low-rankness. 
Next, we will show this low-rank structure that appears in the image domain will also be inherited by the gradient map of the original image, and verify it from both numerical testing and theoretical analysis.


\emph{Numerical testing:} Fig. \ref{fig:lrsstv}-(a) displays a real 3-D HSI cube, expressed as $\mathcal{X} \in \mathbb{R}^{200 \times 200 \times 160}$; (b) shows the difference operator along the spatial height, width and spectral direction; (c) represents the gradient maps of original HSI in spatial vertical, horizontal and spectral directions are expressed as $\mathcal{G}_n \in \mathbb{R}^{200 \times 200 \times 160}, n = 1, 2, 3$. Each tensor is stacked by a gradient map of 160 slices. Except for the sparsity shown in Fig. \ref{fig:lrsstv}-(f). It can be seen from Fig. \ref{fig:lrsstv}-(e) that although the sparsity in different bands of the gradient map is not the same or even independent, there is a clear correlation between the different bands of the HSI gradient map. 
Further, Fig. \ref{fig:lrsstv}-(d) shows the gradient map in the Fourier domain. 
Compared with the original gradient map, one can see that the low-rankness of the gradient map in the Fourier domain is more obvious, especially the gradient map in two directions of spatial dimension. 
Due to the fact that most of the singular values are small and close to 0, only a few singular values have large values.

\emph{Theoretical analysis:}
In fact, gradient transformation hardly changes the rank of the original image, i.e., the gradient map inherits the low-rankness of original image in image domain.  
Next, we prove this conclusion theoretically.

\noindent \textbf{Lemma 4.1. } For any $\textbf{A} \in \mathbb{R}^{m \times k}$ and $\textbf{B} \in \mathbb{R}^{k \times n}.$ $r(\textbf{A})$ denotes the rank of $\textbf{A}$ and $r(\textbf{B})$ is the rank of $\textbf{B}$. Then we have
\begin{equation}
r(\textbf{A})+r(\textbf{B})-k \leq r(\textbf{A} \textbf{B}) \leq \min \{r(\textbf{A}), r(\textbf{B})\}	
\end{equation}

\emph{Proof.}
Firstly, we prove that $r(\textbf{A})+r(\textbf{B})-k  \leq r(\textbf{AB})$. 
Let $r(\textbf{A})=r_1$, $r(\textbf{B})=r_2$, $r(\textbf{AB})=r$, then there are invertible matrices \textbf{A} and \textbf{B} such that the following equation holds,

$$
\textbf{PAQ}=\left[\begin{array}{ll}
\textbf{E}_{r_1} & \textbf{0} \\
\textbf{0} & \textbf{0}
\end{array}\right]
$$
Let $
\textbf{Q}^{-1} \textbf{B}=\left[\begin{array}{c}
\textbf{B}_{r_{1} \times m} \\
\textbf{B}_{\left(n-r_{1}\right) \times m}
\end{array}\right]
$, we have 
$$
r=r(\textbf{A} \textbf{B})=r\left(\textbf{PAQQ}^{-1}\textbf{B}\right),
$$
due to
$$
\textbf{P} \textbf{A} \textbf{Q} \textbf{Q}^{-1} \textbf{B}=\left(\begin{array}{cc}
\textbf{E}_{r_{1}} & \textbf{O} \\
\textbf{O} & \textbf{O}
\end{array}\right)\left(\begin{array}{c}
\textbf{B}_{r_{1} \times m} \\
\textbf{B}_{\left(n-r_{1}\right) \times m}
\end{array}\right)=\left(\begin{array}{c}
\textbf{B}_{r_{1} \times m} \\
\textbf{0}
\end{array}\right),
$$
we have $
r\left(\textbf{B}_{r_{1} \times m}\right)=r(\textbf{A} \textbf{B})=r,
$ however, $r(\textbf{Q}^{-1} \textbf{B})=r_2$, this means that the number of linearly independent rows in $\textbf{B}_{\left(n-r_{1}\right) \times m}$ is $r_2-r$, and the total number of rows is $n-r_1$. Therefore, $r_2-r \leq n-r_1$, that is, $r \geq r_1 + r_2 -n$.
Secondly, to prove that $r(\textbf{A} \textbf{B}) \leq \min \{r(\textbf{A}), r(\textbf{B})\}$, 
we only need to prove $r(\textbf{A} \textbf{B}) \leq r(\textbf{A})$ and $r(\textbf{A} \textbf{B}) \leq r(\textbf{B})$ at the same time. 
The proofs of these two inequalities please refer to the Appendix.

\noindent \textbf{Theorem 4.1.} Given any 3-D hyperspectral image $\mathcal{L} \in \mathbb{R}^{m \times n \times p}$ and its gradient tensors $\mathcal{D}_{x} \mathcal{L}, \mathcal{D}_{y} \mathcal{L}, \mathcal{D}_{z} \mathcal{L} \in \mathbb{R}^{m \times n \times p}$.
we have 
$$r_{tc}\left(\mathcal{D}_{x} \mathcal{L} \right) \approx r_{tc}\left(\mathcal{D}_{y} \mathcal{L} \right) \approx r_{tc}(\mathcal{L}) \approx r_{tc}\left(\mathcal{D}_{z} \mathcal{L}\right)$$
where the Tucker rank of $\mathcal{A}$ is defined as
$r_{tc}(\mathcal{A})=\left(r(\mathbf{A}_{(1)}), r(\mathbf{A}_{(2)}), r(\mathbf{A}_{(3)})\right)$, and $r(\mathbf{A}_{(i)})$ is the rank of mode-$i$ matricization of $\mathcal{A}$.

\emph{Proof.}
Due to the three modes in multi-tubal rank has equivalent status, we first prove the theorem in one model, then it can easily be deduced to the remaining two modes. Then, we have
\begin{equation}
\begin{aligned}
r_{tc}(\mathcal{L})=\left(r(\mathbf{L}_{(1)}), r(\mathbf{L}_{(2)}), r( \mathbf{L}_{(3)})\right)\\
r_{tc}(\mathcal{D}_{x} \mathcal{L})=\left(r(\mathcal{D}_{x} \mathbf{L}_{(1)}), r_{2}(\mathcal{D}_{x} \mathbf{L}_{(2)}), r_{3}(\mathcal{D}_{x} \mathbf{L}_{(3)})\right)
\end{aligned}
\end{equation} 
To prove $r_{tc}(\mathcal{D}_{x} \mathcal{L}) \approx r_{tc}(\mathcal{L})$, one only need to prove that $r(\mathcal{D}_{x} \mathbf{L}_{(i)}) \approx r(\mathbf{L}_{(i)}), i=1,2,3$. 
Let $\mathcal{D}_{x} \mathbf{L}_{(1)}=\textbf{D}_{x} \mathbf{L}_{(1)},$ where $\textbf{D}_{x} \in \mathbb{R}^{np \times np}$ is the circulant matrix corresponding to the forward finite difference operator $\mathcal{D}_{x}$ with periodic boundary conditions along the $\textbf{x}$-axis. Then we know that $r\left(\textbf{D}_{x}\right)=np-1.$ 
According to Lemma 1, we get that
$$
r(\mathbf{L}_{(1)})-1 \leq r\left(\mathcal{D}_{x} \textbf{L}_{(1)}\right) \leq \min \{np-1, r(\textbf{L}_{(1)})\} \leq r(\textbf{L}_{(1)})
$$
i.e., $r\left(\mathcal{D}_{x} \textbf{L}_{(1)}\right) \approx r(\textbf{L}_{(1)}) .$ 
Similarly, we get 
\begin{equation}
\left\{\begin{array}{l}
r(\textbf{L}_{(2)})-1 \leq r\left(\mathcal{D}_{y} \textbf{L}_{(2)}\right) \leq r(\textbf{L}_{(2)})\\
r(\textbf{L}_{(3)})-1 \leq r\left(\mathcal{D}_{y} \textbf{L}_{(3)}\right) \leq r(\textbf{L}_{(3)})
\end{array}\right.
\end{equation}
and $r\left(\mathcal{D}_{y} \textbf{L}_{(2)}\right) \approx r(\textbf{L}_{(2)}), r\left(\mathcal{D}_{z} \textbf{L}_{(3)}\right) \approx r(\textbf{L}_{(3)}).$ Therefore,
we get that
\begin{equation}
\left\{\begin{array}{l}
\label{equa:rank_equal}
r(\mathcal{D}_{x} \mathbf{L}_{(1)}) \approx r(\mathbf{L}_{(1)}),\\
r(\mathcal{D}_{y} \mathbf{L}_{(2)}) \approx r(\mathbf{L}_{(2)}),\\
r(\mathcal{D}_{z} \mathbf{L}_{(3)}) \approx r(\mathbf{L}_{(3)}),\\
\end{array}\right.
\end{equation}
And for the forward finite difference operators $\mathcal{D}_{x}, \mathcal{D}_{y}, \mathcal{D}_{z}$ with zero boundary conditions $\left(r\left(\textbf{D}_{x}\right)=np-1, r\left(\textbf{D}_{y}\right)=mp-1, , r\left(\textbf{D}_{z}\right)=mn-1\right)$ and symmetric boundary conditions 
$$
r\left(\textbf{D}_{x}\right)=np, r\left(\textbf{D}_{y}\right)=mp, , r\left(\textbf{D}_{z}\right)=mn,
$$ 
we also have the conclusion (\ref{equa:rank_equal}). Therefore, the ranks of $\mathcal{D}_{x} \textbf{L}_{(1)}$, $\mathcal{D}_{y} \textbf{L}_{(2)}$ and $\mathcal{D}_{z} \textbf{L}_{(3)}$ represent the ranks of $\mathbf{L}_{(1)}, \mathbf{L}_{(2)}, \mathbf{L}_{(3)}$, respectively.

From the above numerical tests and theoretical analysis, we get the fact that the gradient map of HSI is not only sparse, but also has a low-rank structure, as long as the original HSI is low-rank. 
Moreover, this low rankness will be greatly enhanced in the Fourier domain.
Based on these facts, we propose a novel TV regularization to simultaneously characterize the sparsity and low-rank priors of the gradient map. 
Specifically, the proposed isotropic LRSTV and anisotropic LRSTV are defined as follows:

\begin{equation}\label{eq:CRSSTV_Ori}
\|\mathbf{\mathcal{L}}\|_{\mathrm{LRSTV}}^{\text{ani}}=\sum_{n=1}^{3}\left(\tau_{n}\left\|D_{n} \mathcal{L}\right\|_{1}+\alpha_{n}r_{tc}(D_{n} \mathcal{L})\right),
\end{equation}

\begin{equation}\label{eq:CRSSTV_iso_Ori}
\|\mathbf{\mathcal{L}}\|_{\mathrm{LRSTV}}^{\text{iso}}=\sqrt{\sum_{n=1}^{3}\tau_{n}\left\|D_{n} \mathcal{L}\right\|_{2}^2}+\sum_{n=1}^{3}\alpha_{n}r_{tc}(D_{n} \mathcal{L}),
\end{equation}
where $\tau_{n}$ and $\alpha_{n}$ are no-negative regularization parameters. 
Then, the key issue is to find  a simple method to estimate Tucker rank.
Here, we utilize the new average rank to represent the low-rankness that exists in the gradient map, 
i.e., replacing $r_{a}(D_{n} \mathcal{L})$ in (\ref{eq:CRSSTV_Ori}) and (\ref{eq:CRSSTV_iso_Ori}) with $r_{a}(D_{n} \mathcal{L})$,
%
where $r_{a}(\mathcal{A})=\frac{1}{n_{3}} \operatorname{rank}(\text {bcirc}(\mathcal{A}))$, and $\operatorname{bcirc}(\mathcal{A}) \in \mathbb{R}^{n_{1}n_{3} \times n_{2}n_{3}}$ is the block circulant matrix of $\mathcal{A}$ defined as follows:
$$
\operatorname{bcirc}(\mathcal{A})=
\left[
\begin{array}{cccc}
{\mathbf{A}^{(1)}} & {\mathbf{A}^{\left(n_{3}\right)}} & {\cdots} & {\mathbf{A}^{(2)}} \\
{\mathbf{A}^{(2)}} & {\mathbf{A}^{(1)}} & {\cdots} & {\mathbf{A}^{(3)}} \\
{\vdots} & {\vdots} & {\ddots} & {\vdots} \\
{\mathbf{A}^{\left(n_{3}\right)}} & {\mathbf{A}^{\left(n_{3}-1\right)}} & {\cdots} & {\mathbf{A}^{(1)}}
\end{array}
\right].
$$
where $\mathbf{A}^{(i)}:=\mathcal{A}(:,:,i)$ is the $i$-th front slice of $\mathcal{A}$.

Acually,
there are some connections between tensor Tucker rank and average rank, and these properties imply that the low Tucker rank or low average rank assumptions are reasonable for their applications in real visual data \cite{lu2019tensor}. 
Let
$r_{\mathrm{tc}}(\mathcal{A})=\left(r\left(\mathbf{A}_{(1)}\right), r\left(\mathbf{A}_{(2)}\right), r\left(\mathbf{A}_{(3)}\right)\right),$ 
where
$\mathbf{A}_{(i)}$ denotes the mode-$i$ matricization of $\mathcal{A},$ be the Tucker rank of $\mathcal{A} .$ Then we have  
$
r_{\mathrm{a}}(\mathcal{A}) \leq r\left(\mathbf{A}_{(1)}\right),
$
i.e., if a tensor with low Tucker rank, it also has low average rank, furthermore, the low average rank hypothesis is weaker than the low rank hypothesis \cite{lu2019tensor}. 
In this way, one can use tensor average rank to represent the low Tucker rank in (\ref{eq:CRSSTV_Ori}) and (\ref{eq:CRSSTV_iso_Ori}). 
Furthermore, with tensor average rank we have following theorem:

\noindent \textbf{Theorem 4.2.} \cite{TRPCA_averageRank}. On the set $\left\{\mathcal{A} \in \mathbb{R}^{n_{1} \times n_{2} \times n_{3}} \mid\|\mathcal{A}\| \leq 1\right\}$, $\mathcal{A}=\mathcal{U} * \mathcal{S} * \mathcal{V}^{*}$ represents the $t-SVD$ of $\mathcal{A} \in \mathbb{R}^{n_{1} \times n_{2} \times n_{3}} .$ 
The tensor tubal rank denoted as $r_{t}(\mathcal{A}),$ is
defined as the number of nonzero singular tubes of $\mathcal{S},$ i.e.,
$
r_{t}(\mathcal{A})=\#\{i, \mathcal{S}(i, i,:) \neq \mathbf{0}\}.
$
The tensor nuclear norm of $\mathcal{A}$ is defined as
$
\|\mathcal{A}\|_{*}:=\langle\mathcal{S}, \mathcal{I}\rangle=\sum_{i=1}^{r} \mathcal{S}(i, i, 1)
$
where $r=r_{t}(\mathcal{A})$.
Then the convex envelope of the tensor average rank $r_{a}(\mathcal{A})$ is the tensor nuclear norm $\|\mathcal{A}\|_{*}$.

According to the Theorem 4.2, (\ref{eq:CRSSTV_Ori}) and (\ref{eq:CRSSTV_iso_Ori}) can be approximated as follows:

\begin{equation}
\label{eq:CRSSTV}
\|\mathbf{\mathcal{L}}\|_{\mathrm{LRSTV}}^{\text{ani}}=\sum_{n=1}^{3}\left(\tau_{n}\left\|D_{n} \mathcal{L}\right\|_{1}+\alpha_{n}\left\|D_{n} \mathcal{L}\right\|_{*}\right),
\end{equation}
\begin{equation}
\label{eq:CRSSTV_iso}
\|\mathbf{\mathcal{L}}\|_{\mathrm{LRSTV}}^{\text{iso}}=\sqrt{\sum_{n=1}^{3}\tau_{n}\left\|D_{n} \mathcal{L}\right\|_{2}^2}+\sum_{n=1}^{3}\alpha_{n}\left\|D_{n}\mathcal{L}\right\|_{*},
\end{equation}
where $\left\|.\right\|_{*}$ represents the new tensor nuclear norm, $D_{n}, n=1,2,3$ denote the gradient operator.
The proposed LRSTV can be regarded as a generalization of classic SSTV, that is, SSTV is a special case of the proposed LRSTV. When the coefficient $\lambda$ is 0, LRSTV degenerates to SSTV.
It is worth pointing out that the new regularization not only imposes sparsity on the gradient map itself, but also calculates low rank on the gradient map after Fourier transform along the spectral dimension (FFT is contained in $\left\|.\right\|_{*}$). 
It naturally encodes the sparsity and low-rank priors of the gradient map, and thus is expected to reflect the inherent structure of the original image more faithfully than common SSTV.

\begin{figure}[H] \centering
	\subfloat[Original] {\includegraphics[width=0.23\columnwidth]{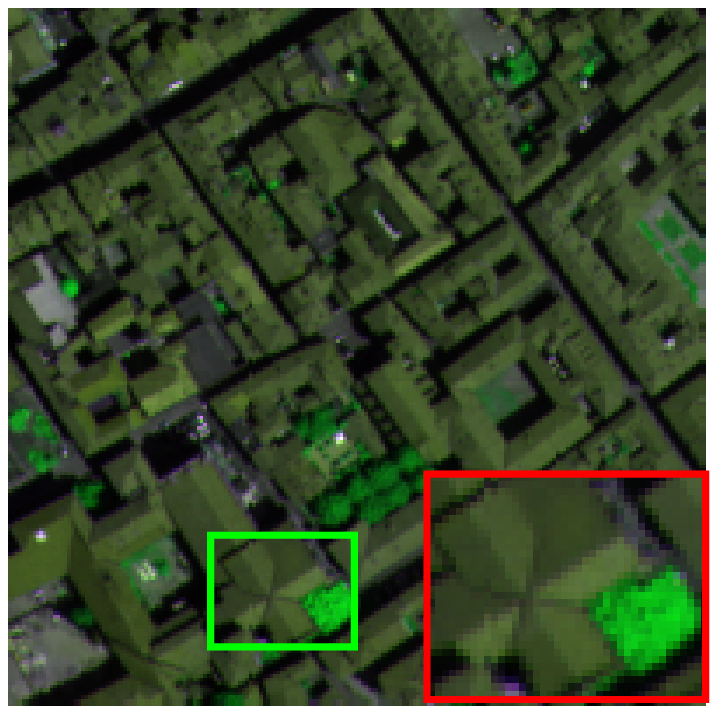}
	}\hfil
	\subfloat[Noisy image ] {\includegraphics[width=0.23\columnwidth]{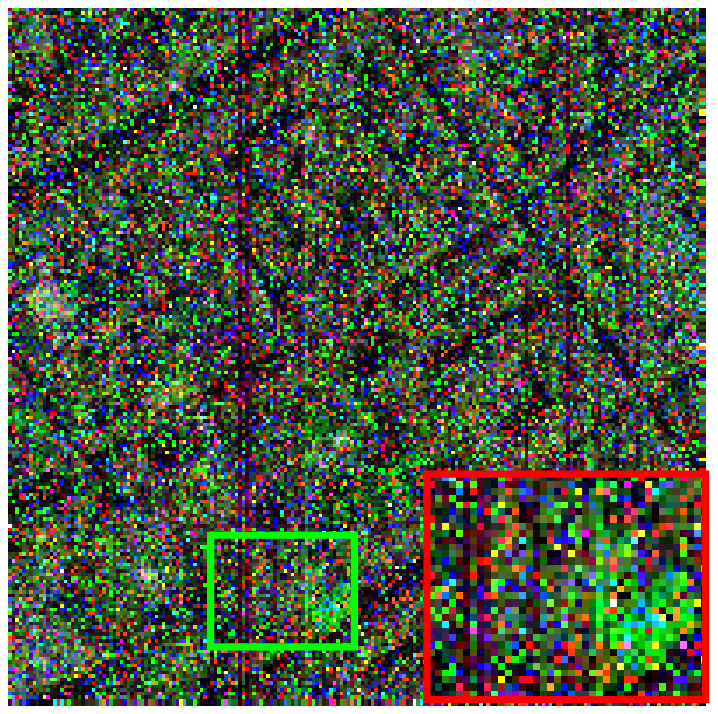}
	}\hfil
	\subfloat[SSTV] {\includegraphics[width=0.23\columnwidth]{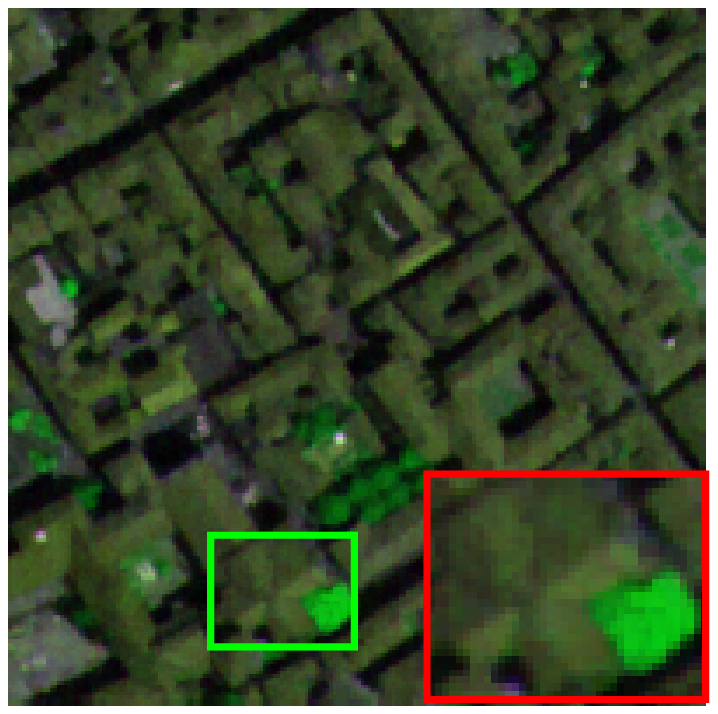}
	}\hfil
	\subfloat[Our TV] {\includegraphics[width=0.23\columnwidth]{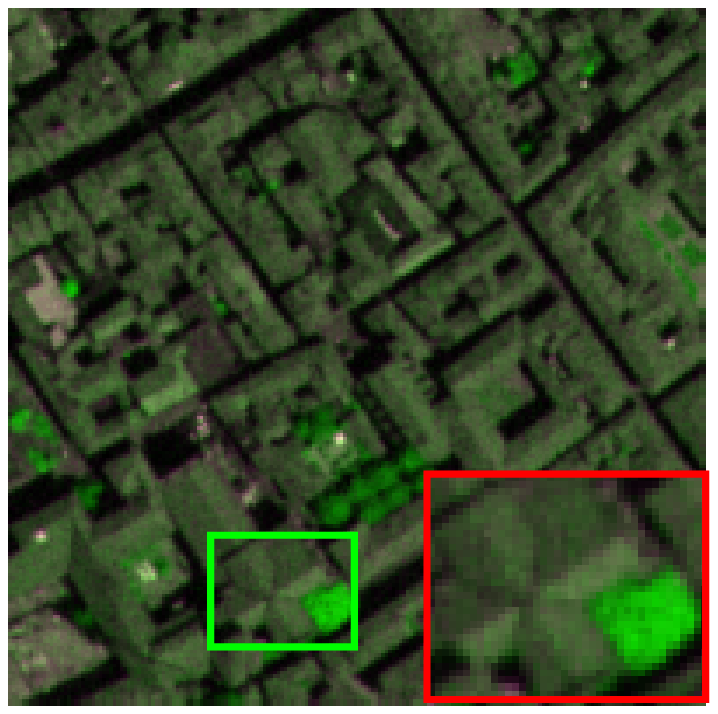}
	}\hfil
	\caption{Effectiveness of the additional low-rank strategy. (a)
		Original false-color image (R:
		31, G: 61, B: 9). (b) Simulated noisy image
		under  zero-mean Gaussian noise, impulse noise, deadline and stripe (the variance value of Gaussian noise and percentages
		of impulse noise being
		randomly selected from 0.1 to 0.2, the width of the deadlines
		was randomly generated from 1 to 3, the number of stripes being randomly selected from 20 to 40, (PSNR = 11.17dB). (c)
		Denoising result of the SSTV method without additional low-rank strategy (PSNR = 31.07dB). (d) Denoising result of the proposed method with additional low-rank strategy (PSNR = 32.63dB).}
	\label{fig:SSTVvsLRSTV_Pavia_imshow}
\end{figure}

As demonstrated in the previous paragraph, since SSTV only explores the sparsity of gradient map, it cannot effectively restore a potentially clear image when the observed image has serious noise pollution. 
LRSTV additionally explores the low-rank prior of the gradient map, which helps to deal with scenarios with high noise intensity. 
Due to image denoising is essentially an inverse problem, 
the more precise and appropriate the constraints added in the solution space, 
the closer the approximate solution to the true solution is.
In Fig. \ref{fig:SSTVvsLRSTV_Pavia_imshow}, we give an intuitive comparison between TV models with and without gradient low-rank strategy. Obviously, the local details of Fig. \ref{fig:SSTVvsLRSTV_Pavia_imshow}(d) is more clear than that of SSTV without gradient low-rank strategy, which shows the effectiveness of low-rank strategy in exploring the prior knowledge of gradient map.

%
%

\subsection{Low-rank Tensor Decomposition with Anisotropic LRSTV}

In addition to the above-mentioned sparse and low-rank priors of the gradient map, the third-order HSI also shows inherent structural information in its spatial and spectral mode in image domain \cite{LRTDTV}.
Here, Tucker decomposition is employed to represent the spatial-spectral low-rankness of HSIs displayed in Section \ref{LRSTV_section}. 
By using the Tucker-3 decomposition, any 3rd-order tensor $\mathcal{L} \in \mathbb{R}^{m \times n \times p} $ with multilinear rank $(r_1,r_2,r_3)$ can be decomposed as follows:
\begin{equation}
\mathcal{L}=\mathcal{C} \times_{1} \mathbf{U}_{1} \times_{2} \mathbf{U}_{2} \times_{3} \mathbf{U}_{3}, \mathbf{U}_{i}^{T} \mathbf{U}_{i}=\mathbf{I}, i=1,2,3,
\label{LRTAmodel}
\end{equation}
where $\mathcal{C}$ denotes the factor tensor, i.e., the so-called core tensor; $\mathbf{U}_{i}$ is the $i$-th factor matrix which has rank $r_i$, $i=1,2,3$.
Finally, by combining the correlation between the spatial mode and the spectral mode, the low rank and sparse structure of the gradient map, we propose LRSTV regularized tensor decomposition (named TDLRSTV) model, i.e.,

\begin{equation}
\label{equa:TDLRSTV}
\begin{aligned}
&\min _{\mathcal{L}, \mathcal{S}} \|\mathcal{L}\|_{\mathrm{LRSTV}}^{\text{ani}}+\lambda\|\mathcal{S}\|_{1}\\
&\quad s.t. \quad \mathcal{Y}=\mathcal{L}+\mathcal{S},\\
&\quad \mathcal{L}=\mathcal{C} \times_{1} \mathbf{U}_{1} \times_{2} \mathbf{U}_{2} \times_{3} \mathbf{U}_{3}, \mathbf{U}_{i}^{T} \mathbf{U}_{i}=\mathbf{I} ,i=1,2,3.
\end{aligned}
\end{equation}

\subsection{Optimization Procedure}

In this subsection, we design an algorithm based on ALM and ADMM \cite{LRTDTV} to solve the proposed TDLRSTV. 
Firstly, we introduce three auxiliary variables to split the multiple regularization terms applied on $\mathcal{L}$. 
In addition, for simplify, we set the same $\tau_{n}$ and $\alpha_{n}$ for all $n$ and denote them as $\tau, \alpha$, respectively, and let $D_{w}(\cdot)=\left[w_{1} \times D_{1}(\cdot); w_{2} \times D_{2}(\cdot); w_{3} \times D_{3}(\cdot)\right]$, where $w_{1}, w_{2}, w_{3}$ are the three weighted parameters added to three $L_1$ norm of gradient maps. Then, by introducing auxiliary variables $\mathcal{Z}$, $\mathcal{F}$ and $\mathcal{E}$, (\ref{equa:TDLRSTV}) can be rewritten as follows:
\begin{equation}
\label{equa:objectiveFunction}
\begin{aligned}
	\min _{\mathcal{C}, \mathbf{U}_{i}, \mathcal{L}, \atop \mathcal{F}, \mathcal{S}, \mathcal{E}, \mathcal{Z}} \tau\|\mathcal{F}\|_{1}+ \alpha \|\mathcal{E}\|_{*} +\lambda\|\mathcal{S}\|_{1} \\
	\text { s.t. } \mathcal{O}=\mathcal{L}+\mathcal{S}, \mathcal{L}=\mathcal{Z}, D_{w}(\mathcal{Z})=\mathcal{F}, \mathcal{E}=\mathcal{F}, \\
	\mathcal{L}=\mathcal{C} \times_{1} \mathbf{U}_{1} \times_{2} \mathbf{U}_{2} \times_{3} \mathbf{U}_{3}, \mathbf{U}_{i}^{T} \mathbf{U}_{i}=\mathbf{I}.
\end{aligned}
\end{equation}
According to ALM method, we rewrite the objective function of (\ref{equa:objectiveFunction}) into following augmented Lagrangian function:
\begin{equation}
\label{equa:Lagrangian}
\begin{array}{l}
L\left(\mathcal{L}, \mathcal{S}, \mathcal{Z}, \mathcal{F},\mathcal{E}\right)=\tau\|\mathcal{F}\|_{1}+ \alpha \|\mathcal{E}\|_{*}+\lambda \|\mathcal{S}\|_{1} \\
\left\langle\Gamma_{1}, \mathcal{O}-\mathcal{L}-\mathcal{S}\right\rangle+\left\langle\Gamma_{2}, \mathcal{L}-\mathcal{Z}\right\rangle
\quad+\left\langle\Gamma_{3}, D_{w}(\mathcal{Z})-\mathcal{F}\right\rangle \\
+ \left\langle\Gamma_{4}, \mathcal{E}-\mathcal{F}\right\rangle
+\frac{\mu}{2}\left(\|\mathcal{O}-\mathcal{L}-\mathcal{S}\|_{F}^{2}\right. \\
\left.+\|\mathcal{L}-\mathcal{Z}\|_{F}^{2}+\left\|D_{w}(\mathcal{Z})-\mathcal{F}\right\|_{F}^{2}+\|\mathcal{E}-\mathcal{F}\|_{F}^{2}\right)
\end{array}
\end{equation}
with the constraints: $\mathcal{L}=\mathcal{C} \times_{1} \mathbf{U}_{1} \times_{2} \mathbf{U}_{2} \times_{3} \mathbf{U}_{3}, \mathbf{U}_{i}^{T} \mathbf{U}_{i}=$
I, where $\mu$ is the penalty parameter, and $\Gamma_{i}(i=1,2,3)$ are the Lagrange multipliers. 
Next we utilized alternating minimization based on ADMM to split (\ref{equa:Lagrangian}) into multiple sub-problems, then update each variable $\mathcal{L, Z, F, E, S}$ alternately. 
In $(k+1)$ th iteration, variables involved in (\ref{equa:Lagrangian}) can be updated as follows:

1) Update $\mathcal{C}, \mathbf{U}_{i}, \mathcal{L}$. With the other variables fixed, the subproblem of $\mathcal{L}$ can be reformulated as follows:
\begin{equation}
	\begin{aligned}
	\min _{\mathbf{U}_{i}^{T} \mathbf{U}_{i}=\mathbf{I}} \mu \| \mathcal{C} \times_{1} \mathbf{U}_{1} \times_{2} \mathbf{U}_{2} \times_{3} \mathbf{U}_{3}-\frac{1}{2}\left(\mathcal{O}-\mathcal{S}^{(k)}\right. \\
	\left.\quad+\mathcal{Z}^{(k)}+\left(\Gamma_{1}^{(k)}-\Gamma_{2}^{(k)}\right) / \mu\right) \|_{F}^{2}
	\end{aligned}
	\nonumber
\end{equation}
Based on the HOOI algorithm \cite{HOOI}, one can get the $\mathcal{C}^{(k+1)},$ and $\mathbf{U}_{1}^{(k+1)}, \mathbf{U}_{2}^{(k+1)}$ and $\mathbf{U}_{3}^{(k+1)},$ then we have
\begin{equation}
\label{equation:X_solution}
\mathcal{L}^{(k+1)}=\mathcal{C}^{(k+1)} \times_{1} \mathbf{U}_{1}^{(k+1)} \times_{2} \mathbf{U}_{2}^{(k+1)} \times_{3} \mathbf{U}_{3}^{(k+1)}
\end{equation}

2) Update $\mathcal{Z}$. With the other variables fixed, the subproblem of $\mathcal{Z}$ can be reformulated as follows:
\begin{equation}
	\begin{aligned}
	\mathcal{Z}^{(k+1)} &=\operatorname{argmin}_{\mathcal{Z}}\left\langle\Gamma_{2}^{(k)}, \mathcal{L}^{(k+1)}-\mathcal{Z}\right\rangle+\left\langle\Gamma_{3}^{(k)}, D_{w}(\mathcal{Z})-\right.\\
	\left.\mathcal{F}^{(k)}\right\rangle &+\frac{\mu}{2}\left(\left\|\mathcal{L}^{(k+1)}-\mathcal{Z}\right\|_{F}^{2}+\left\|D_{w}(\mathcal{Z})-\mathcal{F}^{(k)}\right\|_{F}^{2}\right)
	\end{aligned}
	\nonumber
\end{equation}
which can be treated as solving the following linear system:
$$
\left(\mu \mathbf{I}+\mu D_{w}^{*} D_{w}\right) \mathcal{Z}=\mu \mathcal{L}^{(k+1)}+\mu D_{w}^{*}\left(\mathcal{F}^{(k)}\right)+\Gamma_{2}^{(k)}-D_{w}^{*}\left(\Gamma_{3}^{(k)}\right)
$$
where $D_{w}^{*}$ denotes the adjoint operator of $D_{w}$. 
By using the fast Fourier transform (FFT) method \cite{LRTDTV}, we have

\begin{equation}
\label{equation:Z_solution}
\left\{\begin{array}{l}
\mathrm{T}_{z}=\mu \mathcal{L}^{(k+1)}+\mu D_{w}^{*}\left(\mathcal{F}^{(k)}\right)+\Gamma_{2}^{(k)}-D_{w}^{*}\left(\Gamma_{3}^{(k)}\right) \\
\mathrm{P}_{z}=w_{1}^{2}\left|\mathrm{fftn}\left(D_{1}\right)\right|^{2}+w_{2}^{2}\left|\mathrm{fftn}\left(D_{2}\right)\right|^{2}+w_{3}^{2}\left|\mathrm{fftn}\left(D_{3}\right)\right|^{2} \\
\mathcal{Z}^{(k+1)}=\operatorname{ifftn}\left(\frac{\mathrm{fftn}\left(\mathrm{T}_{z}\right)}{\mu \mathbf{1}+\mu \mathrm{P}_{z}}\right)
\end{array}\right.
\end{equation}
where $\mathrm{fftn}$ and $\mathrm{ifftn}$ indicate fast $3 \mathrm{D}$ Fourier transform and its inverse transform, respectively. $|\cdot|^{2}$ is the elements-wise square, and the division is also performed element-wisely.

3) Update $\mathcal{F}$. With the other variables fixed, the subproblem of $\mathcal{F}$ can be reformulated as follows:
\begin{equation}
\begin{aligned}
\mathcal{F}^{(k+1)}=\underset{\mathcal{F}}{\operatorname{argmin}} \tau\|\mathcal{F}\|_{1}+\frac{\mu}{2}\left\|\mathcal{F}-\frac{T_1+T_2}{2}\right\|_{F}^{2}
\end{aligned}
\nonumber
\end{equation}
where $T_1=D_{w}\left(\mathcal{Z}^{(k+1)}\right)+\frac{\Gamma_{3}^{(k)}}{\mu}, T_2=\mathcal{E}^{(k+1)}+\frac{\Gamma_{4}^{(k)}}{\mu}$.
By introducing the so-called soft-thresholding operator:
\begin{equation}
\nonumber
	\mathcal{R}_{\Delta}(\mathbf{x})=\left\{\begin{array}{rlr}
	x-\Delta, & \text { if } & \mathbf{x}>\Delta \\
	x+\Delta, & \text { if } & \mathbf{x}<\Delta \\
	0, & \text {otherwise}
	\end{array}\right.
\end{equation}
where $x \in \mathbb{R}$ and $\Delta>0,$ then we can update $\mathcal{F}^{k+1}$ as
\begin{equation}
\label{equation:F_solution}
\mathcal{F}^{k+1}=\mathcal{R}_{\frac{\tau}{\mu}}\left(\frac{T_1+T_2}{2}\right)
\end{equation}

4) Update $\mathcal{E}$. With the other variables fixed, the subproblem of $\mathcal{E}$ can be reformulated as follows:
\begin{equation}
\label{equation:E_solution}
\begin{aligned}
&\mathcal{E}^{(k+1)}=\underset{\mathcal{E}}{\operatorname{argmin}} \|\mathcal{E}\|_{*}+\frac{\mu}{2 \alpha}\left\|\mathcal{E}-\left(\mathcal{F}-\frac{\Gamma_{4}^{(k)}}{\mu}\right)\right\|_{F}^{2}\\
&=\operatorname{SVT}\left(\mathcal{F}-\frac{\Gamma_{4}^{(k)}}{\mu}, \frac{\alpha}{\mu}\right)
\end{aligned}
\end{equation}

) Update $\mathcal{S}$. Similarly, we should consider
\begin{equation}
\begin{aligned}
&\mathcal{S}^{(k+1)}=\underset{\mathcal{S}}{\operatorname{argmin}} \lambda \|\mathcal{S}\|_{1}+\frac{\mu}{2}\left\|\mathcal{S}-\left(\mathcal{O}-\mathcal{L}^{(k+1)}+\frac{\Gamma_{1}^{(k)}}{\mu}\right)\right\|_{F}^{2},
\end{aligned}
\nonumber
\end{equation}
similar to the update of $\mathcal{F}$, we have
\begin{equation}
\label{equation:S_solution}
	\mathcal{S}^{k+1}=\mathcal{R}_{\frac{\lambda}{\mu}}\left(\mathcal{O}-\mathcal{L}^{(k+1)}+\frac{M_{1}^{(k)}}{\mu}\right)
\end{equation}

Finally, we update multipliers $\Gamma_{i} (i=1, 2, 3)$, according to the ALM,
\begin{equation}
\label{equa:Multipliers}
	\left\{\begin{array}{l}
	\Gamma_{1}^{(k+1)}=\Gamma_{1}^{(k)}+\mu\left(\mathcal{O}-\mathcal{L}^{(k+1)}-\mathcal{S}^{(k+1)}\right) \\
	\Gamma_{2}^{(k+1)}=\Gamma_{2}^{(k)}+\mu\left(\mathcal{L}^{(k+1)}-\mathcal{Z}^{(k+1)}\right) \\
	\Gamma_{3}^{(k+1)}=\Gamma_{3}^{(k)}+\mu\left(D_{w}\left(\mathcal{Z}^{(k+1)}\right)-\mathcal{F}^{(k+1)}\right)
	\end{array}\right.
\end{equation}

\subsection{Time Complexity Analysis}

The main per-iteration cost lies in the update of $\mathcal{E}, \mathcal{Z}, \mathcal{F}, \mathcal{S}$.
The update of $\mathcal{E}$ requires computing FFT and $p$ SVDs of $m\times n$ matrices.
The updates of $\mathcal{S}$ and $\mathcal{N}$ only need to perform basic tensor operations.
Thus, the per-iteration complexity is $O(m n p \log (p)+m_{(1)} n_{(1)}^{2} p)$,
where $m_{(1)} = \max(m, n)$, $n_{(1)} = \min(m, n)$.

\begin{algorithm}
	\caption{TDLRSTV for HSI denoising.} \label{algorightm-1}
	\begin{algorithmic}[1]
		\Require
		$m \times n \times p$ observed HSI $ \mathcal{O}$, stopping criterion $\varepsilon$, and regularization parameters $\lambda$, $\tau$, $\alpha$.
		\Ensure
		Denoised image $\mathcal{L}$;
		\State Initialize: 		
					
		1st step: Update $\mathcal{L}, $ via	(\ref{equation:X_solution}) 
		
		2nd step: Update $\mathcal{Z}$ via	(\ref{equation:Z_solution})
		
		3rd step: Update $\mathcal{F}$ via (\ref{equation:F_solution})
		
		4th step: Update $\mathcal{E}$ via	(\ref{equation:E_solution}) 
		
		5th step: Update $\mathcal{S}$ via (\ref{equation:S_solution})
		
		6th step: Update $\Gamma_{1}, \Gamma_{2}, \Gamma_{3}$ via (\ref{equa:Multipliers})
		
		7th step: Update the parameter via
		
		 $\mu:=\min \left(\rho \mu, \mu_{\max }\right)$
		
		\State Check the convergence condition\\
		$\max \left\{\left\|\mathcal{O}-\mathcal{L}^{k+1}-\mathcal{S}^{k+1}\right\|_{\infty},\left\|\mathcal{L}^{k+1}-\mathcal{Z}^{k+1}\right\|_{\infty}\right\}\leq \varepsilon.$
	\end{algorithmic}
\end{algorithm}

\section{Experimental Results and Discussion}
\label{results}

In this section, we applied the proposed TDLRSTV to both public simulation and real HSI datasets, and compare the results with that of several state-of-the-arts to evaluate the performance of TDLRSTV.

\subsection{Experimental Setting}

\noindent \textbf{Benchmark Datasets.} Four HSIs datasets are tested:

\begin{figure}
	\centering
	\includegraphics[width=1\linewidth]{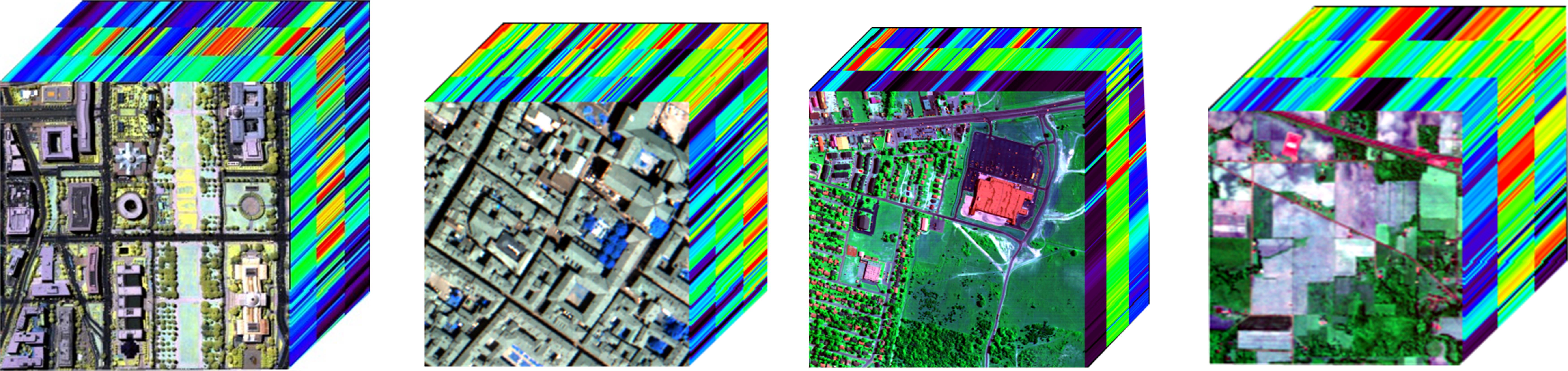}
	\caption{Simulation and real HSI data sets tested in the experiments. From left to right: Washington DC Mall data set, Pavia Centre data set, HYDICE urban data set and AVIRIS Indian Pines data set.}
	\label{fig:3ddatasets}
\end{figure}

\begin{itemize}
	\item
Pavia City Center \cite{PaviaCentre} dataset.
The raw Pavia Center dataset has a size of 1096$\times$1096$\times$102, i.e., there is 102 bands and each band is a $1096\times1096$ 2-D image.
After removing the bands heavily polluted by noise, sub-block with a size of $200 \times 200 \times 80$ are selected for simulation experiments in this section.

\item
Washington DC Mall data \cite{AVIRIS_Indian_Pines}, the raw data has a size of 1208$\times$307$\times$191.
As \cite{zengTGRS}, sub-block with size 256$\times$256$\times$191 are selected.

\item
Indian Pines dataset \cite{AVIRIS_Indian_Pines} with size $145\times145 \times 220$ was collected by the Airborne Visible Infrared Imaging Spectrometer (AVIRIS). This dataset was selected for real data experiments.

\item
HYDICE Urban HSI dataset \cite{Urban} has size of $207\times207\times210$.
After deleting the water absorption band, the remaining 189 bands were selected for testing. 
This dataset was also selected for real data experiments. 
\end{itemize}

\noindent \textbf{Baselines.} 
Seven SOTA HSI denoising methods
are employed as the comparison models, i.e., LRTA based on Tucker decomposition \cite{LRTA_tucker}, BM4D \cite{BM4D} based on nonlocal self-similarity, LRMR via low-rank matrix approximation \cite{LRMR}, LRTV which combines total variation and LRMR \cite{LRTV}, NAILRMA based on iterative LRMR and noise estimation \cite{NAILRMA}, LRTDTV based on total variation
and low-rank tensor Tucker decomposition \cite{LRTDTV} and spectral-spatial total variation (SSTV) \cite{SSTV}.

\noindent \textbf{Evaluation Indexes.} To measure the performance of tested models, both qualitative visual evaluation and quantitative index are utilized.
Quantitative picture quality indices (PQI): 
peak signal-to-noise ratio (PSNR), 
structural similarity (SSIM) and feature similarity (FSIM) \cite{FSIM}, which evaluate the quality of the HSI spatial dimension, and erreur relative globale adimensionnelle
de synthèse (EGRAS) \cite{EGRAS} and spectral angle mapper (SAM) \cite{SAM}, 
which evaluate the quality of the HSI spectral dimension. 
In addition, all experiments are performed on MATLAB(2019b), Intel core i7@2.2GHz with 64 GB RAM (Windows).

\subsection{Simulated HSI Data Experiments}

\noindent \textbf{Mixed noise setting.} 
To simulate HSI's degradation process of noise, nine different noise cases are simulated for the selected HSI data-sets. 
The details are listed as follows:
\begin{itemize}
	\item Case 1, Gaussian noise is added to the all band of the selected data, the mean and variance are setted as zero and  0.1, respectively.

	\item Case 2, the same as Case 1, but the variance is aggravated to 0.2.

	\item Case 3, both Gaussian noise (with 0 mean and 0.05 variance) and sparse pepper-and-salt noise (with percentage 0.1) are added for all bands.

	\item Case 4, the same as Case 3, but the variance of Gaussian noise and the percentage of sparse pepper-and-salt noise are enhanced to 0.075 and 0.15, respectively.

	\item Case 5, the same as Case 4, but the variance of Gaussian noise and the percentage of sparse pepper-and-salt noise are further aggravated to 0.1 and 0.2, respectively.


	\item Case 6, the same as Case 6, but the variance of Gaussian noise is further enhanced to 0.2;


	\item Case 7, the same as Case 8, in addition, the stripe in Case 7 is also added;

	\item Case 8, dead line and stripe are the same as Case 9, while all the bands are added with the same Gaussian noise (0.15 variance) and sparse pepper-and-salt noise (0.2 percentage);

	\item Case 9, the same as Case 10, but the variance of Gaussian noise is aggravated to 0.2.

\end{itemize}

\begin{figure*}[!t] \centering
	\includegraphics[width=1.5\columnwidth]{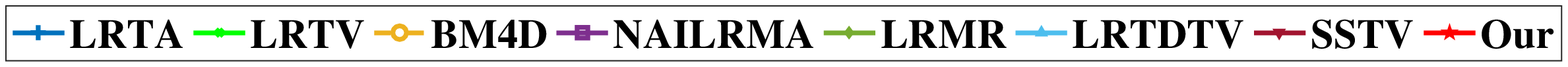}\\ 
	\subfloat[Case 3] {\includegraphics[width=0.31\columnwidth]{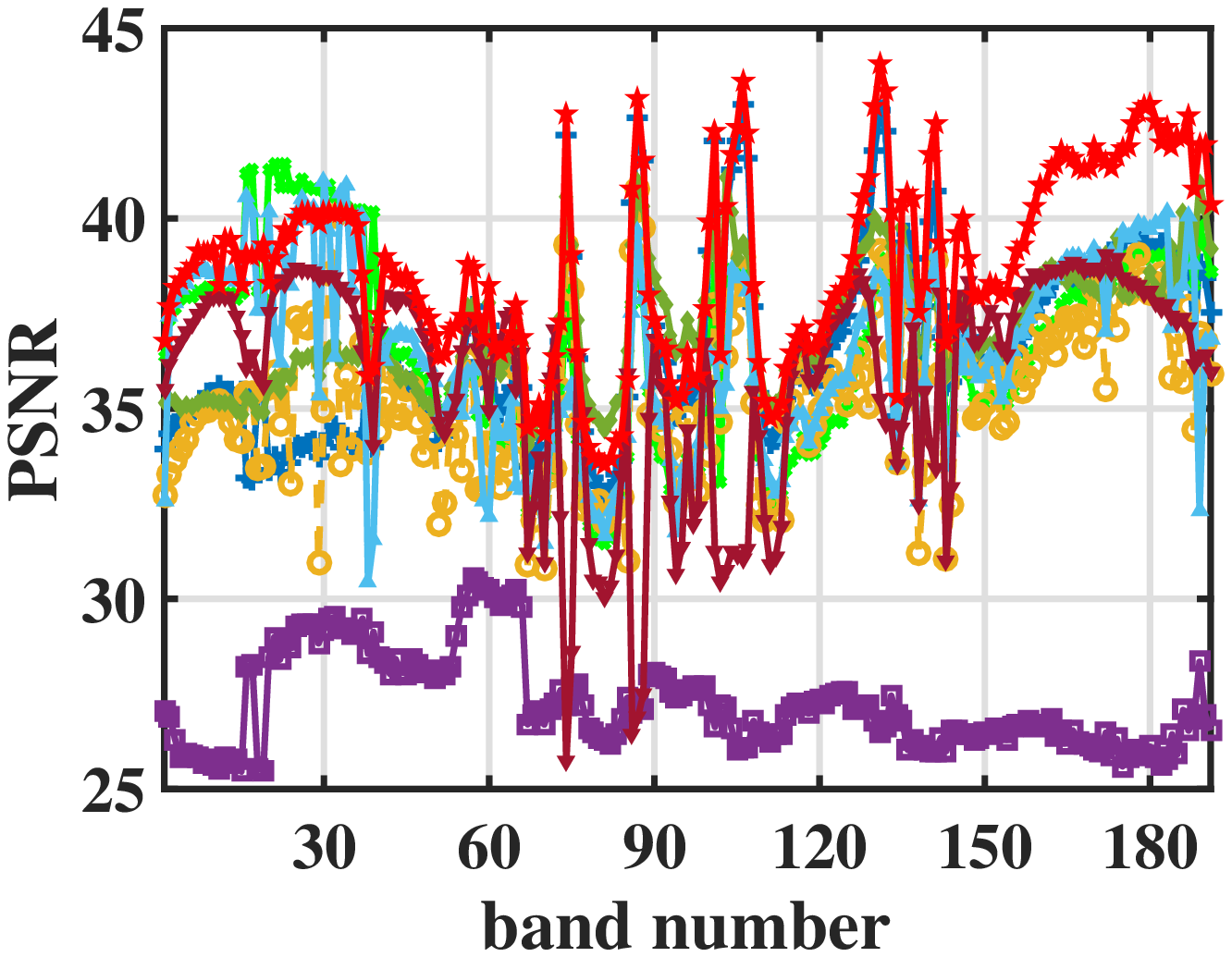}
	}\hspace{0.1mm}
	\subfloat[Case 3] {\includegraphics[width=0.31\columnwidth]{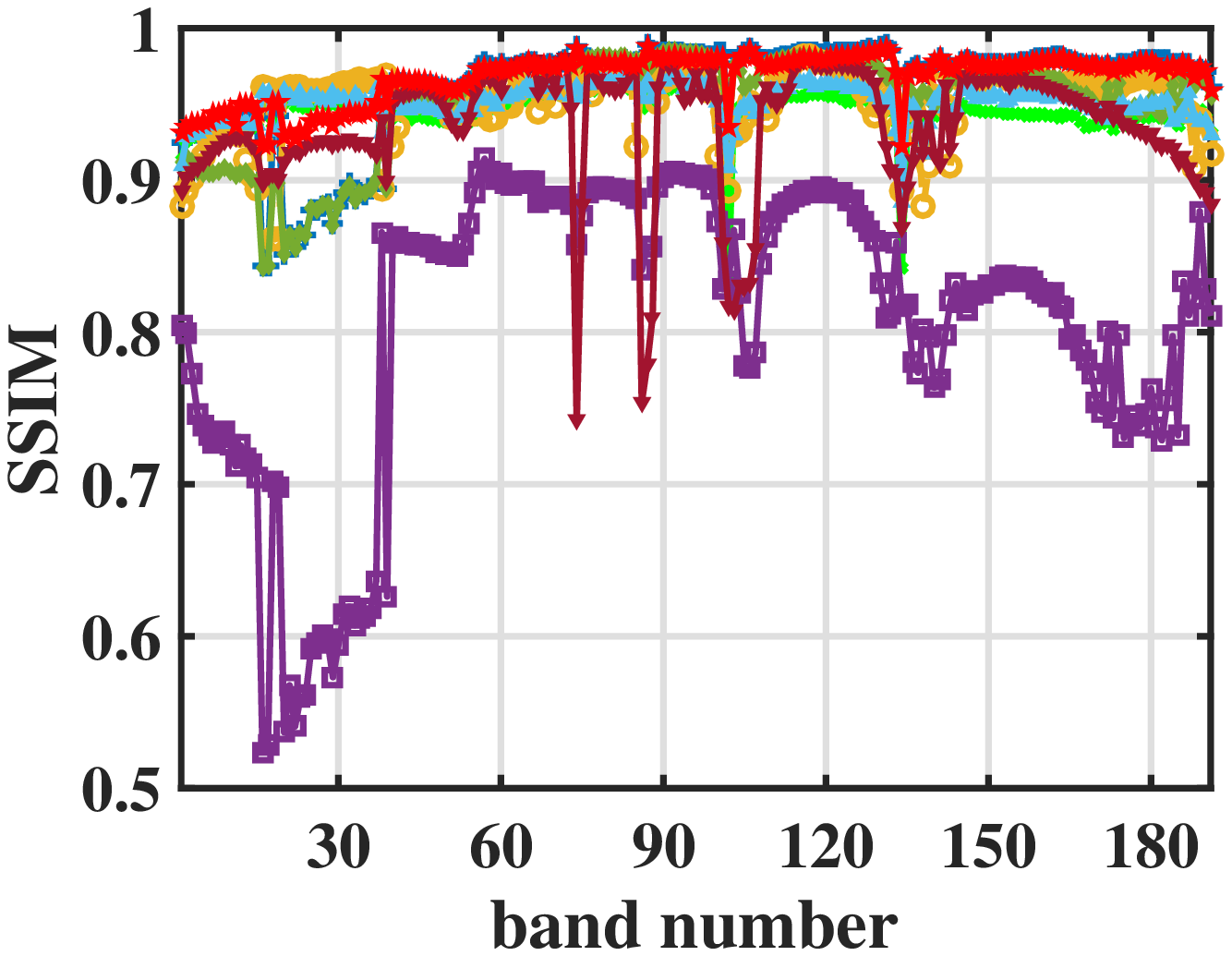}
	}\hspace{0.1mm}
	\subfloat[Case 8]{\includegraphics[width=0.31\columnwidth]{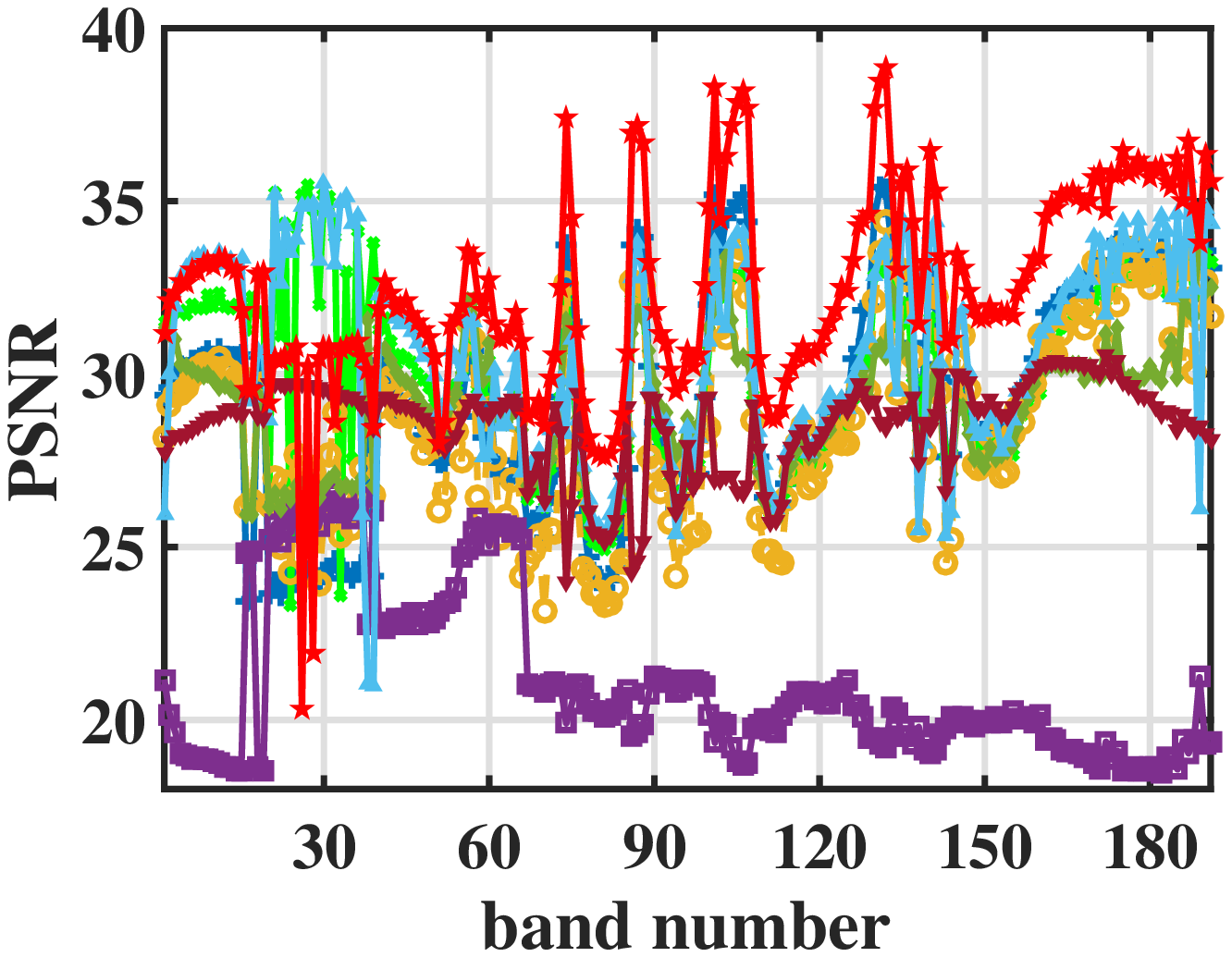}
	}\hspace{0.1mm}
	\subfloat[Case 8]{\includegraphics[width=0.31\columnwidth]{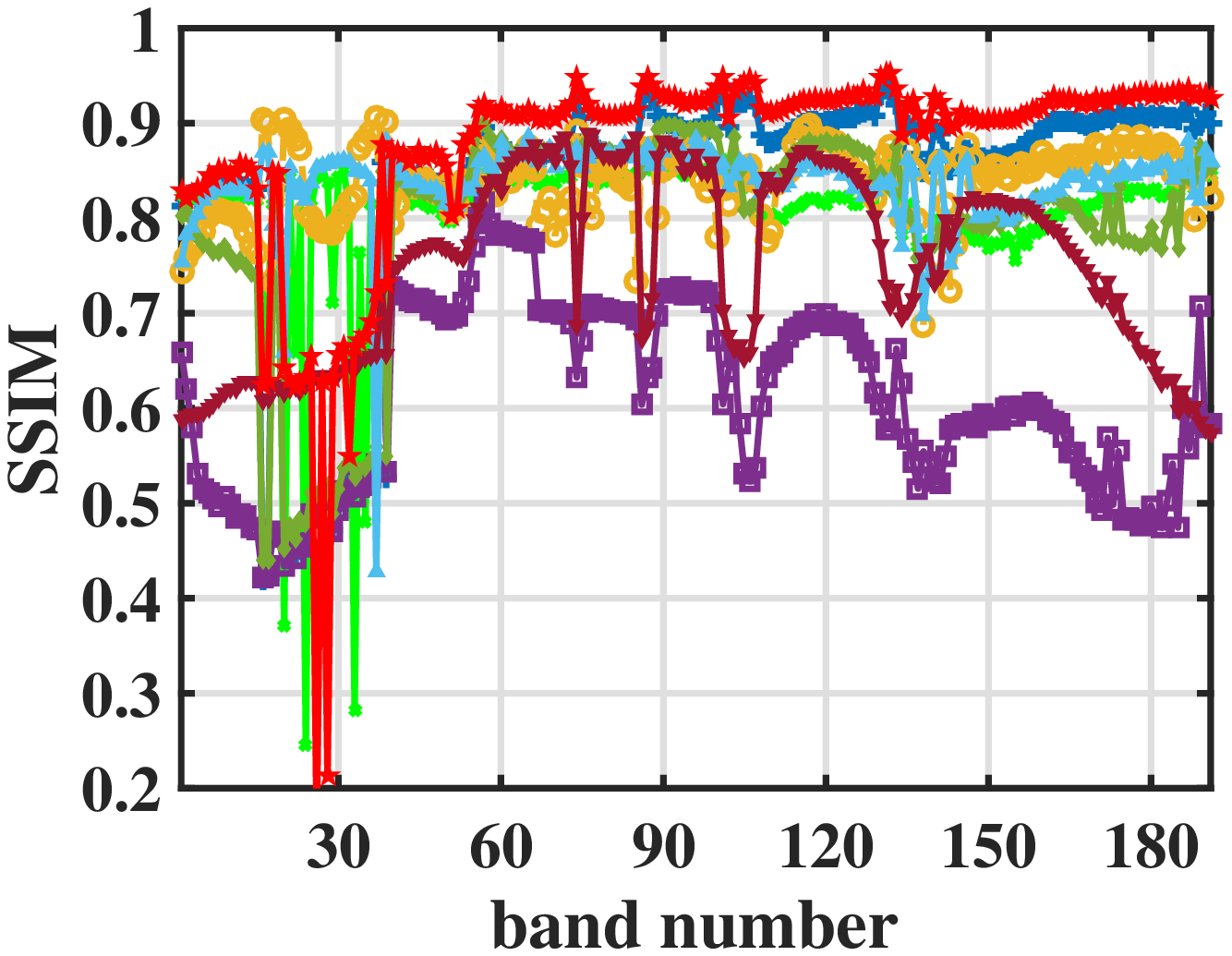}
	}\hspace{0.1mm}
	\subfloat[Case 9]{\includegraphics[width=0.31\columnwidth]{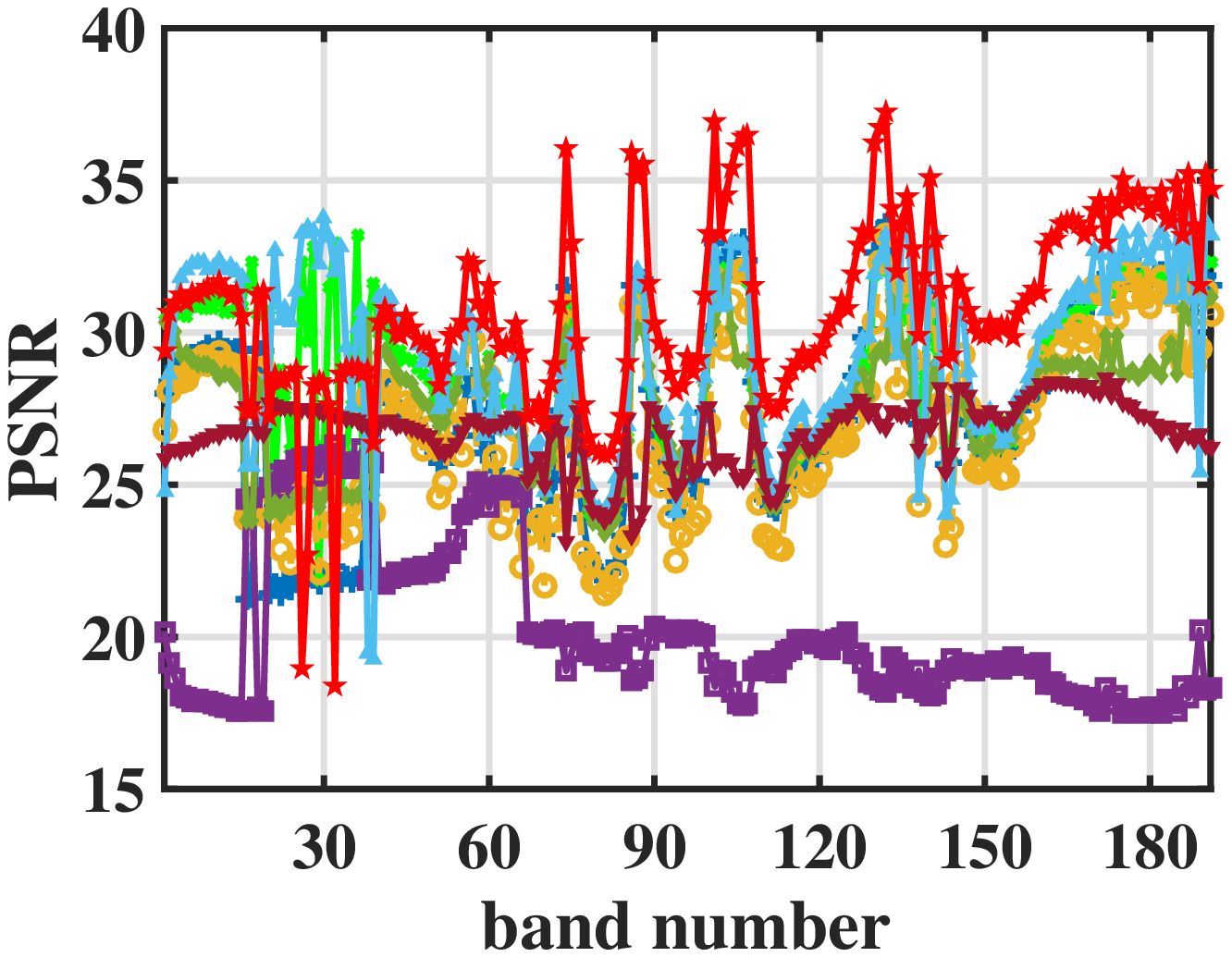}}
	\hspace{0.1mm}
	\subfloat[Case 9]{\includegraphics[width=0.31\columnwidth]{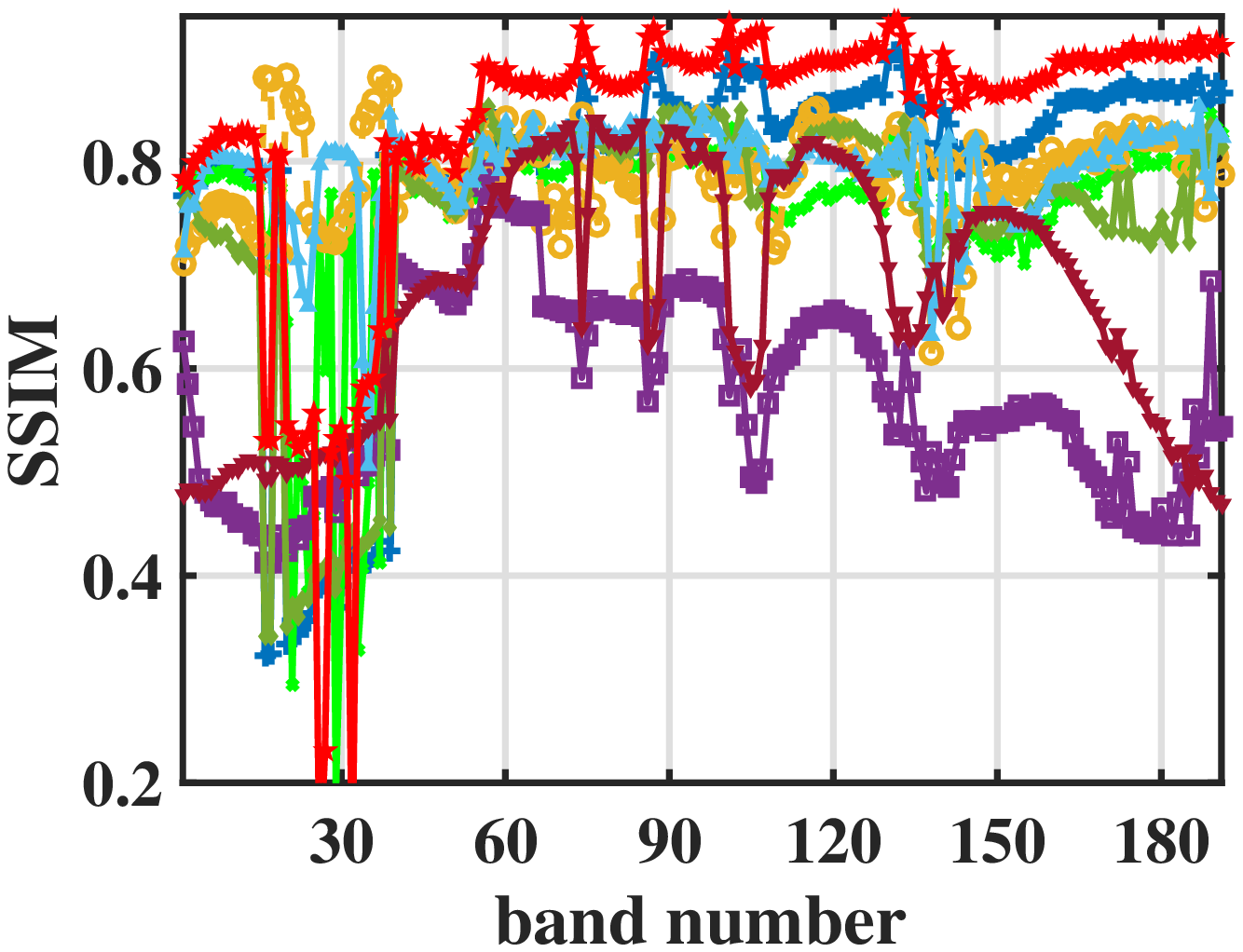}}
	\hspace{0.1mm}
	\caption{Comparison of quantitative evaluation index values, PSNR and SSIM, of each band on the Washington DC Mall data by LRTA, LRTV, BM4D, NAILRMA, LRMR, LRTDTV, SSTV and the proposed TDLRSTV.}
	\label{DC_imshow0302_PSNR_SSIM}
\end{figure*}

\begin{figure*}[htbp] \centering
	\subfloat[Original image] {\includegraphics[width=0.32\columnwidth]{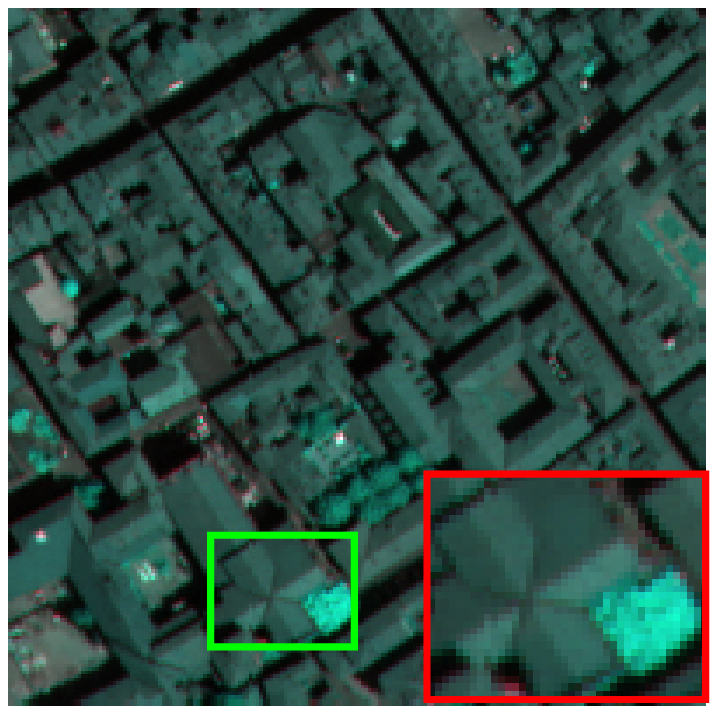}
	}\hspace{0.1mm}
	\subfloat[Noisy one(11.18dB)] {\includegraphics[width=0.32\columnwidth]{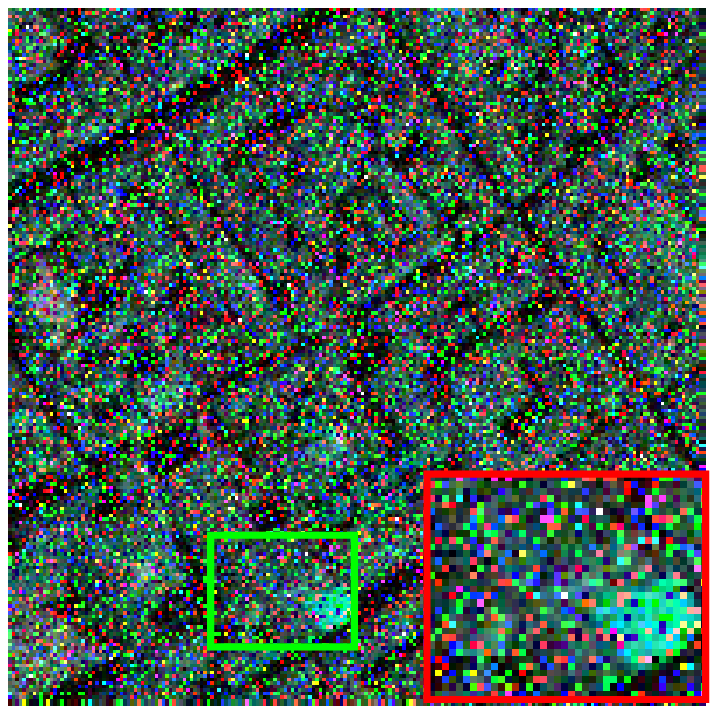}
	}\hspace{0.1mm}
	\subfloat[LRTA(28.38dB)] {\includegraphics[width=0.32\columnwidth]{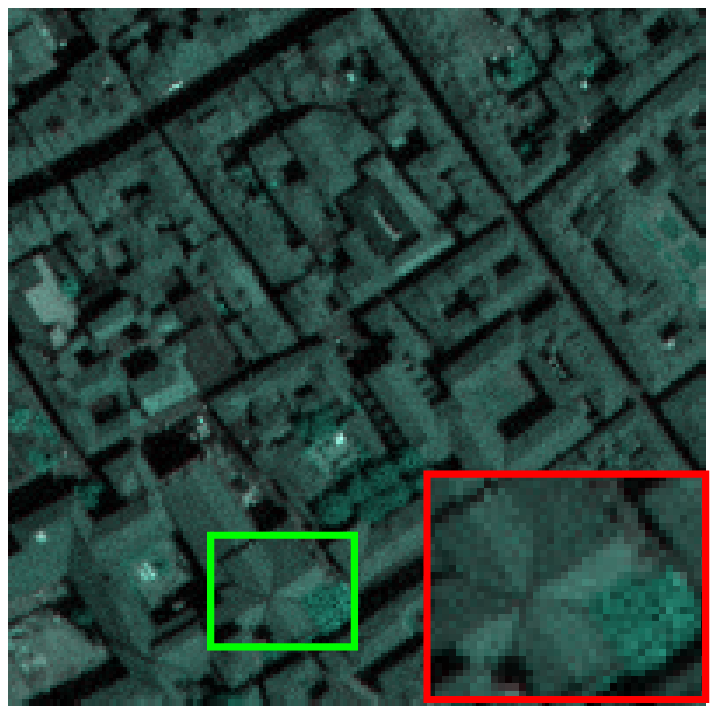}
	}\hspace{0.1mm}
	\subfloat[LRTV(29.38dB)] {\includegraphics[width=0.32\columnwidth]{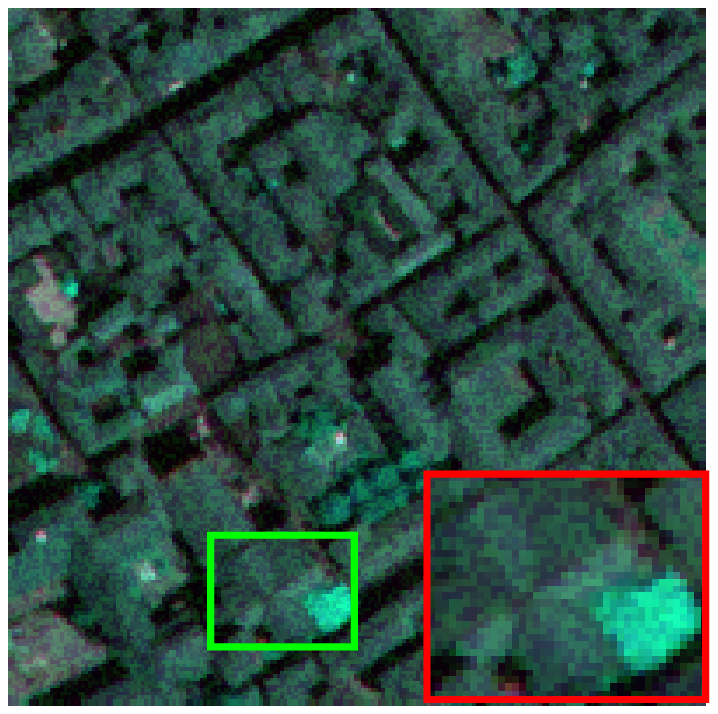}
	}\hspace{0.1mm}
	\subfloat[BM4D(28.52dB)] {\includegraphics[width=0.32\columnwidth]{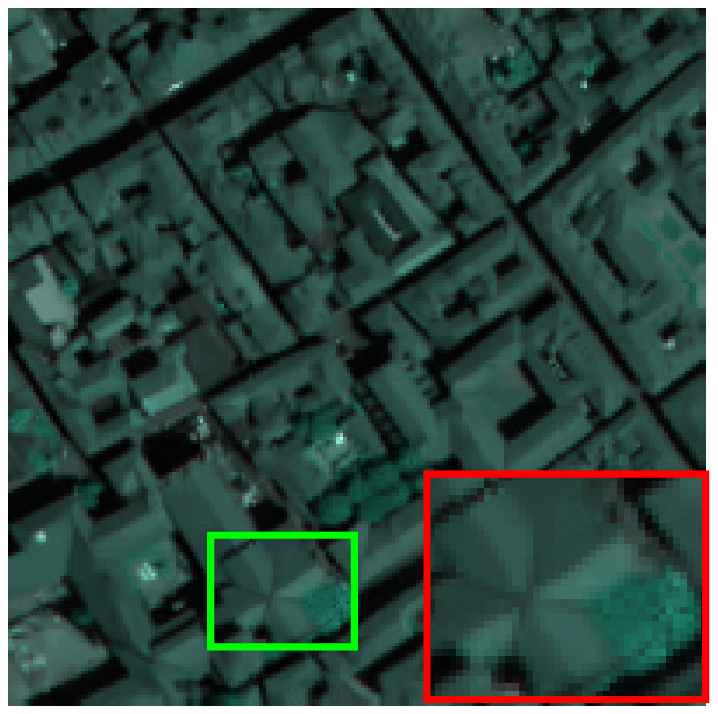}
	}\hspace{0.1mm}
	\subfloat[NALRMA(21.87dB)] {\includegraphics[width=0.32\columnwidth]{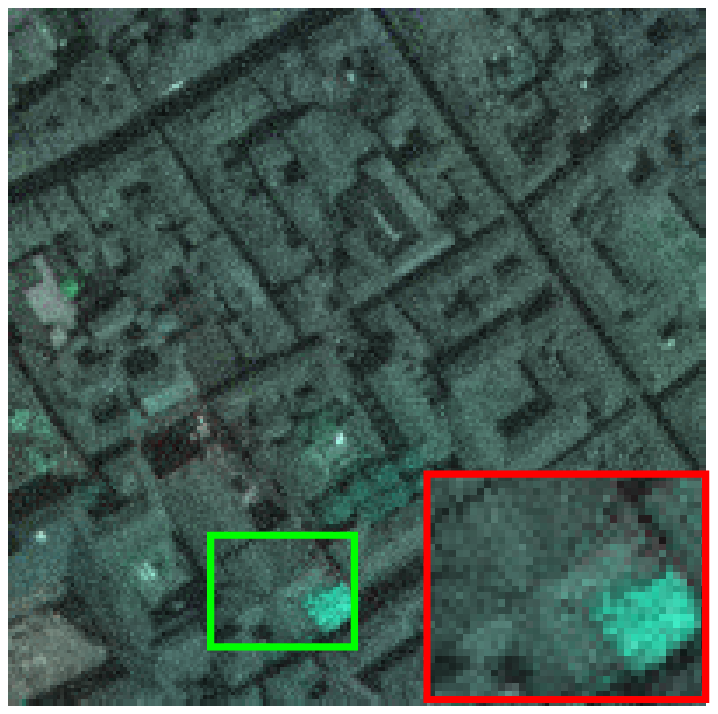}
	}\hspace{0.1mm}
	\subfloat[LRMR(30.77dB)] {\includegraphics[width=0.32\columnwidth]{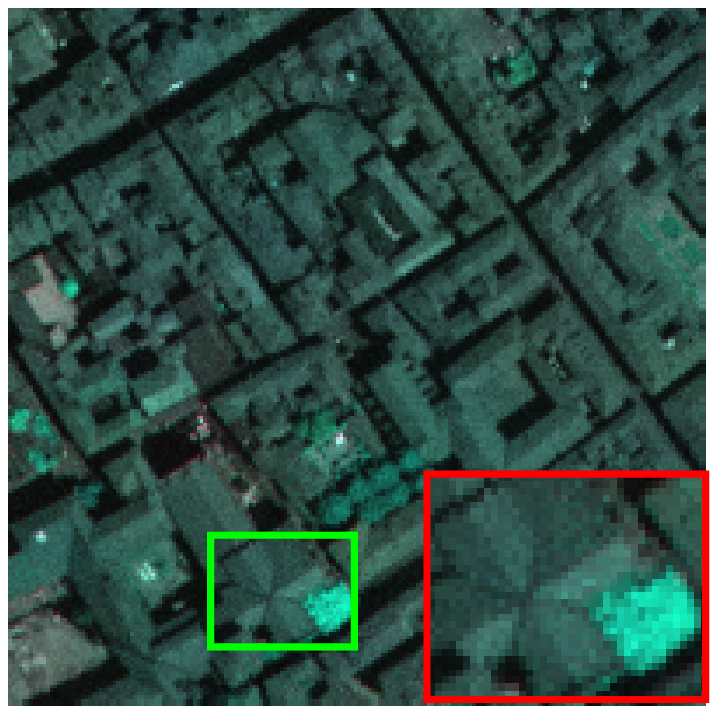}
	}\hspace{0.1mm}
	\subfloat[LRTDTV(31.02dB)] {\includegraphics[width=0.32\columnwidth]{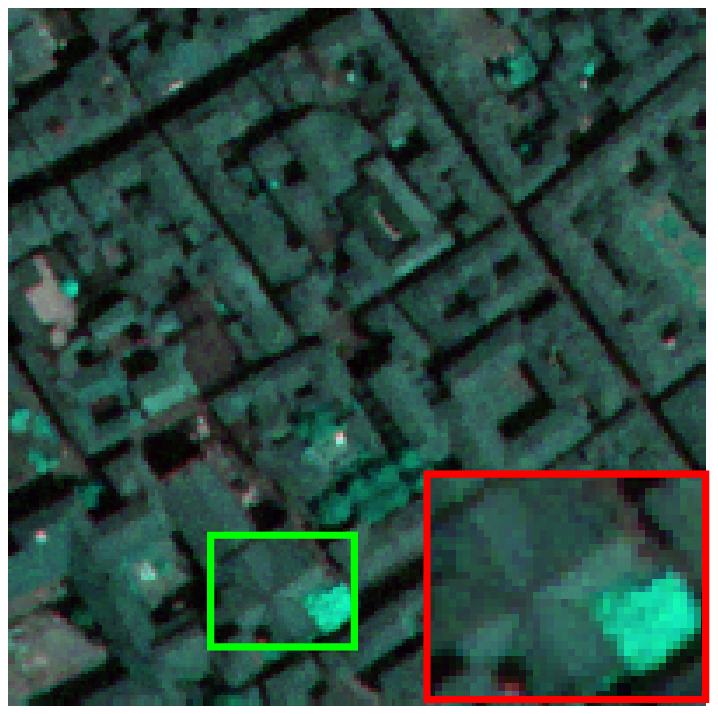}
	}\hspace{0.1mm}
	\subfloat[SSTV(31.31dB)] {\includegraphics[width=0.32\columnwidth]{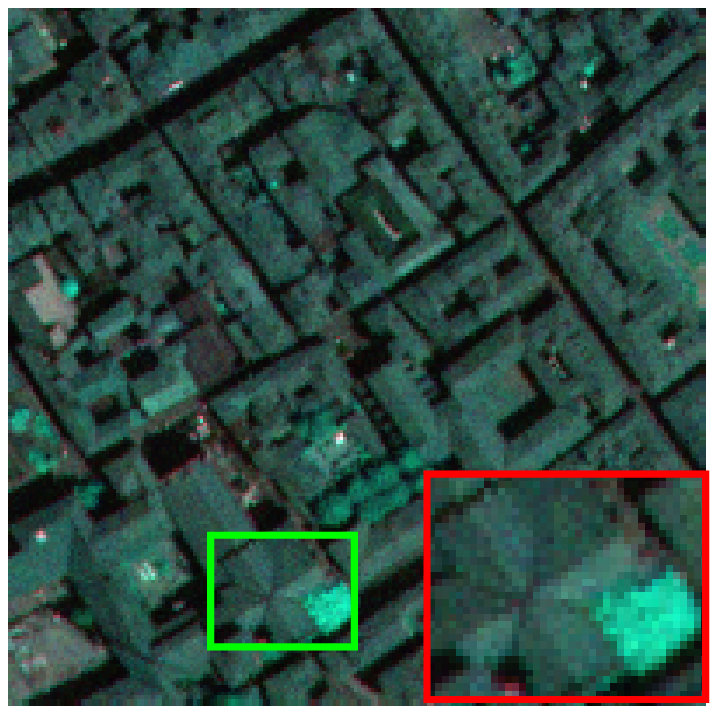}
	}\hspace{0.1mm}
	\subfloat[Our(33.10dB)] {\includegraphics[width=0.32\columnwidth]{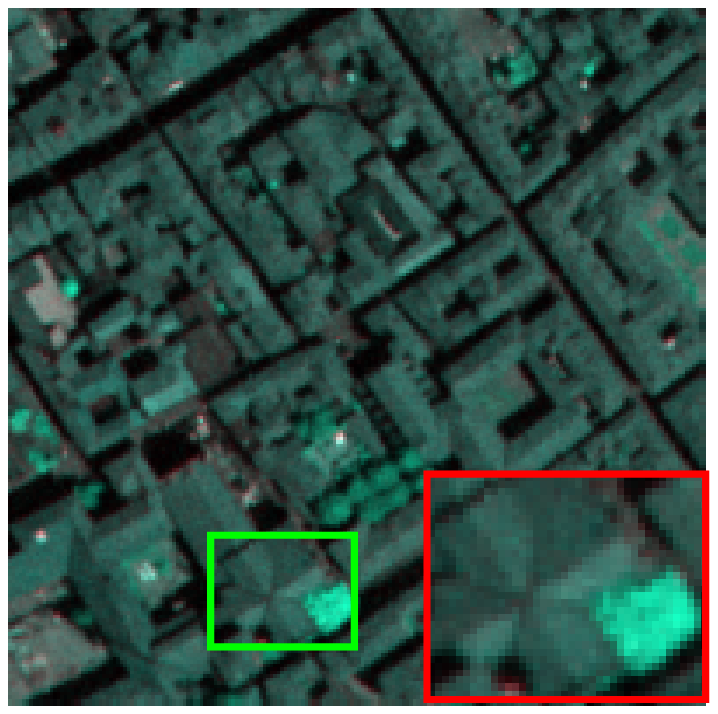}
	}\hspace{0.1mm}
	\caption{Recovery performance comparison on the Pavia centre data by LRTA, LRTV, BM4D, NAILRMA, LRMR, LRTDTV, SSTV and the proposed TDLRSTV. The noise level: Gaussian noise (with 0 mean and 0.1 variance) and sparse pepper-and-salt noise (with percentage 0.2) are added for all bands, i.e., $P=0.2, G=0.2 $. The color image is composed of bands 14 77 58 for the red, green, and blue channels, respectively.}
	\label{Pavia_imshow0202_146}
\end{figure*}

\noindent \textbf{Visual comparison.}
To visually illustrate the denoising performance of the proposed LRSTV, in two different noise situations, for the Washington DC Mall and Pavia data sets, we have selected 3 bands to synthesize a pseudo-color image, as shown in Fig. \ref{DC_imshow0302_PSNR_SSIM} and Fig. \ref{Pavia_imshow0202_146}. In these two figures, the red demarcated window is the magnified area of the area marked by the green demarcated window. From the magnified area, one can clearly see that compared with other competing methods, the proposed method effectively removes the noise while recovering the tiny details and textures.

\noindent \textbf{Quantitative comparison.} For Washington DC Mall and Pavia, Table \ref{Table:WDC} and Table \ref{Table:Pavia} list the five quantitative evaluation index values (PQIs) of all competing methods. Bold shows the optimal PSNR, SSIM, FSIM, ERGAS and SAM values. 
The following observations can be obtained from the table. 
First of all, under most noise conditions, the proposed LRSTV always achieves the best performance among the five evaluation indicators, which fully proves the effectiveness of LRSTV for HSI denoising tasks. 
Second, for the mixed noise in Washington DC Mall, as the noise level increases, the advantages of our method over other methods become more obvious. 
One can observe that when the Gaussian noise variance is 0.2, the proposed method almost exceeds 2.2dB than LRTDTV. 
In addition, we took four noise cases as an example, and plotted the PSNR and SSIM values of each band, as shown in Fig. \ref{DC_imshow0302_PSNR_SSIM}. 
From the figure one can see that the PSNR value and SSIM value of most bands obtained by the proposed model are significantly higher than other methods.

\begin{figure*}[htbp] \centering
	\captionsetup[subfloat]{labelsep=none,format=plain,labelformat=empty}
	\subfloat[Noisy image] {\includegraphics[width=0.35\columnwidth]{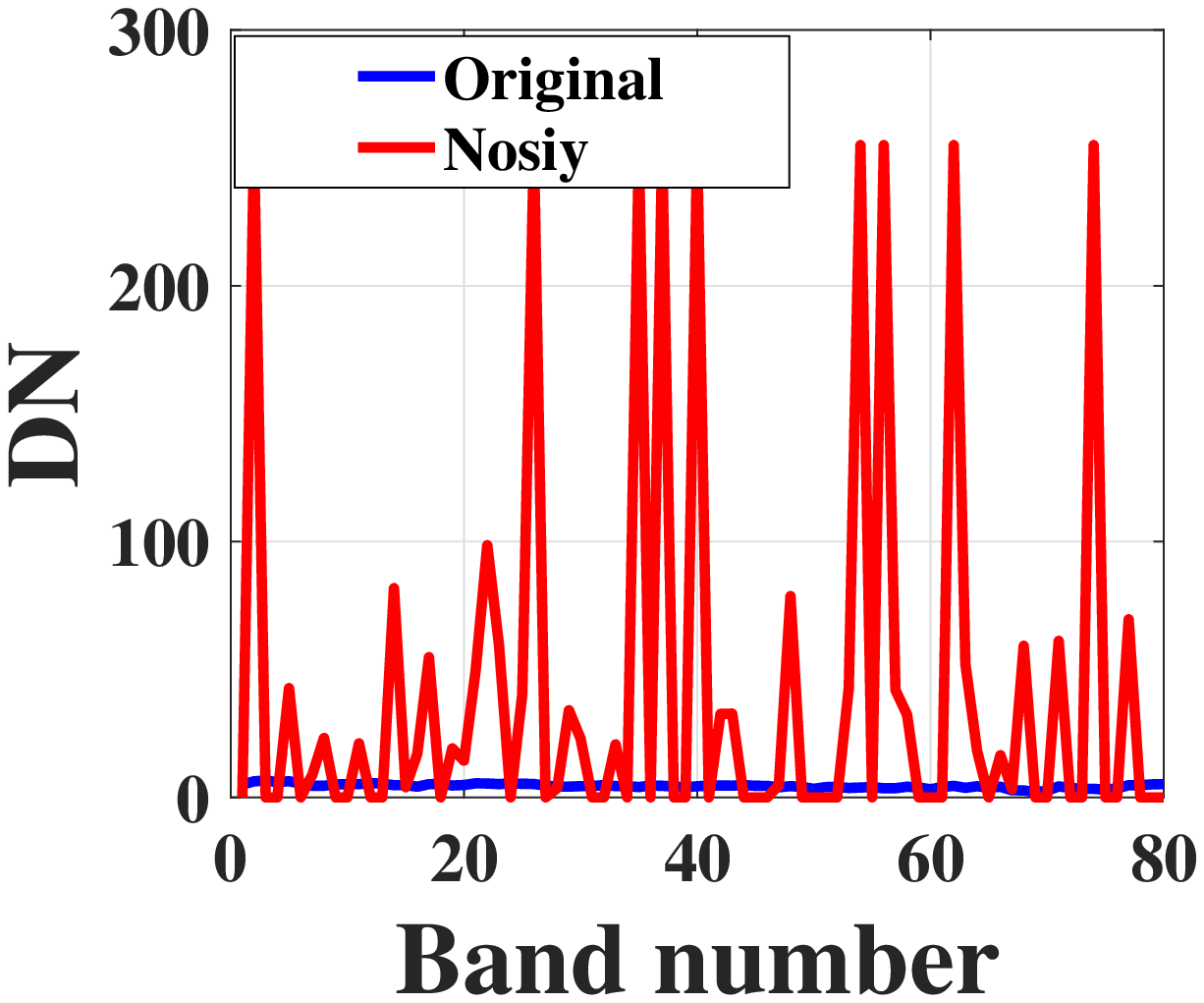}
	}\hspace{0.3mm}
	\subfloat[LRTA] {\includegraphics[width=0.35\columnwidth]{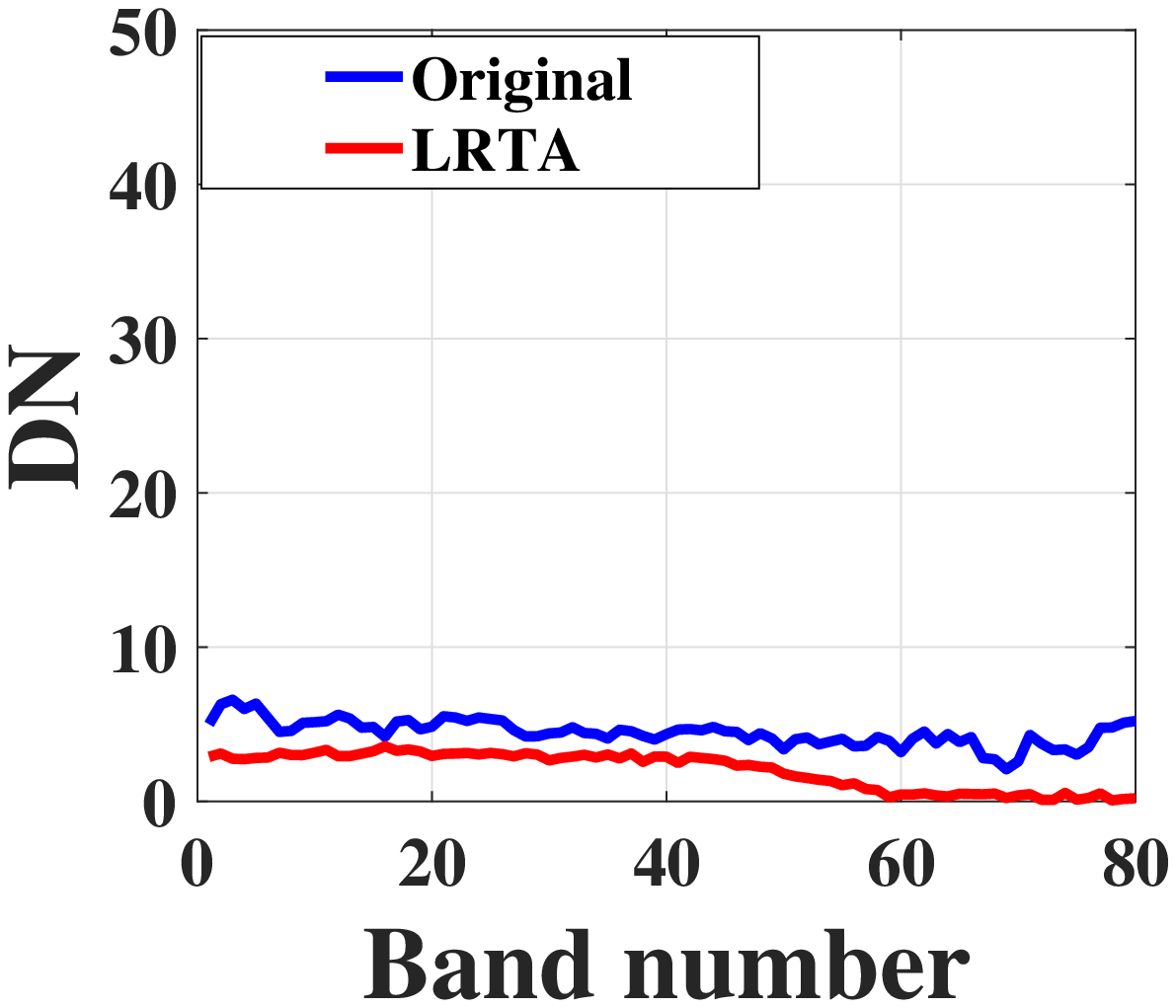}
	}\hspace{0.3mm}
	\subfloat[LRTV] {\includegraphics[width=0.35\columnwidth]{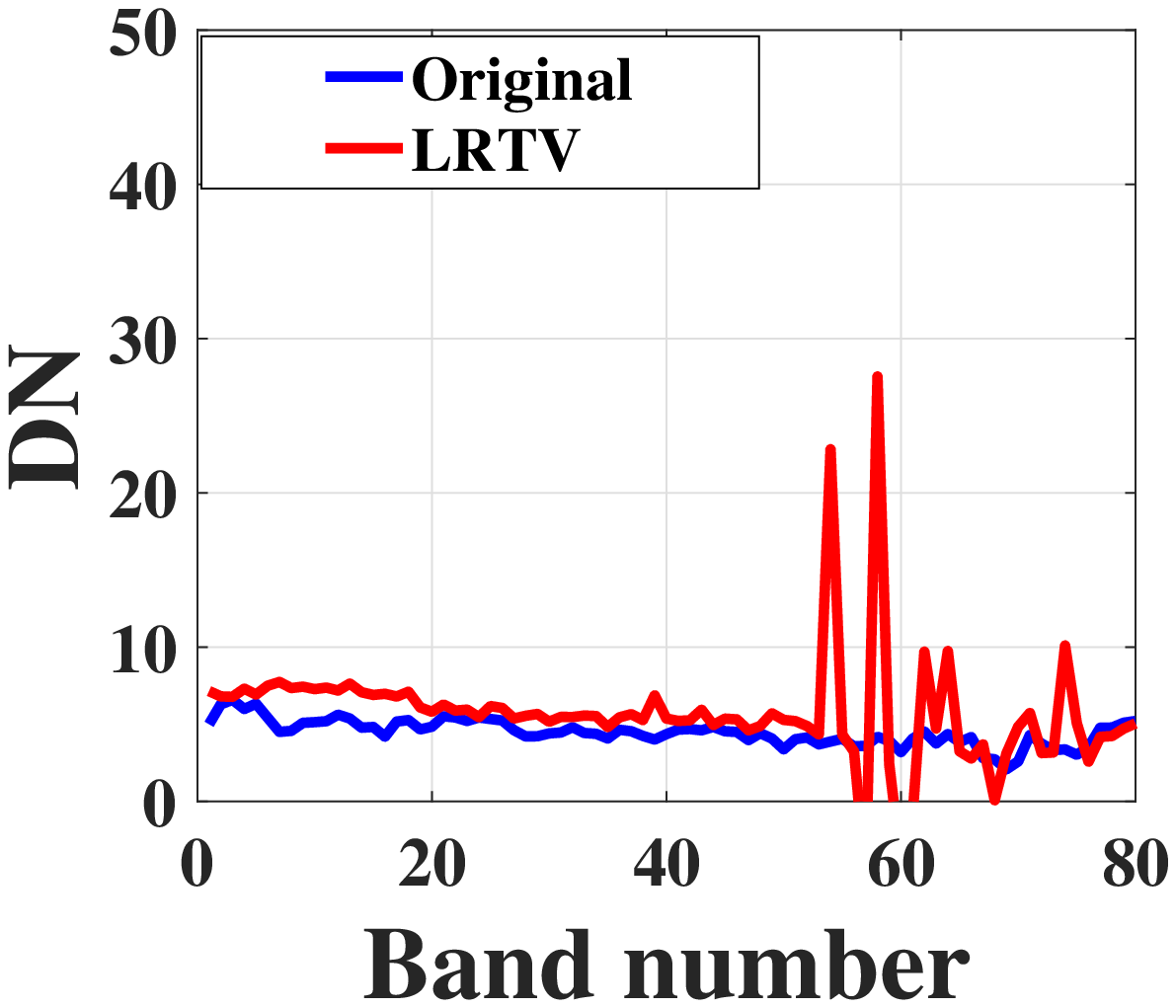}
	}\hspace{0.3mm}
	\subfloat[BM4D] {\includegraphics[width=0.35\columnwidth]{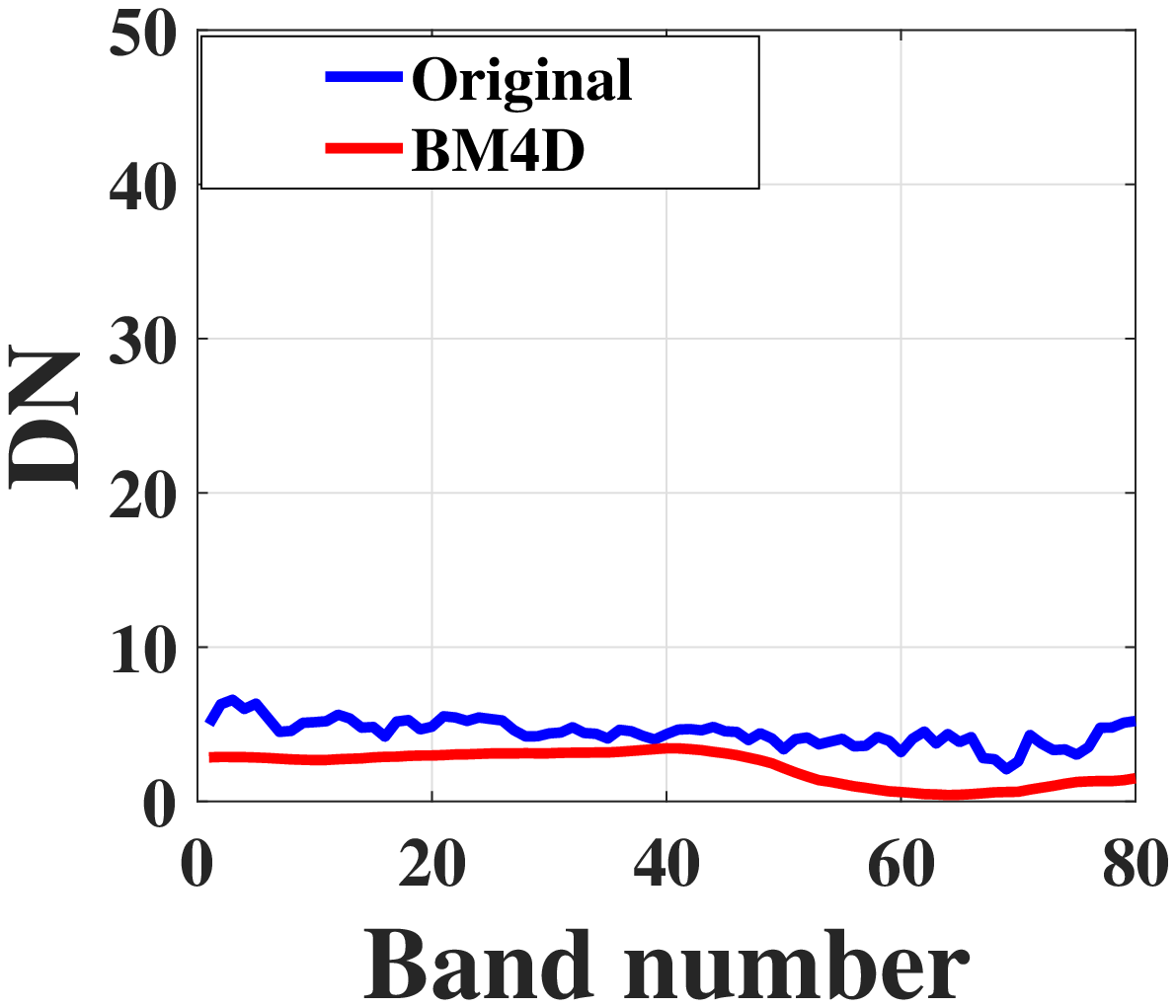}
	}\hspace{0.3mm}
	\subfloat[LRMR] {\includegraphics[width=0.35\columnwidth]{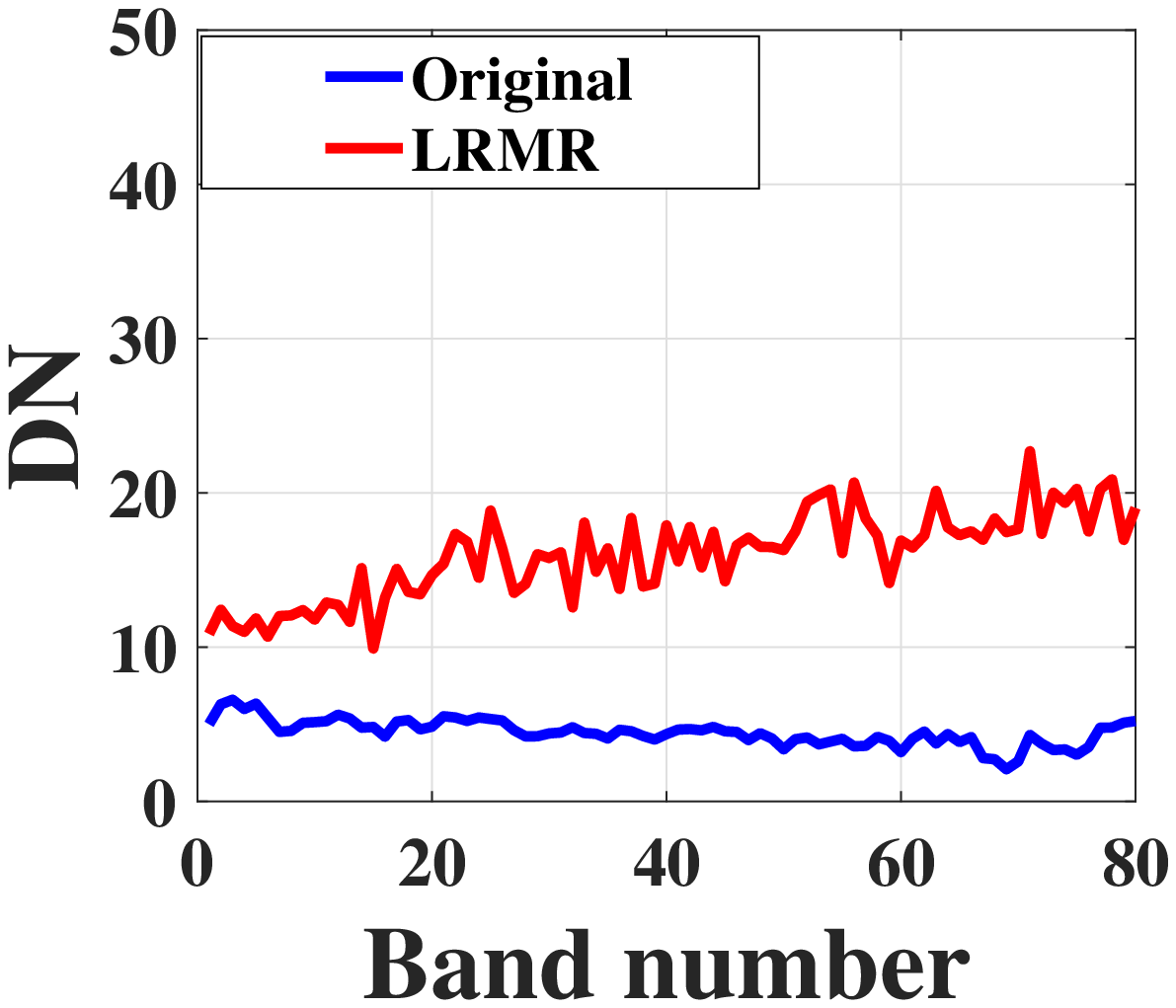}
	}\hspace{0.3mm}
	\subfloat[LRTDTV] {\includegraphics[width=0.35\columnwidth]{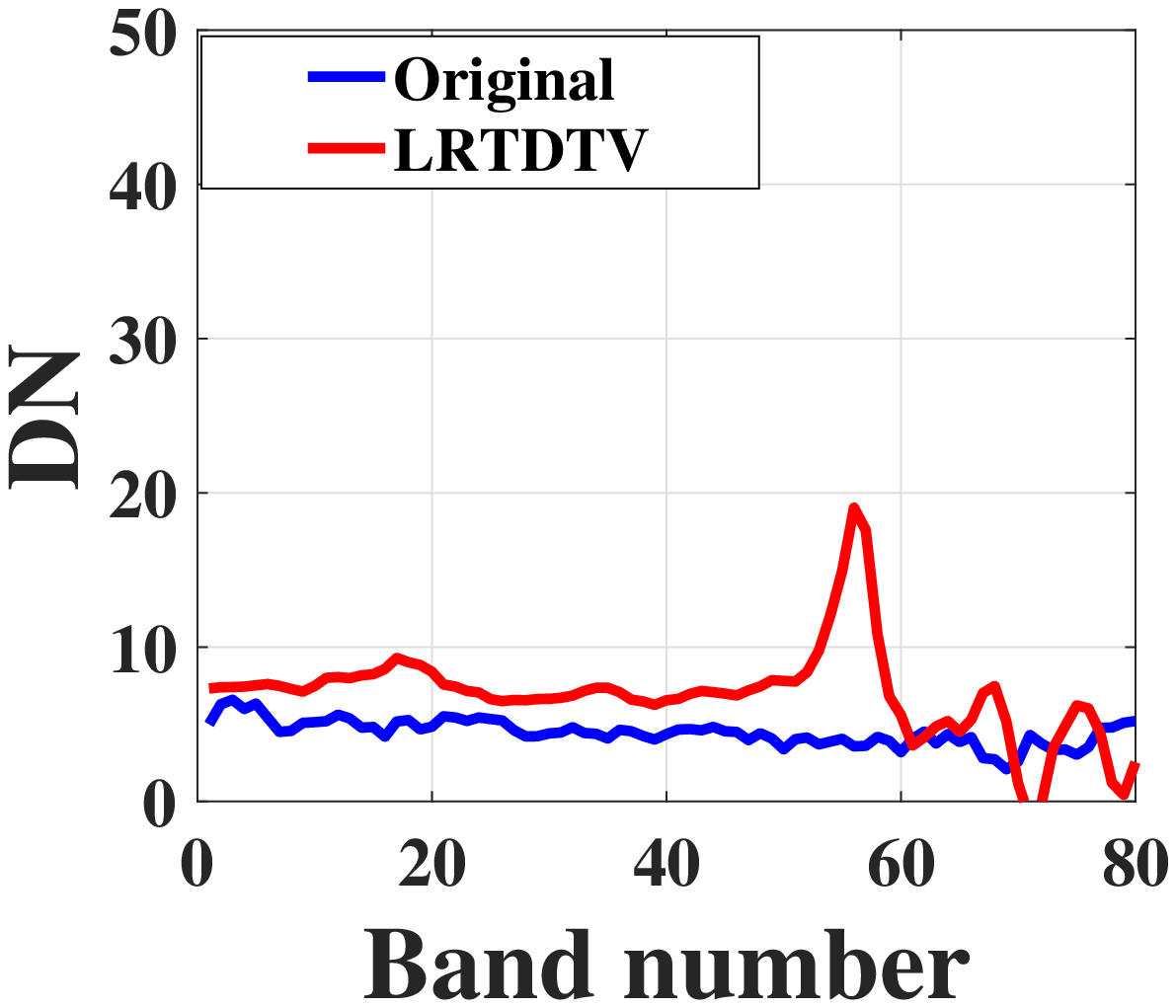}
	}\hspace{0.3mm}
	\subfloat[SSTV] {\includegraphics[width=0.35\columnwidth]{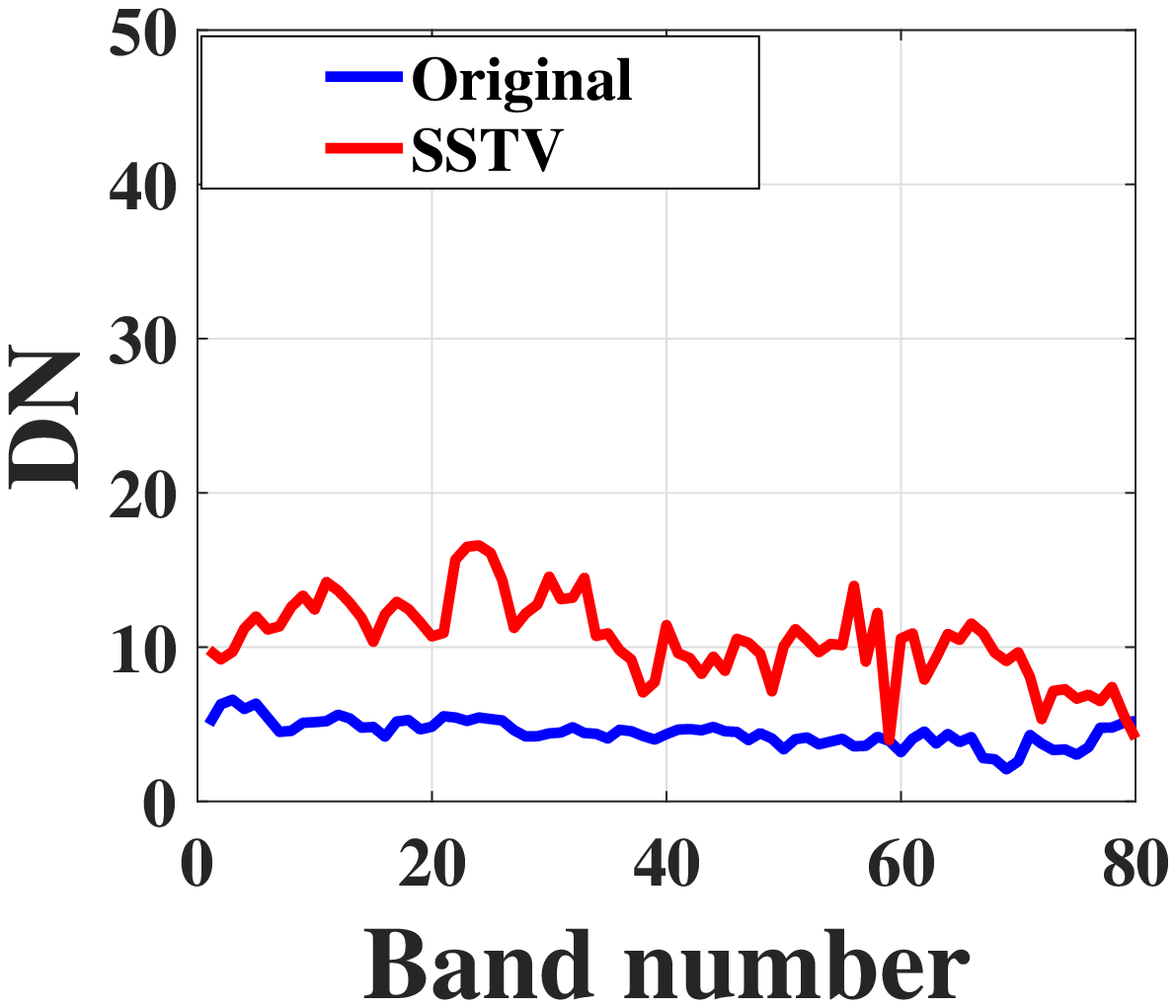}
	}\hspace{0.3mm}
	\subfloat[Our] {\includegraphics[width=0.35\columnwidth]{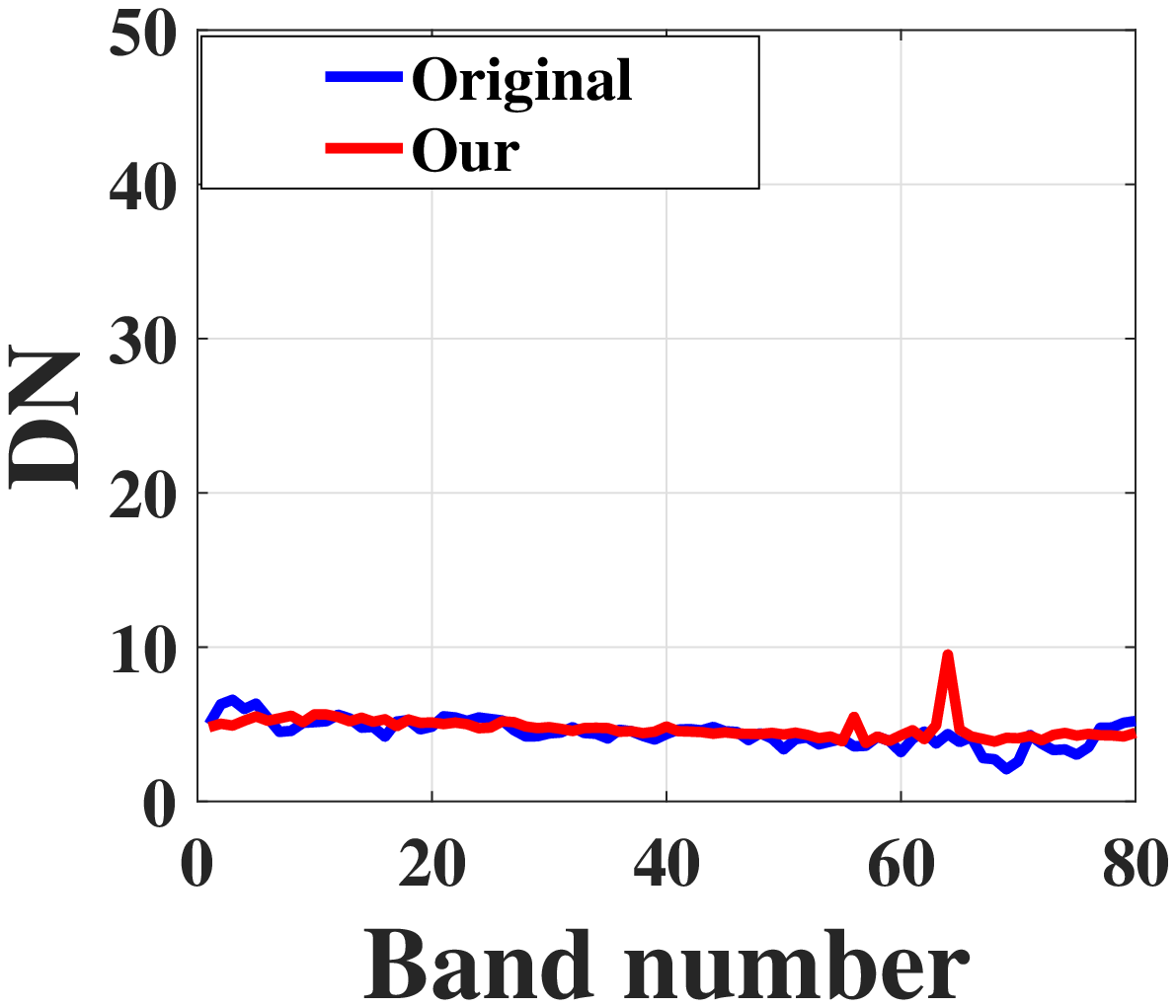}
	}\hspace{0.3mm}
	\caption{The reflectance comparison of special pixel of Pavia in (33,44) with noise Case 10.}
	\label{Pavia_imshow_pixel_3344}
\end{figure*}

To further assess and compare the performance of the tested algorithms, Fig. \ref{Pavia_imshow0202_146} and Fig. \ref{Pavia_imshow_pixel_3344} show the spectral characteristics of the pixels (21, 140) of Washington DC Mall and (33, 44) of Pavia before and after the denoising. 
Combining the ERGAS and SAM values in Table \ref{Table:WDC} and Table \ref{Table:Pavia}, one can see that among all competing methods, the HSI recovered by the proposed LRSTV method has the spectral characteristics closest to the original clean HSI.

Another interesting observation is that the proposed LRSTV based on low rank and sparseness can obtain better results than the classic SSTV based on sparseness alone and LRTDTV that combines tensor decomposition and SSTV.
This results demonstrates the effectiveness of a sparseness criterion
combining low-rank and  gradient sparsity in
HSI processing.

\begin{table*}[!t]
	\centering
	\caption{QUANTITATIVE EVALUATION RESULTS OF THE DIFFERENT DENOISING METHODS WITH THE SIMULATED NOISE IN CASES 1–11 ON Washington DC Mall DATASET (the bold one is the best one)}
	\scalebox{0.85}{
		\begin{tabular}{cccccccccccc}
			\toprule 
			Noise case & Level & PQIs   & Noisy & LRTA  & LRTV  & BM4D & NAILRMA & LRMR  & LRTDTV & SSTV & Our \\
			\midrule
			\multirow{5}[2]{*}{Case 1} & \multirow{5}[2]{*}{G=0.1} & PSNR  & 20.001 & 33.289 & 33.938 & 31.795 & 37.109 & 34.931 & 34.035 & 32.324 & \textbf{36.627} \\
			&       & SSIM  & 0.339 & 0.92  & 0.901 & 0.903 & 0.942 & 0.925 & 0.916 & 0.866 & \textbf{0.943} \\
			&       & FSIM  & 0.662 & 0.956 & 0.933 & 0.936 & 0.97  & 0.955 & 0.948 & 0.93  & \textbf{0.967} \\
			&       & ERGA  & 425.199 & 86.098 & 84.529 & 104.339 & 55.119 & 71.889 & 82.903 & 110.015 & \textbf{60.245} \\
			&       & MSAM  & 0.357 & 0.063 & 0.06  & 0.074 & \textbf{0.046} & 0.059 & 0.065 & 0.089 & 0.05 \\
			\midrule
			\multirow{5}[2]{*}{Case 2} & \multirow{5}[2]{*}{G=0.2} & PSNR  & 13.98 & 29.198 & 30.664 & 28.269 & \textbf{32.486} & 30.253 & 30.872 & 27.721 & 32.415 \\
			&       & SSIM  & 0.139 & 0.83  & 0.813 & 0.815 & \textbf{0.862} & 0.821 & 0.842 & 0.705 & 0.861 \\
			&       & FSIM  & 0.482 & 0.909 & 0.872 & 0.876 & \textbf{0.928} & 0.899 & 0.903 & 0.852 & 0.923 \\
			&       & ERGA  & 850.399 & 136.107 & 121.66 & 153.445 & \textbf{92.835} & 120.515 & 120.088 & 181.394 & 96.695 \\
			&       & MSAM  & 0.636 & 0.101 & 0.092 & 0.112 & \textbf{0.077} & 0.101 & 0.096 & 0.143 & 0.082 \\
			\midrule
			\multirow{5}[2]{*}{Case 3} & \multirow{3}[1]{*}{G=0.05} & PSNR  & 14.234 & 36.577 & 36.594 & 35.314 & 27.258 & 37.041 & 36.486 & 35.876 & \textbf{38.781} \\
			&       & SSIM  & 0.195 & 0.962 & 0.945 & 0.952 & 0.808 & 0.952 & 0.951 & 0.937 & \textbf{0.968} \\
			&       & FSIM  & 0.599 & 0.978 & 0.964 & 0.969 & 0.926 & 0.972 & 0.971 & 0.965 & \textbf{0.983} \\
			& \multirow{2}[1]{*}{P=0.1} & ERGA  & 867.536 & 60.581 & 62.391 & 71.652 & 199.283 & 59.042 & 62.301 & 76.999 & \textbf{47.085} \\
			&       & MSAM  & 0.598 & 0.042 & 0.042 & 0.049 & 0.145 & 0.047 & 0.048 & 0.063 & \textbf{0.038} \\
			\midrule
			\multirow{5}[2]{*}{Case 4} & \multirow{3}[1]{*}{G=0.075} & PSNR  & 12.387 & 33.968 & 34.186 & 32.802 & 24.127 & 33.926 & 34.318 & 33.22 & \textbf{36.404} \\
			&       & SSIM  & 0.122 & 0.931 & 0.908 & 0.923 & 0.724 & 0.913 & 0.92  & 0.893 & \textbf{0.942} \\
			&       & FSIM  & 0.514 & 0.962 & 0.938 & 0.949 & 0.893 & 0.952 & 0.95  & 0.944 & \textbf{0.968} \\
			& \multirow{2}[1]{*}{P=0.15} & ERGA  & 1070.64 & 80.096 & 82.94 & 93.325 & 297.986 & 83.734 & 80.374 & 98.23 & \textbf{61.431} \\
			&       & MSAM  & 0.689 & 0.057 & 0.057 & 0.065 & 0.211 & 0.067 & 0.063 & 0.08  & \textbf{0.051} \\
			\midrule
			\multirow{5}[2]{*}{Case 5} & \multirow{3}[1]{*}{G=0.1} & PSNR  & 11.068 & 31.926 & 32.343 & 31.014 & 21.839 & 31.583 & 32.61 & 31.138 & \textbf{34.684} \\
			&       & SSIM  & 0.085 & 0.896 & 0.87  & 0.893 & 0.647 & 0.867 & 0.887 & 0.841 & \textbf{0.915} \\
			&       & FSIM  & 0.457 & 0.944 & 0.912 & 0.929 & 0.863 & 0.927 & 0.93  & 0.919 & \textbf{0.952} \\
			& \multirow{2}[1]{*}{P=0.2} & ERGA  & 1243.266 & 100.259 & 101.336 & 113.16 & 397.968 & 108.636 & 98.449 & 121.489 & \textbf{74.45 }\\
			&       & MSAM  & 0.753 & 0.072 & 0.069 & 0.08  & 0.272 & 0.088 & 0.078 & 0.098 & \textbf{0.063} \\
			\midrule
			\multirow{5}[2]{*}{Case 6} & \multirow{2}[1]{*}{G=0.2} & PSNR  & 13.95 & 29.076 & 30.738 & 28.212 & 32.295 & 30.069 & 30.839 & 27.641 & \textbf{32.878} \\
			&       & SSIM  & 0.138 & 0.828 & 0.82  & 0.814 & 0.861 & 0.818 & 0.843 & 0.703 & \textbf{0.881} \\
			& \multirow{2}[0]{*}{dead-line=131:160} & FSIM  & 0.481 & 0.909 & 0.875 & 0.875 & 0.927 & 0.898 & 0.902 & 0.852 & \textbf{0.932} \\
			&       & ERGA  & 853.148 & 138.063 & 127.715 & 154.822 & 95.051 & 123.604 & 122.056 & 182.814 & \textbf{94.21} \\
			& stripe=111:140 & MSAM  & 0.638 & 0.101 & 0.095 & 0.112 & \textbf{0.079} & 0.103 & 0.098 & 0.144 & 0.08 \\
			\midrule
			\multirow{5}[2]{*}{Case 7} &       & PSNR  & 13.645 & 33.403 & 33.46 & 31.965 & 25.635 & 32.563 & 33.899 & 32.52 & \textbf{35.22} \\
			& G=(0,0.2) & SSIM  & 0.165 & 0.93  & 0.891 & 0.916 & 0.761 & 0.886 & 0.913 & 0.881 & \textbf{0.933} \\
			& P=(0,0.2) & FSIM  & 0.54  & 0.959 & 0.927 & 0.944 & 0.906 & 0.936 & 0.947 & 0.938 & \textbf{0.962} \\
			& dead-line=131:160 & ERGA  & 971.828 & 89.598 & 97.423 & 102.326 & 276.556 & 99.8  & 89.609 & 103.511 & \textbf{80.96} \\
			& stripe=111:140 & MSAM  & 0.652 & \textbf{0.064} & 0.075 & 0.071 & 0.197 & 0.082 & 0.073 & 0.086 & 0.069 \\
			\midrule
			\multirow{5}[2]{*}{Case 8} &       & PSNR  & 10.665 & 29.274 & 30.172 & 28.707 & 21.089 & 29.069 & 30.631 & 28.552 & \textbf{32.441} \\
			& G=0.15 & SSIM  & 0.072 & 0.837 & 0.809 & 0.838 & 0.612 & 0.796 & 0.838 & 0.751 & \textbf{0.874} \\
			& P=0.2   & FSIM  & 0.43  & 0.914 & 0.872 & 0.893 & 0.847 & 0.889 & 0.9   & 0.878 & \textbf{0.929} \\
			& dead-line=131:160 & ERGA  & 1288.726 & 134.921 & 128.901 & 146.256 & 441.014 & 143.18 & 124.861 & 160.407 & \textbf{95.866} \\
			& stripe=111:140 & MSAM  & 0.772 & 0.096 & 0.092 & 0.102 & 0.296 & 0.117 & 0.097 & 0.127 & \textbf{0.083} \\
			\midrule
			\multirow{5}[2]{*}{Case 9} &       & PSNR  & 10.212 & 27.416 & 28.794 & 27.061 & 20.242 & 27.551 & 29.255 & 26.739 & \textbf{30.815} \\
			& G=0.2 & SSIM  & 0.061 & 0.785 & 0.755 & 0.787 & 0.578 & 0.74  & 0.792 & 0.671 & \textbf{0.835} \\
			& P=0.2   & FSIM  & 0.408 & 0.887 & 0.837 & 0.858 & 0.834 & 0.861 & 0.874 & 0.84  & \textbf{0.908} \\
			& dead-line=131:160 & ERGA  & 1344.758 & 166.394 & 148.07 & 175.666 & 492.804 & 168.475 & 145.641 & 196.318 & \textbf{115.867} \\
			& stripe=111:140 & MSAM  & 0.791 & 0.116 & 0.108 & 0.121 & 0.324 & 0.139 & 0.114 & 0.152 & \textbf{0.101} \\
			\bottomrule
	\end{tabular}}
	\label{Table:WDC}
\end{table*}%

\begin{table*}[htbp]
	\centering
	\caption{QUANTITATIVE EVALUATION RESULTS OF THE DIFFERENT DENOISING METHODS WITH THE SIMULATED NOISE IN CASES 1–10 ON Pavia DATASET (the bold one is the best one)}
	\scalebox{0.85}{
		\begin{tabular}{cccccccccccc}
			\toprule
			Noise case & Level & PQIs   & Noisy & LRTA  & LRTV  & BM4D  & NAILRMA & LRMR  & LRTDTV & SSTV  & Our \\
			\midrule
			\multirow{5}[2]{*}{Case 1} & \multirow{5}[2]{*}{G=0.1} & PSNR  & 20.001 & 31.02 & 33.234 & 30.99 & \textbf{35.19}8 & 33.748 & 33.26 & 33.434 & 35.187 \\
			&       & SSIM  & 0.436 & 0.929 & 0.916 & 0.932 & \textbf{0.952} & 0.942 & 0.926 & 0.927 & 0.951 \\
			&       & FSIM  & 0.722 & 0.958 & 0.949 & 0.956 & \textbf{0.972} & 0.962 & 0.953 & 0.957 & 0.968 \\
			&       & ERGA  & 368.912 & 104.831 & 89.259 & 104.88 & 65.422 & 78.364 & 80.014 & 80.014 & \textbf{64.881} \\
			&       & MSAM  & 0.548 & 0.127 & 0.128 & 0.098 & 0.107 & 0.124 & 0.109 & 0.146 & \textbf{0.105} \\
			\midrule
			\multirow{5}[2]{*}{Case 2} & \multirow{5}[2]{*}{G=0.2} & PSNR  & 13.98 & 26.841 & 29.133 & 27.07 & 30.41 & 29.192 & 29.554 & 28.007 & \textbf{30.87} \\
			&       & SSIM  & 0.182 & 0.838 & 0.83  & 0.857 & \textbf{0.879} & 0.854 & 0.841 & 0.801 & 0.876 \\
			&       & FSIM  & 0.546 & 0.914 & 0.896 & 0.91  & 0.93  & 0.916 & 0.903 & 0.892 & \textbf{0.923} \\
			&       & ERGA  & 737.824 & 167.054 & 144.929 & 162.526 & 111.83 & 128.298 & 122.327 & 149.419 & \textbf{107.545} \\
			&       & MSAM  & 0.828 & 0.183 & 0.207 & 0.125 & 0.161 & 0.191 & 0.152 & 0.239 & \textbf{0.168} \\
			\midrule
			\multirow{5}[2]{*}{Case 3} & \multirow{3}[1]{*}{G=0.05} & PSNR  & 14.29 & 34.101 & 36.289 & 34.096 & 26.891 & 36.385 & 36.128 & \textbf{37.771} & 37.352 \\
			&       & SSIM  & 0.247 & 0.965 & 0.962 & 0.967 & 0.846 & 0.967 & 0.962 & 0.971 & \textbf{0.973} \\
			&       & FSIM  & 0.665 & 0.978 & 0.975 & 0.978 & 0.925 & 0.979 & 0.976 & \textbf{0.984} & 0.983 \\
			& \multirow{2}[1]{*}{P=0.1} & ERGA  & 717.692 & 75.447 & 59.183 & 75.13 & 168.129 & 57.436 & 57.449 & \textbf{48.403} & 50.018 \\
			&       & MSAM  & 0.704 & 0.088 & 0.094 & 0.077 & 0.13  & 0.084 & 0.083 & 0.077 & \textbf{0.072} \\
			\midrule
			\multirow{5}[2]{*}{Case 4} & \multirow{3}[1]{*}{G=0.075} & PSNR  & 12.412 & 31.441 & 33.686 & 31.531 & 23.947 & 33.339 & 33.752 & 34.403 & \textbf{35.274} \\
			&       & SSIM  & 0.154 & 0.938 & 0.931 & 0.944 & 0.773 & 0.937 & 0.936 & 0.94  & \textbf{0.955} \\
			&       & FSIM  & 0.575 & 0.963 & 0.956 & 0.963 & 0.898 & 0.962 & 0.958 & 0.969 & \textbf{0.972} \\
			& \multirow{2}[1]{*}{P=0.15} & ERGA  & 890.259 & 101.073 & 82.052 & 99.823 & 236.001 & 80.798 & 76.2  & 71.321 & \textbf{64.003} \\
			&       & MSAM  & 0.77  & 0.111 & 0.126 & 0.093 & 0.147 & 0.105 & 0.11  & 0.101 & \textbf{0.089} \\
			\midrule
			\multirow{5}[2]{*}{Case 5} & \multirow{3}[1]{*}{G=0.1} & PSNR  & 11.082 & 29.404 & 31.57 & 29.628 & 21.724 & 31.166 & 31.67 & 31.786 & \textbf{33.41} \\
			&       & SSIM  & 0.105 & 0.905 & 0.896 & 0.919 & 0.701 & 0.902 & 0.902 & 0.898 & \textbf{0.93} \\
			&       & FSIM  & 0.514 & 0.947 & 0.935 & 0.947 & 0.872 & 0.944 & 0.937 & 0.95  & \textbf{0.957} \\
			& \multirow{2}[1]{*}{P=0.2} & ERGA  & 1037.013 & 126.62 & 105.887 & 123.37 & 305.688 & 102.759 & 96.904 & 96.406 & \textbf{79.551} \\
			&       & MSAM  & 0.809 & 0.128 & 0.156 & 0.106 & 0.16  & 0.121 & 0.134 & 0.125 & \textbf{0.105} \\
			\midrule
			\multirow{5}[2]{*}{Case 6} & \multirow{2}[1]{*}{G=0.2} & PSNR  & 13.91 & 26.521 & 28.655 & 26.729 & 30.193 & 28.965 & 29.295 & 27.792 & \textbf{30.553} \\
			&       & SSIM  & 0.178 & 0.834 & 0.823 & 0.854 & 0.877 & 0.851 & 0.836 & 0.797 & \textbf{0.884} \\
			& dead-line=54:74 & FSIM  & 0.543 & 0.913 & 0.89  & 0.909 & \textbf{0.929} & 0.915 & 0.9   & 0.891 & 0.927 \\
			& stripe=54:64 & ERGA  & 742.599 & 174.279 & 153.902 & 169.729 & 114.464 & 131.541 & 127.949 & 152.421 & \textbf{114.408} \\
			&       & MSAM  & 0.837 & 0.187 & 0.232 &\textbf{ 0.126} & 0.165 & 0.198 & 0.17  & 0.248 & 0.18 \\
			\midrule
			\multirow{5}[2]{*}{Case 7} &       & PSNR  & 11.62 & 27.349 & 29.594 & 27.748 & 22.5  & 29.369 & 29.864 & 29.395 & \textbf{31.05} \\
			& G=(0.1,0.2) & SSIM  & 0.115 & 0.865 & 0.843 & 0.894 & 0.716 & 0.863 & 0.856 & 0.843 & \textbf{0.897} \\
			& P=(0.1,0.2) & FSIM  & 0.517 & 0.93  & 0.905 & 0.934 & 0.873 & 0.925 & 0.912 & 0.924 & \textbf{0.937} \\
			& dead-line=54:74 & ERGA  & 982.617 & 159.906 & 150.144 & 152.46 & 284.596 & 125.431 & 119.332 & 127.077 & \textbf{115.822} \\
			& stripe=54:64 & MSAM  & 0.806 & 0.145 & 0.222 & \textbf{0.118} & 0.172 & 0.139 & 0.143 & 0.157 & 0.165 \\
			\midrule
			\multirow{5}[2]{*}{Case 8} &       & PSNR  & 10.666 & 26.684 & 28.908 & 27.002 & 21.082 & 28.683 & 29.042 & 28.51 & \textbf{30.78} \\
			& G=0.15 & SSIM  & 0.087 & 0.844 & 0.83  & 0.872 & 0.663 & 0.843 & 0.837 & 0.818 & \textbf{0.882} \\
			& P=0.2   & FSIM  & 0.485 & 0.922 & 0.896 & 0.92  & 0.856 & 0.917 & 0.899 & 0.913 & \textbf{0.931} \\
			& dead-line=54:74 & ERGA  & 1085.555 & 172.021 & 147.64 & 165.688 & 331.455 & 135.095 & 131.268 & 139.649 & \textbf{109.451} \\
			& stripe=54:64 & MSAM  & 0.827 & 0.146 & 0.205 & \textbf{0.12}  & 0.173 & 0.144 & 0.164 & 0.169 & 0.142 \\
			\midrule
			\multirow{5}[2]{*}{Case 9} &       & PSNR  & 10.227 & 24.819 & 27.281 & 25.185 & 20.202 & 27.053 & 27.702 & 26.369 & \textbf{29.13} \\
			& G=0.2 & SSIM  & 0.073 & 0.784 & 0.774 & 0.824 & 0.626 & 0.784 & 0.789 & 0.74  & \textbf{0.849} \\
			& P=0.2   & FSIM  & 0.463 & 0.899 & 0.868 & 0.892 & 0.842 & 0.893 & 0.873 & 0.878 & \textbf{0.911} \\
			& dead-line=54:74 & ERGA  & 1140.084 & 211.928 & 167.908 & 202.9 & 367.514 & 162.357 & 151.926 & 178.448 & \textbf{134.873} \\
			& stripe=54:64 & MSAM  & 0.835 & 0.156 & 0.215 &\textbf{ 0.129} & 0.179 & 0.158 & 0.175 & 0.202 & 0.173 \\
			\bottomrule
	\end{tabular}}
	\label{Table:Pavia}
\end{table*}%

\begin{figure*}[!t] \centering
	\subfloat[Original image] {\includegraphics[width=0.32\columnwidth]{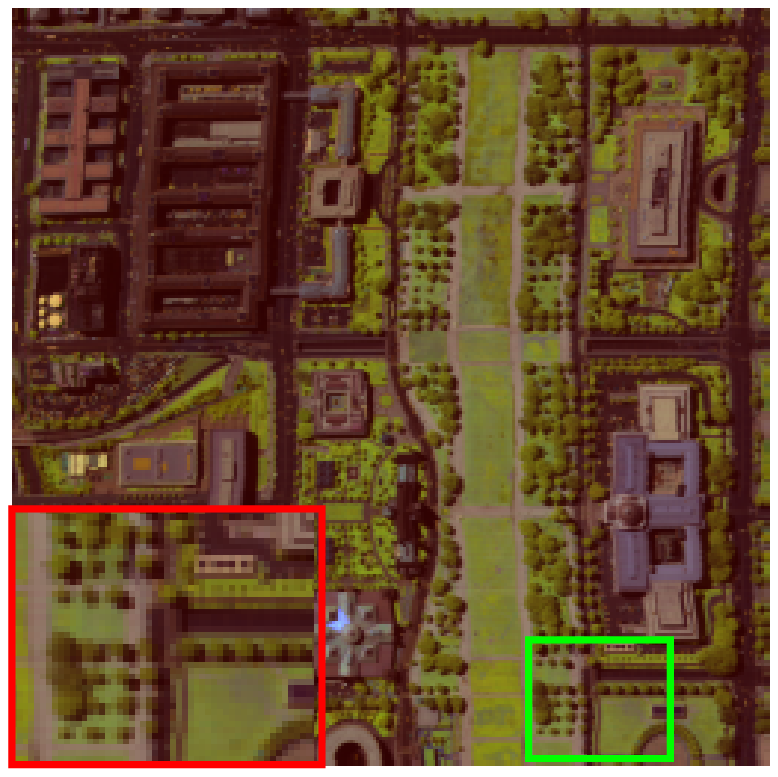}
	}\hspace{0.1mm}
	\subfloat[Noisy one(10.91dB)] {\includegraphics[width=0.32\columnwidth]{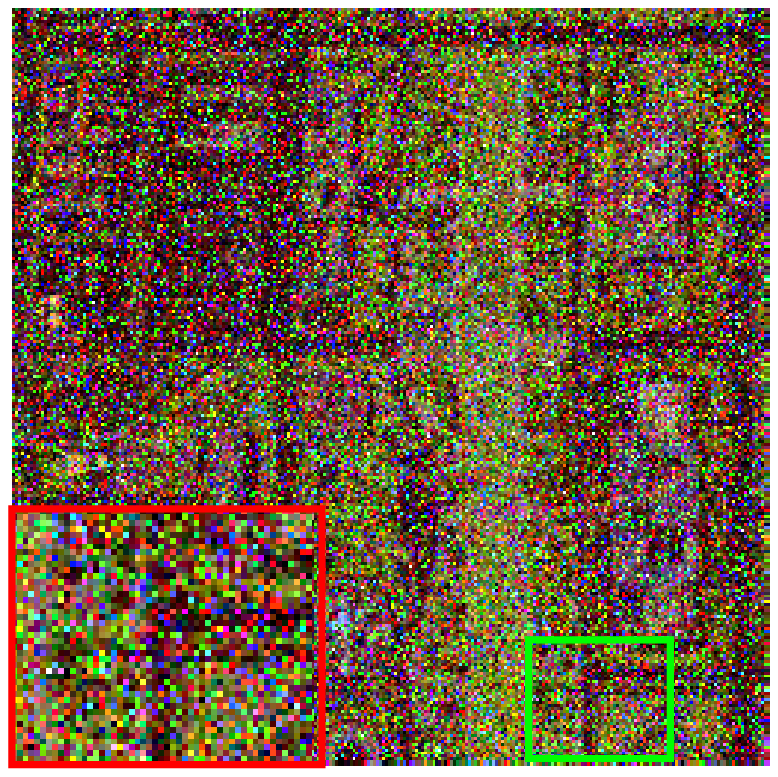}
	}\hspace{0.1mm}
	\subfloat[LRTA(28.07dB)] {\includegraphics[width=0.32\columnwidth]{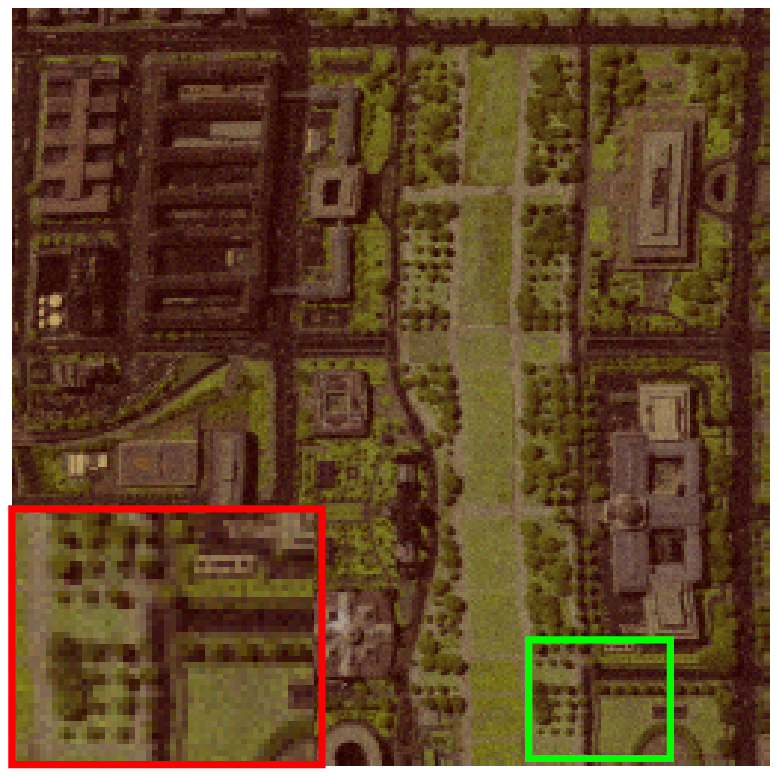}
	}\hspace{0.1mm}
	\subfloat[LRTV(28.35dB)] {\includegraphics[width=0.32\columnwidth]{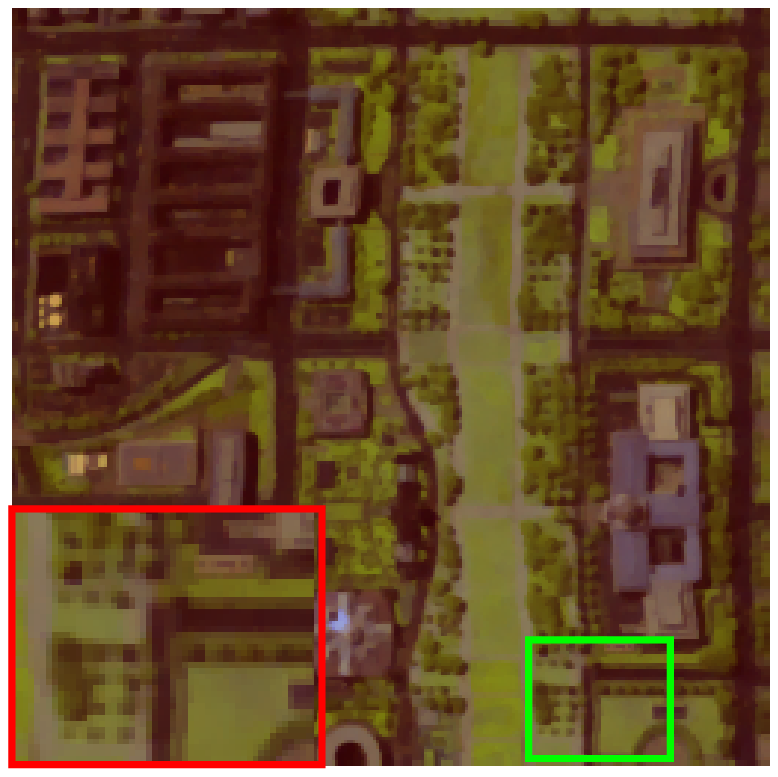}
	}\hspace{0.1mm}
	\subfloat[BM4D(26.41dB)] {\includegraphics[width=0.32\columnwidth]{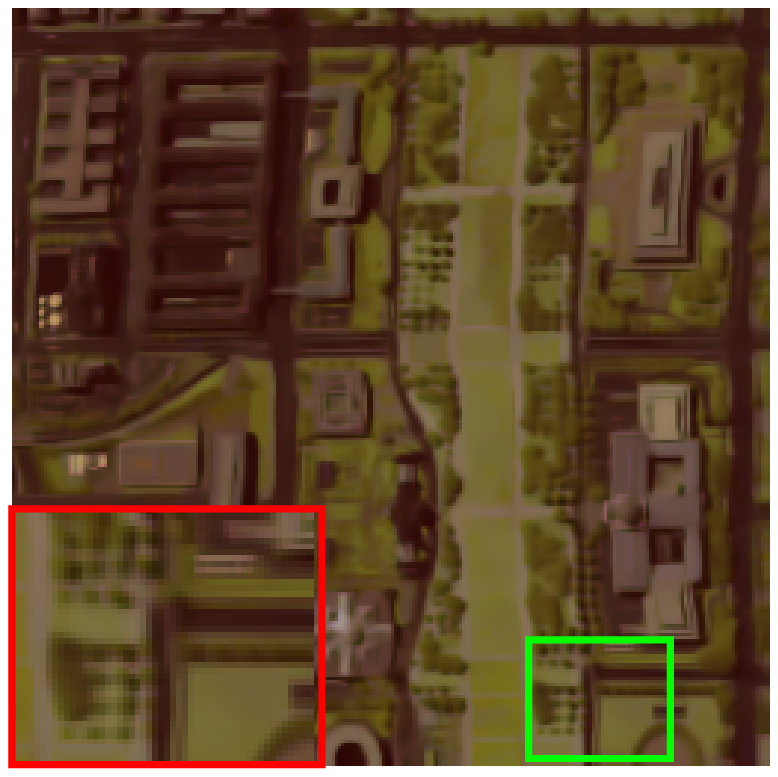}
	}\hspace{0.1mm}
	\subfloat[NALRMA(22.29dB)] {\includegraphics[width=0.32\columnwidth]{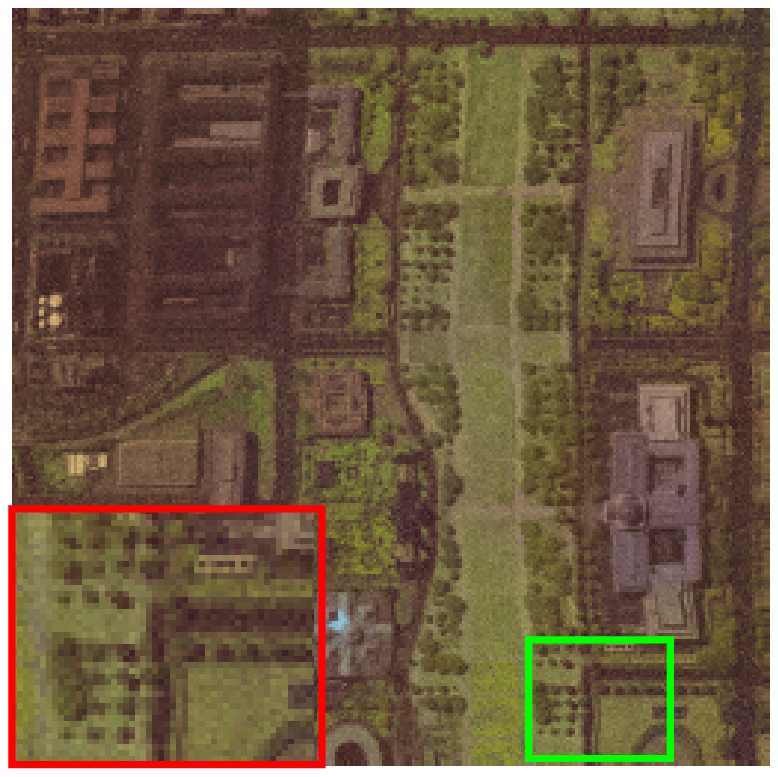}
	}\hspace{0.1mm}
	\subfloat[LRMR(28.87dB)] {\includegraphics[width=0.32\columnwidth]{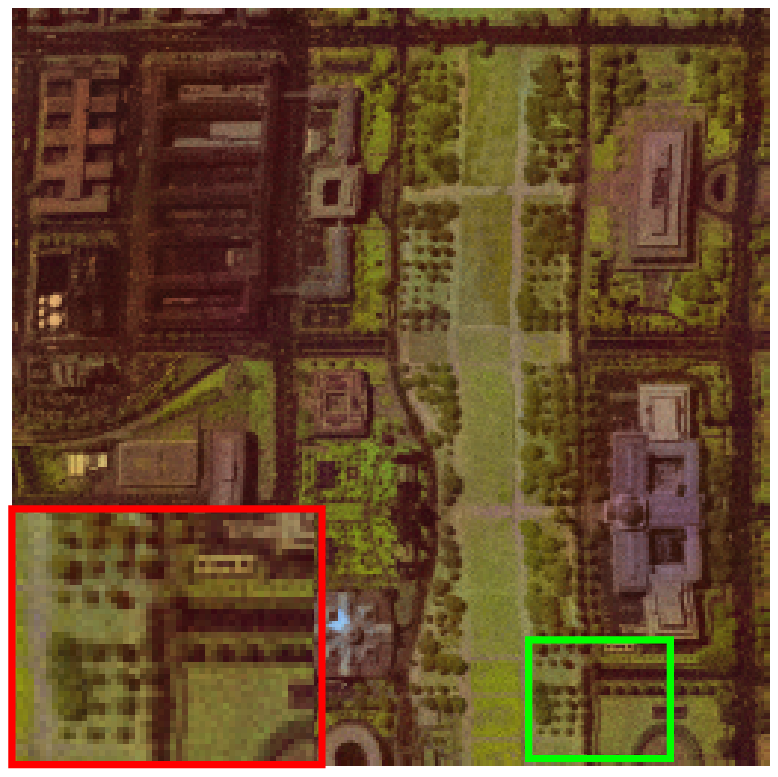}
	}\hspace{0.1mm}
	\subfloat[LRTDTV(28.78dB)] {\includegraphics[width=0.32\columnwidth]{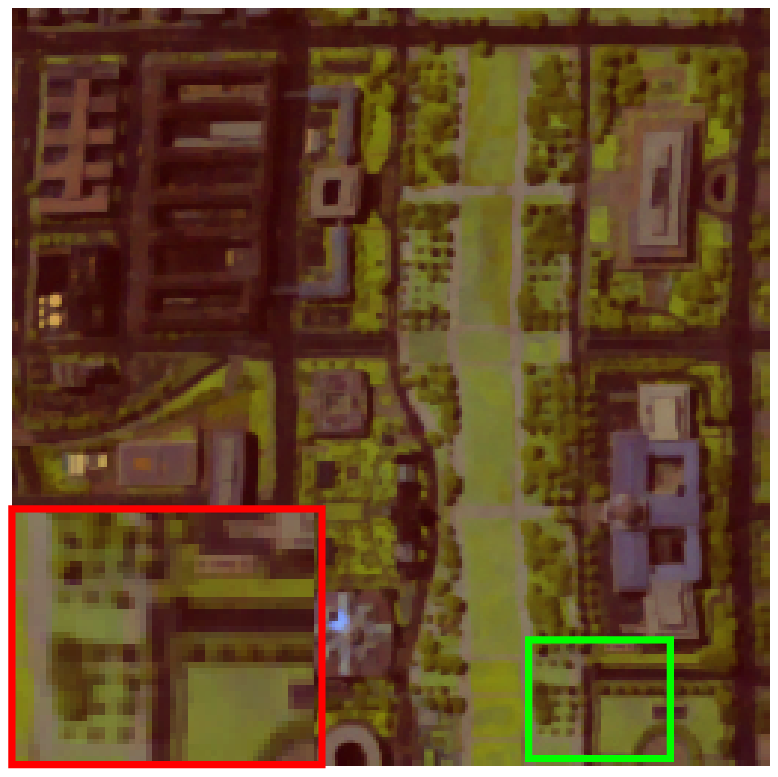}
	}\hspace{0.1mm}
	\subfloat[SSTV(29.05dB)] {\includegraphics[width=0.32\columnwidth]{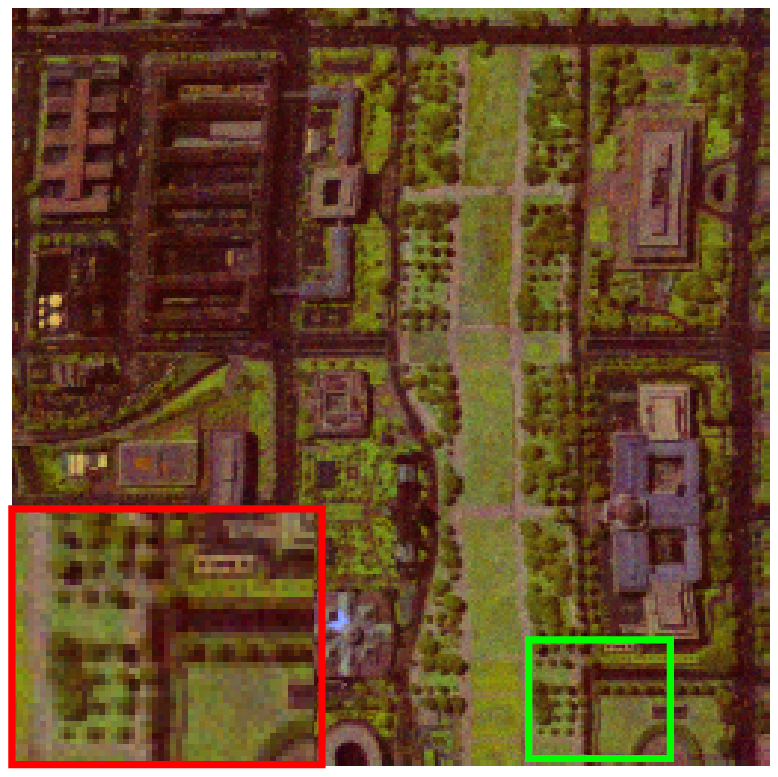}
	}\hspace{0.1mm}
	\subfloat[Our(31.49dB)] {\includegraphics[width=0.32\columnwidth]{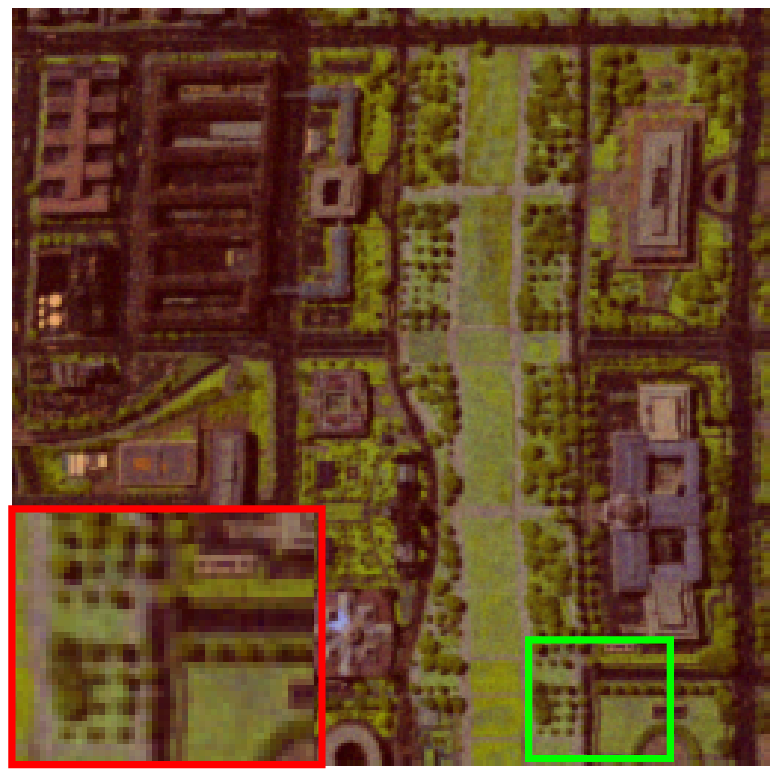}
	}\hspace{0.1mm}
	\caption{Recovery performance comparison on the Washington DC Mall data by LRTA, LRTV, BM4D, NAILRMA, LRMR, LRTDTV, SSTV and the proposed TDLRSTV. The noise level: Gaussian noise (with 0 mean and 0.15 variance) and sparse pepper-and-salt noise (with percentage 0.2) are added for all bands; dead lines are added to the 131-160 band, the number of dead lines varies randomly from 3 to 10, and the width of dead lines varies randomly from 1 to 3; stripes are added to the 111-140 band, where the number of bands varies randomly between 20 and 40. The color image is composed of bands 58 85 155 for the red, green, and blue channels, respectively.}
	\label{DC_imshow0202_146}
\end{figure*}

\subsection{Real HSI Data Experiments}
	
\begin{figure*}[htbp] \centering
	\captionsetup[subfloat]{labelsep=none,format=plain,labelformat=empty}	
	\subfloat[Noisy image] {\includegraphics[width=0.35\columnwidth]{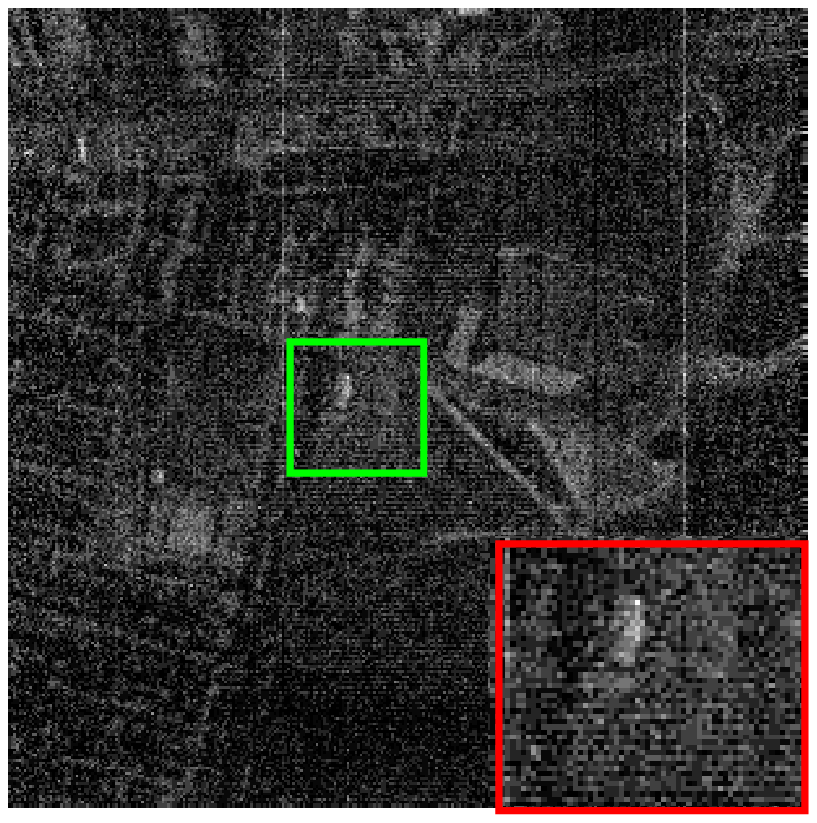}
	}\hspace{0.1mm}
	\subfloat[LRTA] {\includegraphics[width=0.35\columnwidth]{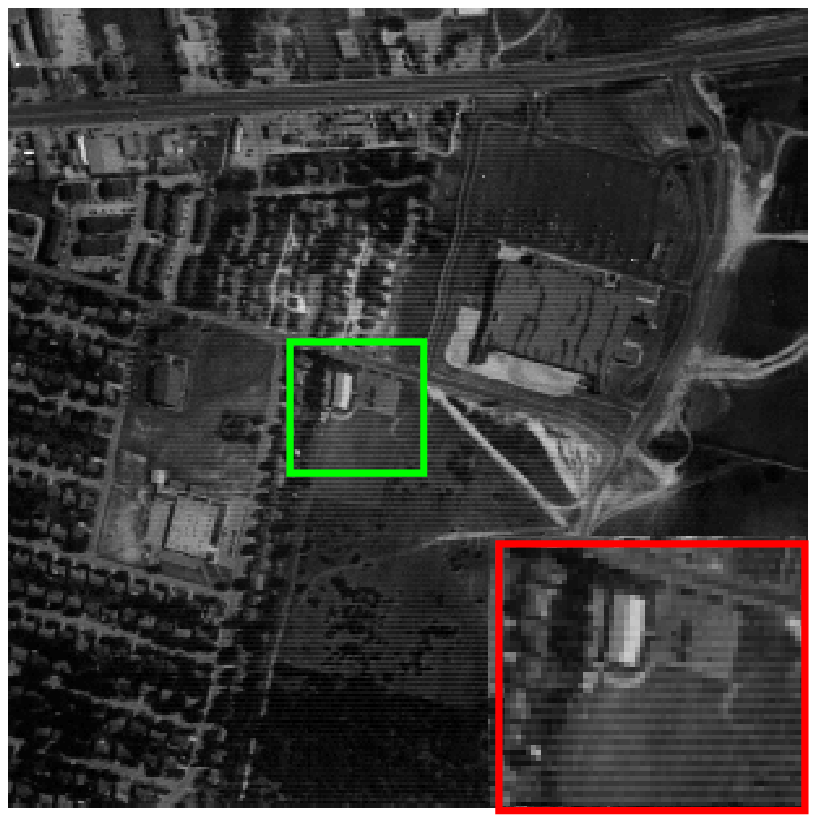}
	}\hspace{0.1mm}
	\subfloat[LRTV] {\includegraphics[width=0.35\columnwidth]{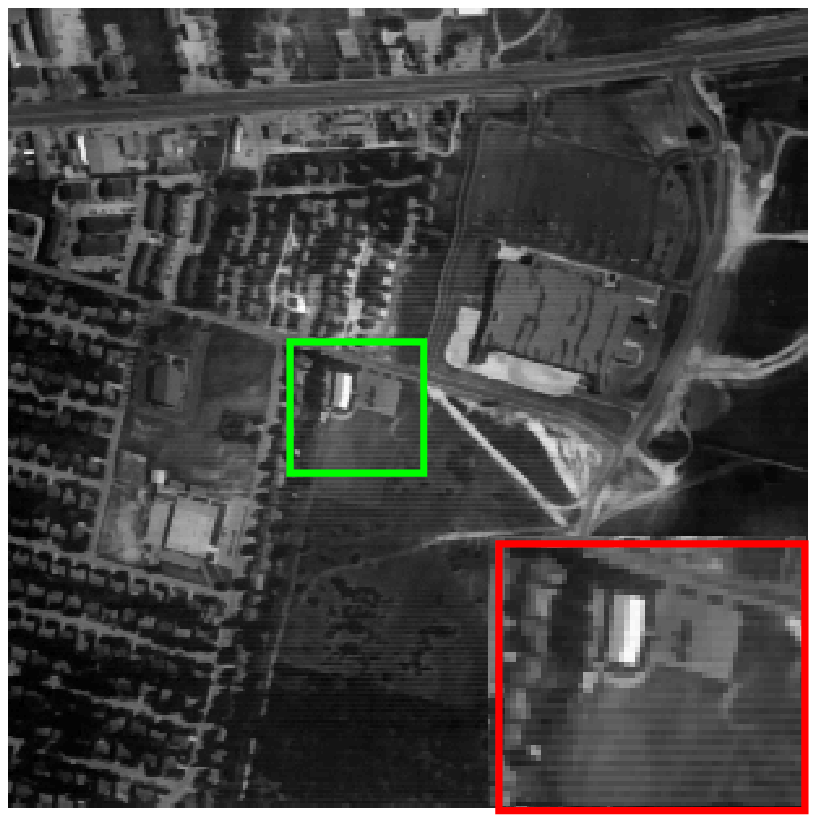}
	}\hspace{0.1mm}
	\subfloat[BM4D] {\includegraphics[width=0.35\columnwidth]{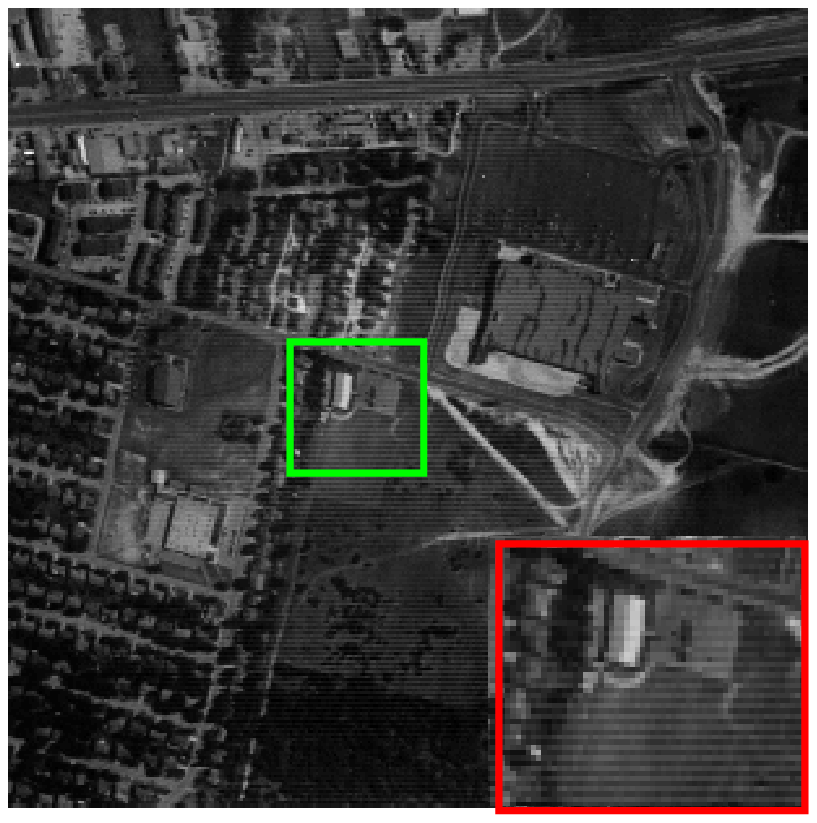}
	}\hspace{0.1mm}
	\subfloat[LRMR] {\includegraphics[width=0.35\columnwidth]{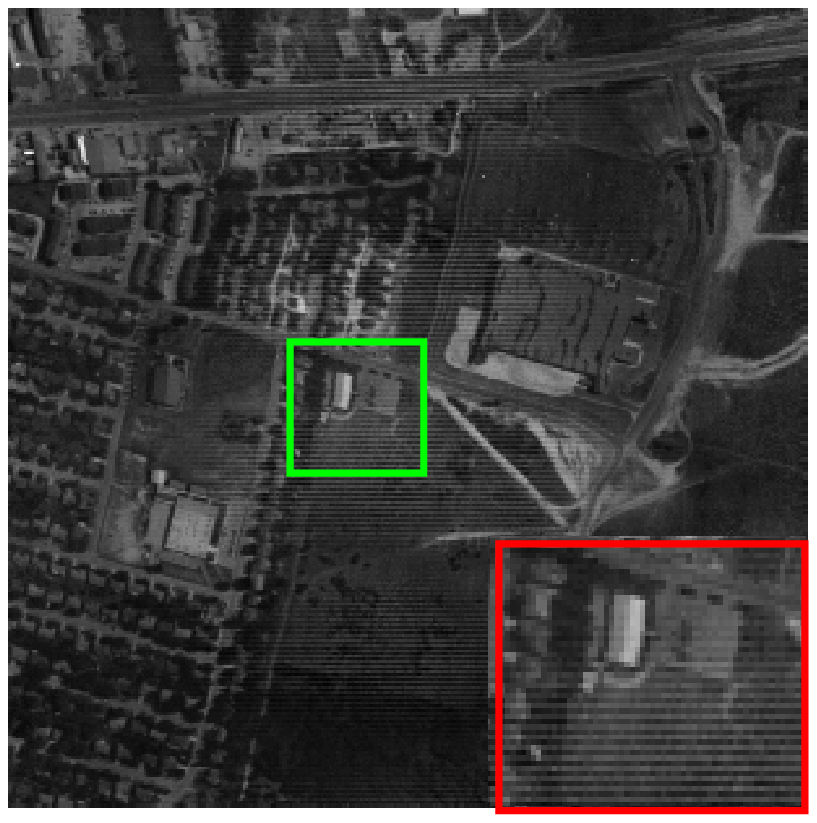}
	}\hspace{0.1mm}
	\subfloat[LRTDTV] {\includegraphics[width=0.35\columnwidth]{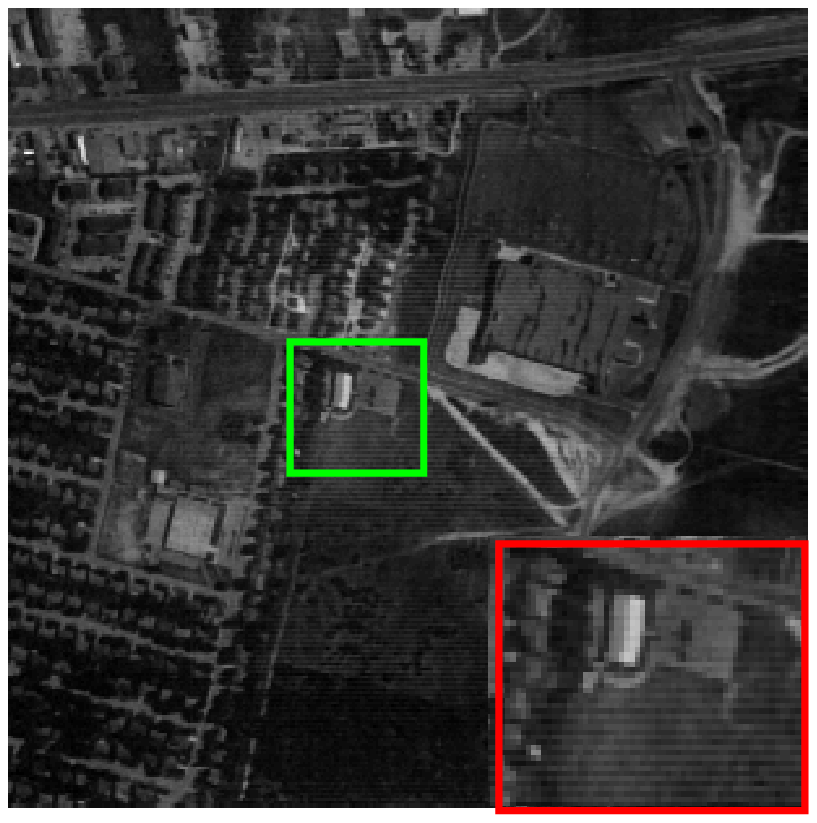}
	}\hspace{0.1mm}
	\subfloat[SSTV] {\includegraphics[width=0.35\columnwidth]{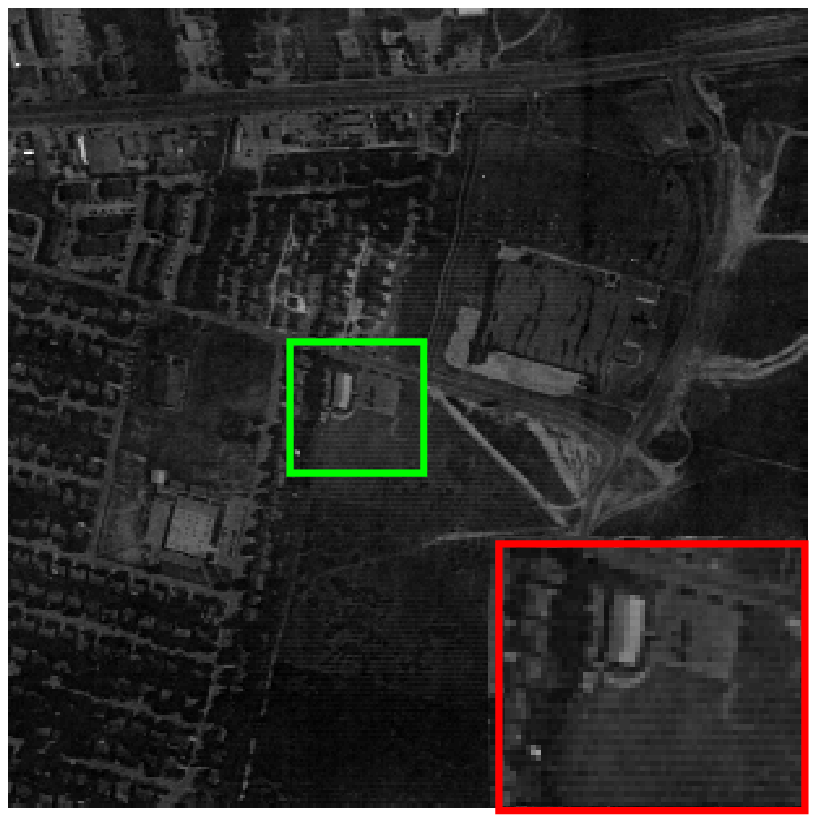}
	}\hspace{0.1mm}
	\subfloat[Our] {\includegraphics[width=0.35\columnwidth]{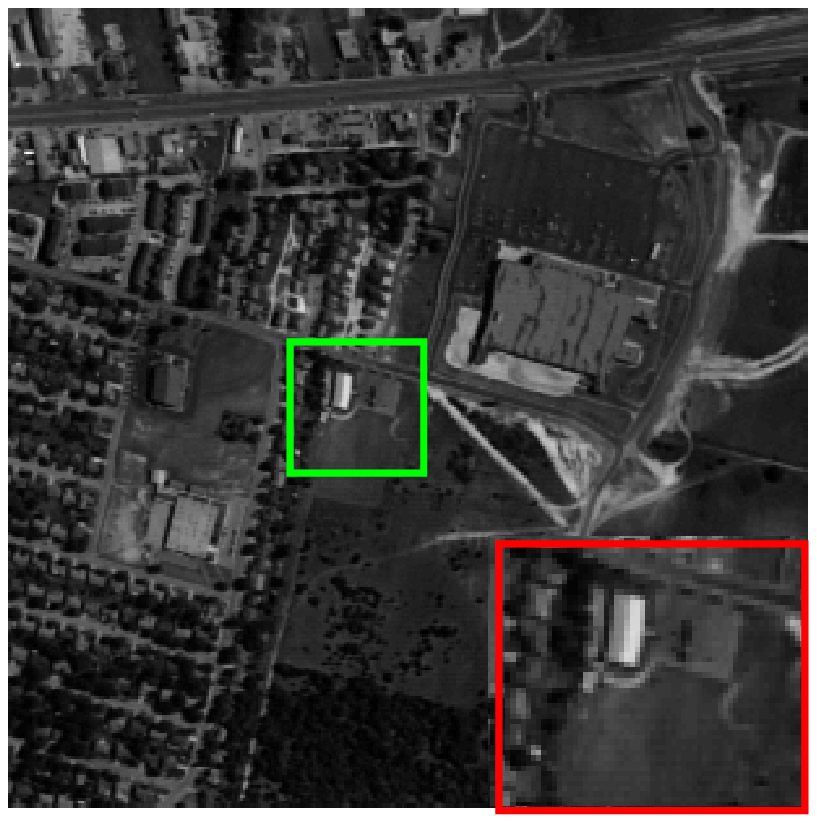}
	}\hspace{0.1mm}
	\caption{Recovery performance comparison on the HYDICE Urban data. The noise is real noise, and the original noisy image in (a) is band 207. }
	\label{real_imshowUrban}
\end{figure*}

\begin{figure*}[htbp] \centering
	\captionsetup[subfloat]{labelsep=none,format=plain,labelformat=empty} 
	\subfloat[Noisy image] {\includegraphics[width=0.32\columnwidth]{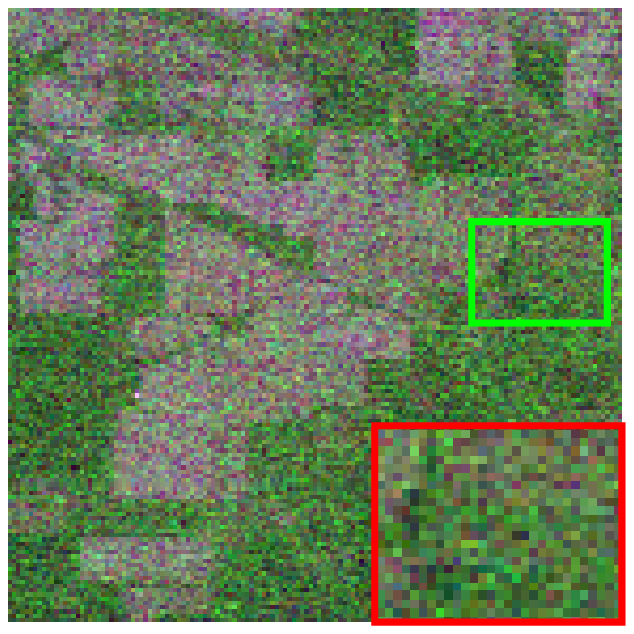}
	}\hspace{0.1mm}
	\subfloat[LRTA] {\includegraphics[width=0.32\columnwidth]{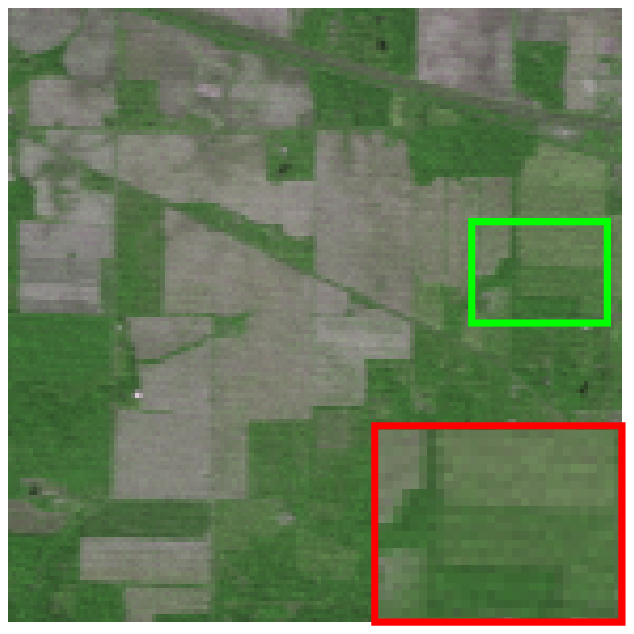}
	}\hspace{0.1mm}
	\subfloat[LRTV] {\includegraphics[width=0.32\columnwidth]{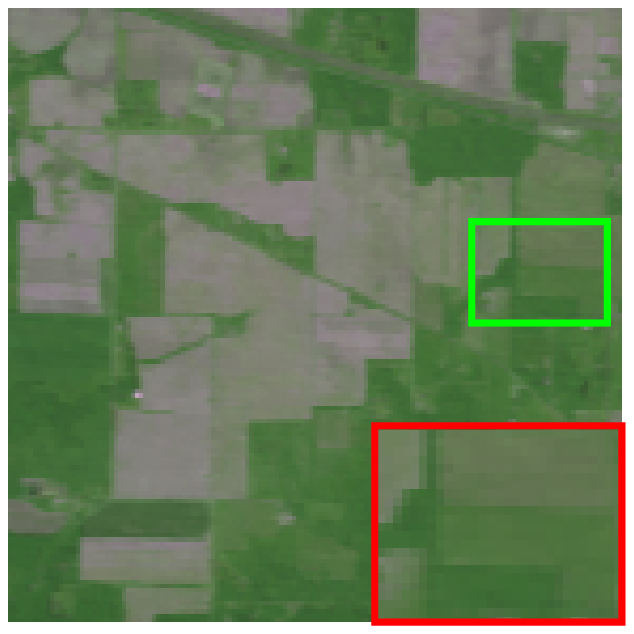}
	}\hspace{0.1mm}
	\subfloat[BM4D] {\includegraphics[width=0.32\columnwidth]{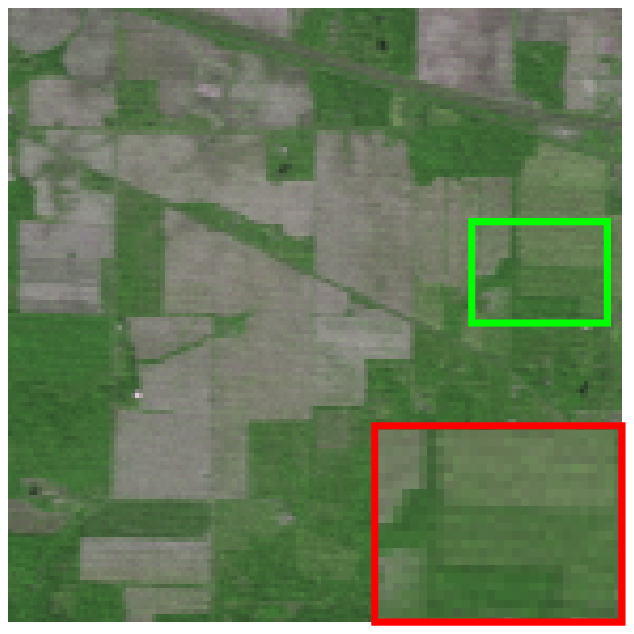}
	}\hspace{0.1mm}
	\subfloat[LRMR] {\includegraphics[width=0.32\columnwidth]{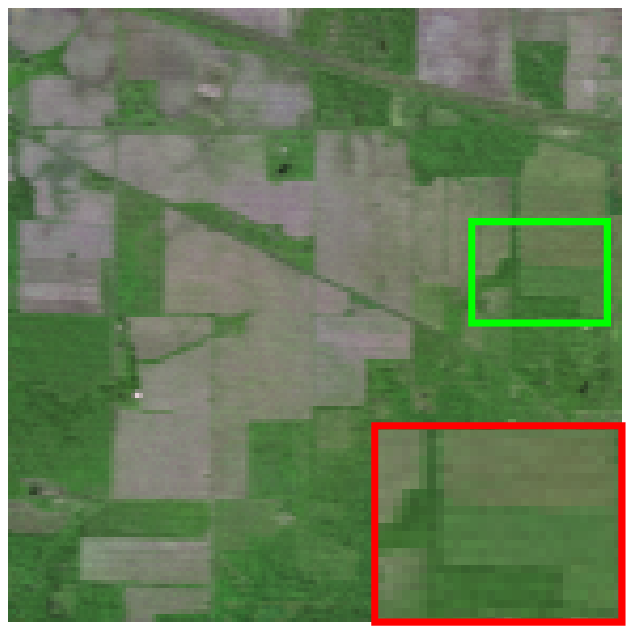}
	}\hspace{0.1mm}
	\subfloat[LRTDTV] {\includegraphics[width=0.32\columnwidth]{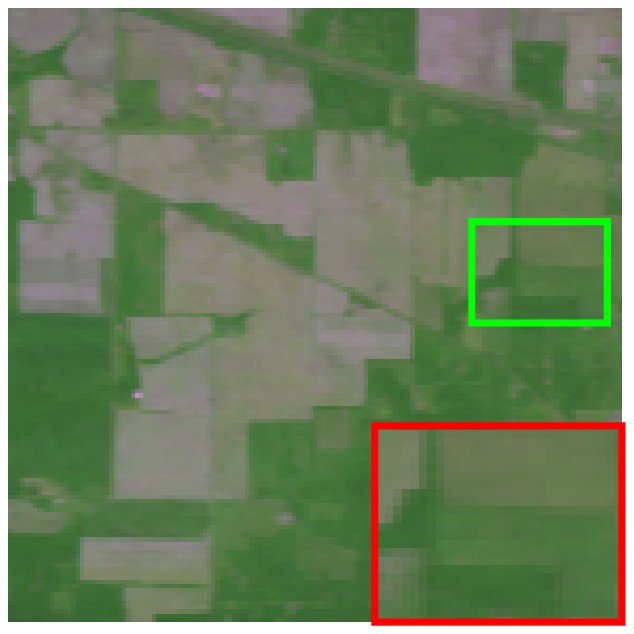}
	}\hspace{0.1mm}
	\subfloat[SSTV] {\includegraphics[width=0.32\columnwidth]{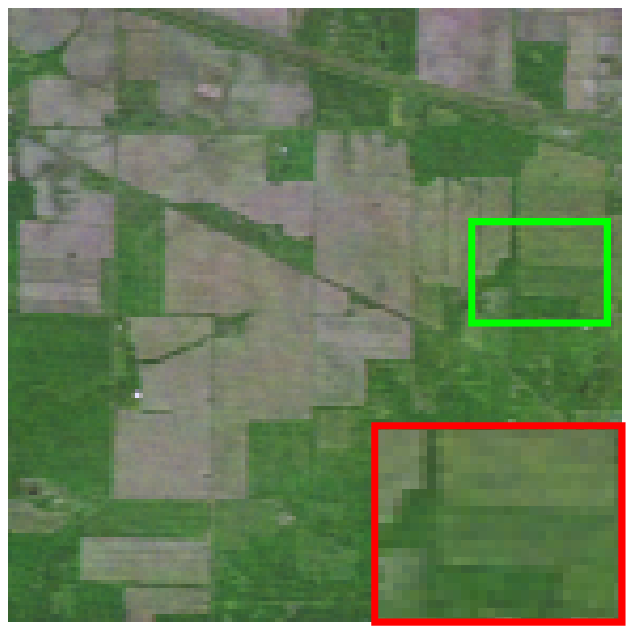}
	}\hspace{0.1mm}
	\subfloat[Our] {\includegraphics[width=0.32\columnwidth]{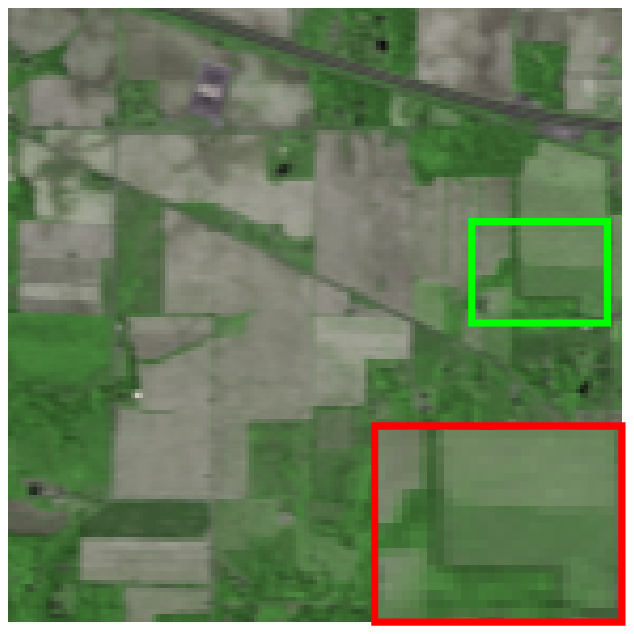}
	}
	\caption{Recovery performance comparison on the AVIRIS Indian Pines data. The noise is real noise, the color image (R: 164, G: 150, B: 218).
	}
	\label{real_imshowIndian}
\end{figure*}

\begin{figure*}[htbp] \centering
	\captionsetup[subfloat]{labelsep=none,format=plain,labelformat=empty} 
	\subfloat[Noisy image] {\includegraphics[width=0.35\columnwidth]{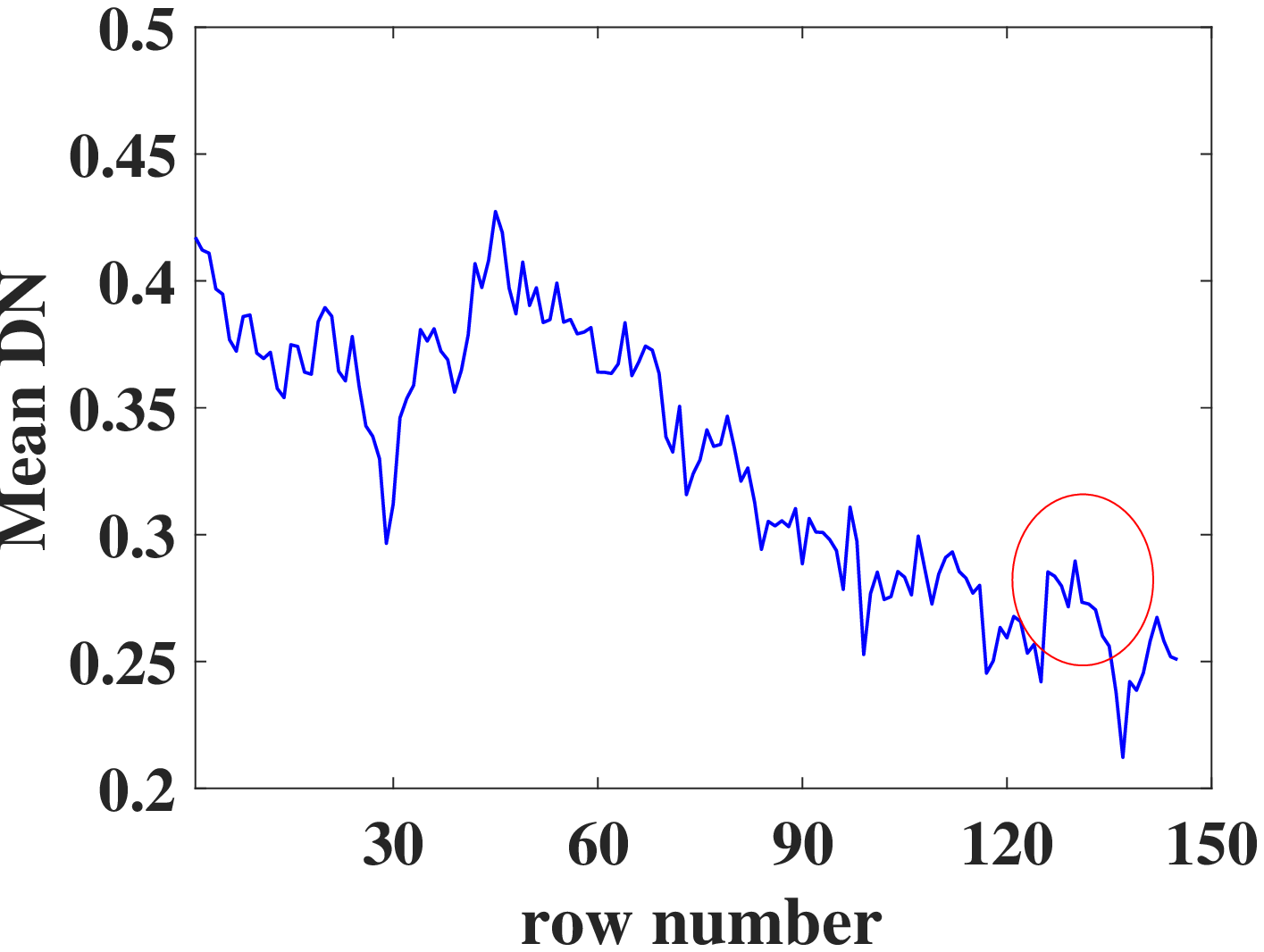}
	}\hspace{0.1mm}
	\subfloat[LRTA] {\includegraphics[width=0.35\columnwidth]{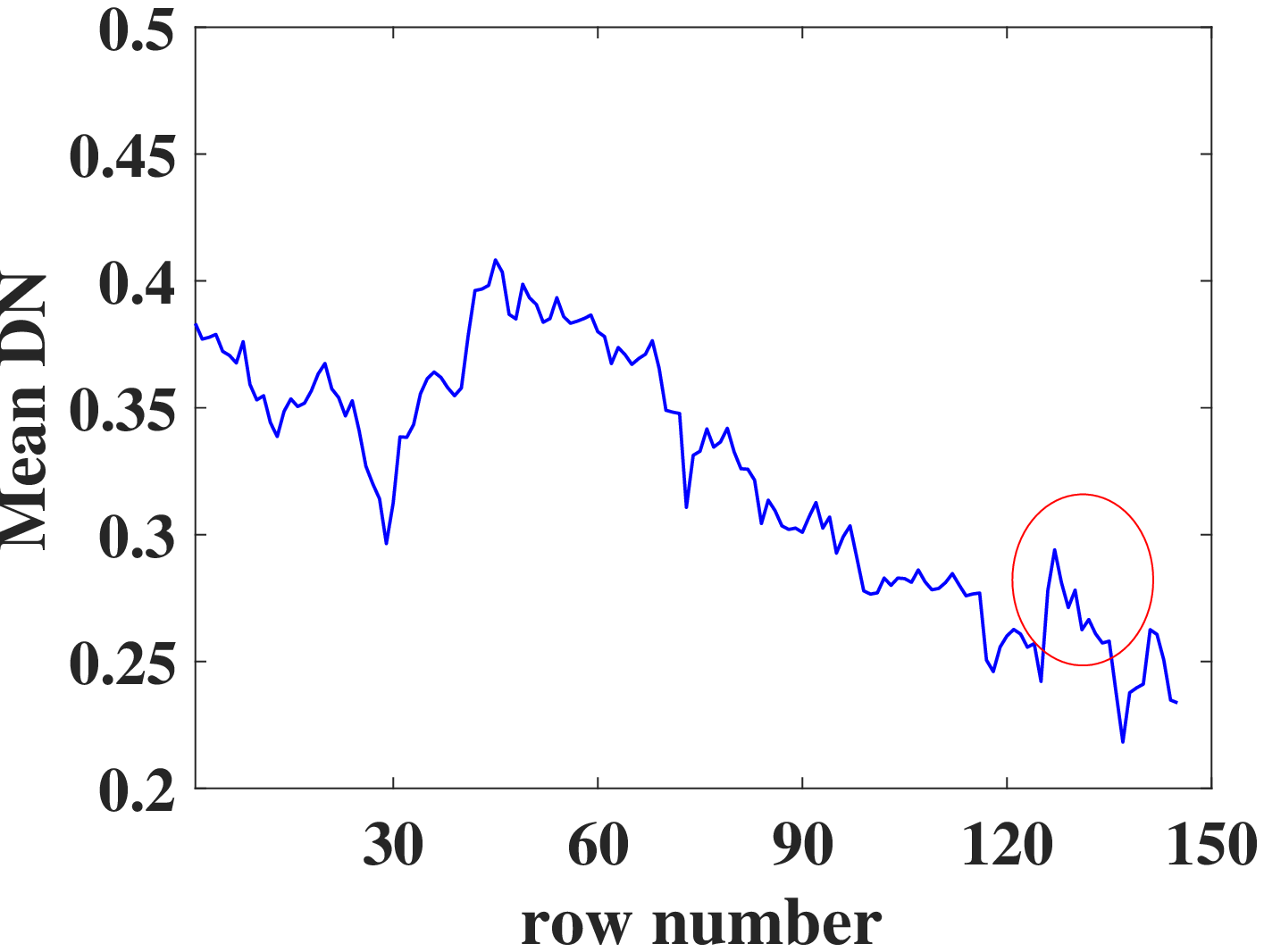}
	}\hspace{0.1mm}
	\subfloat[LRTV] {\includegraphics[width=0.35\columnwidth]{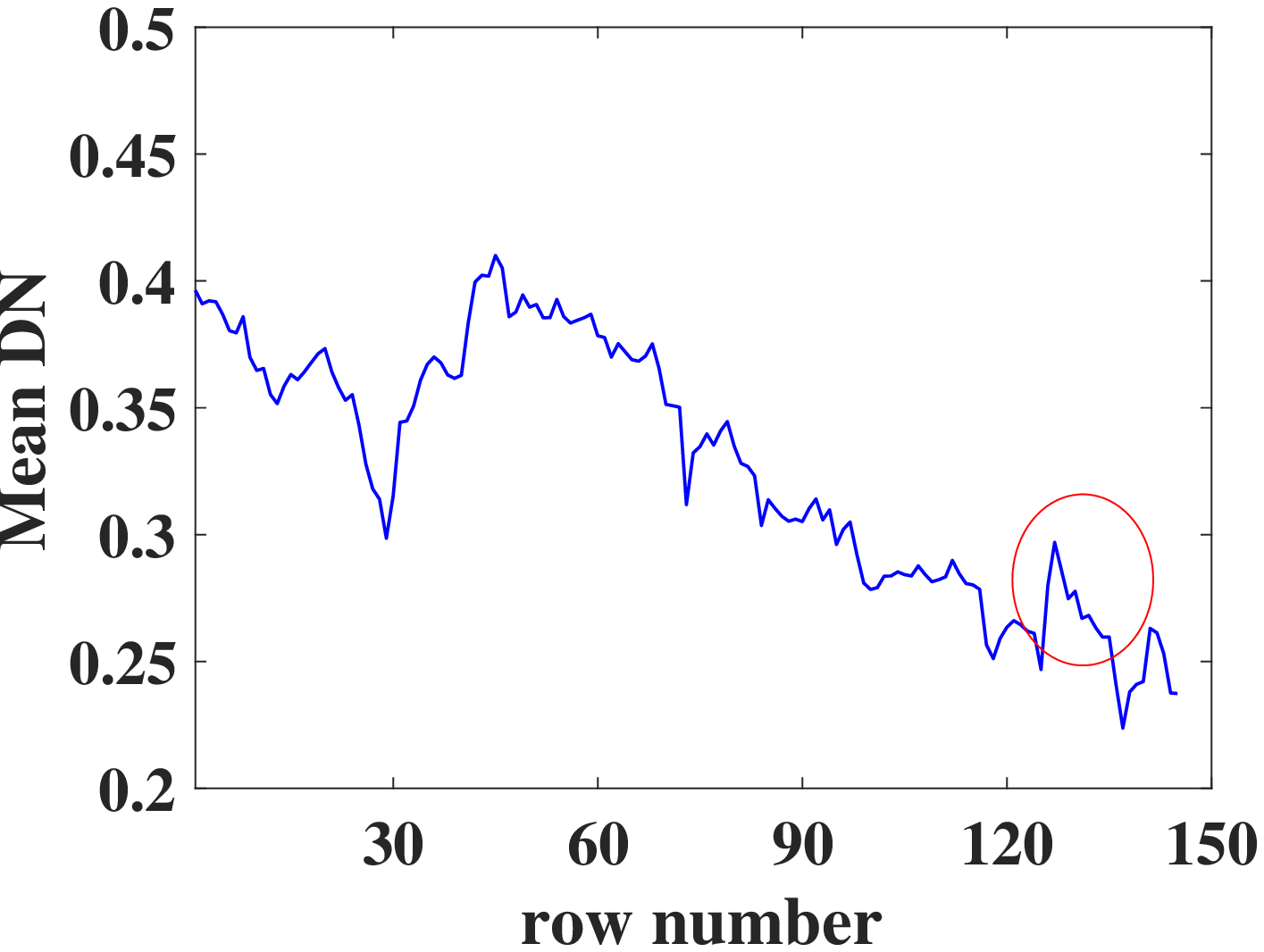}
	}\hspace{0.1mm}
	\subfloat[BM4D] {\includegraphics[width=0.35\columnwidth]{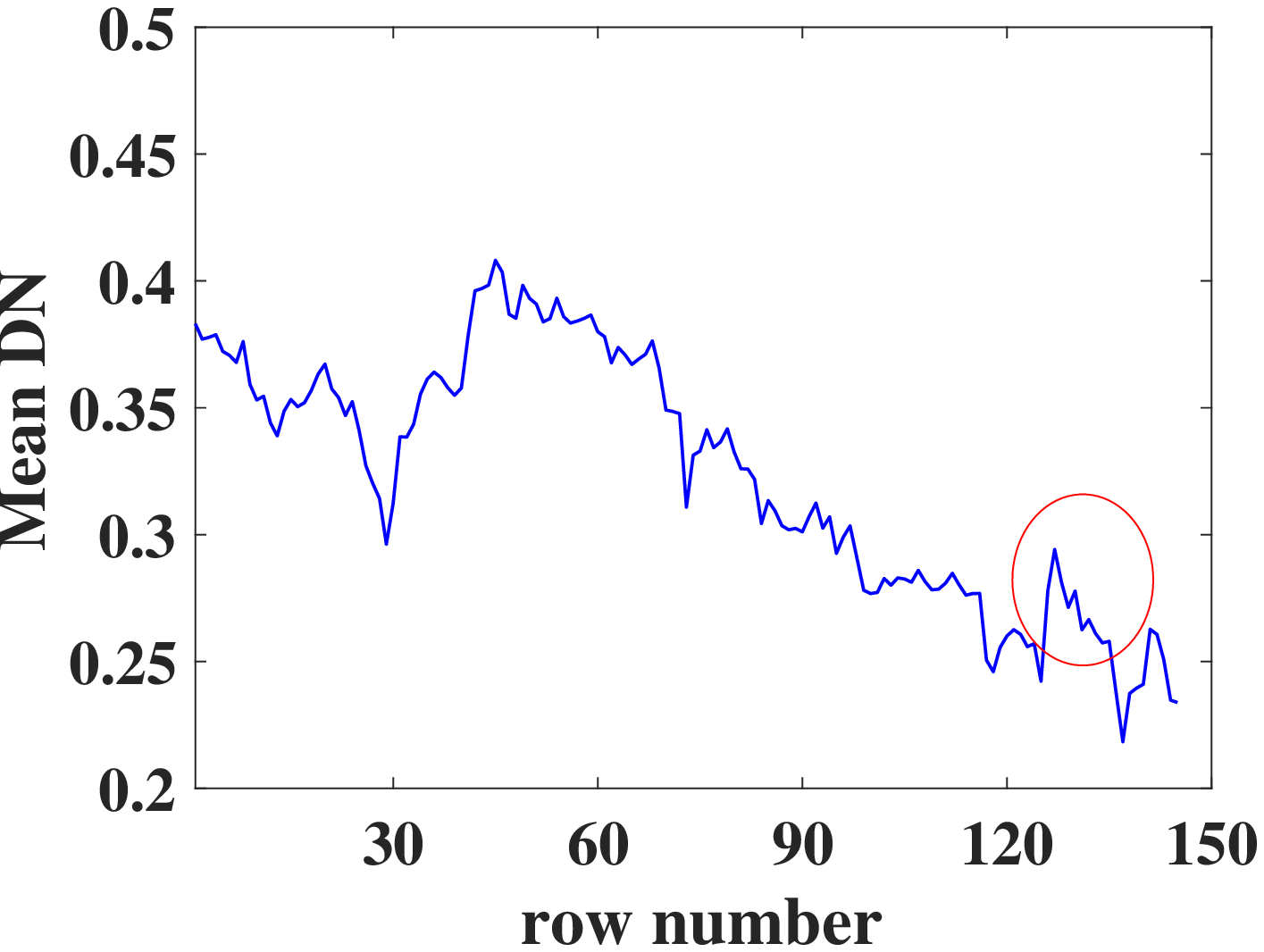}
	}\hspace{0.1mm}
	\subfloat[LRMR] {\includegraphics[width=0.35\columnwidth]{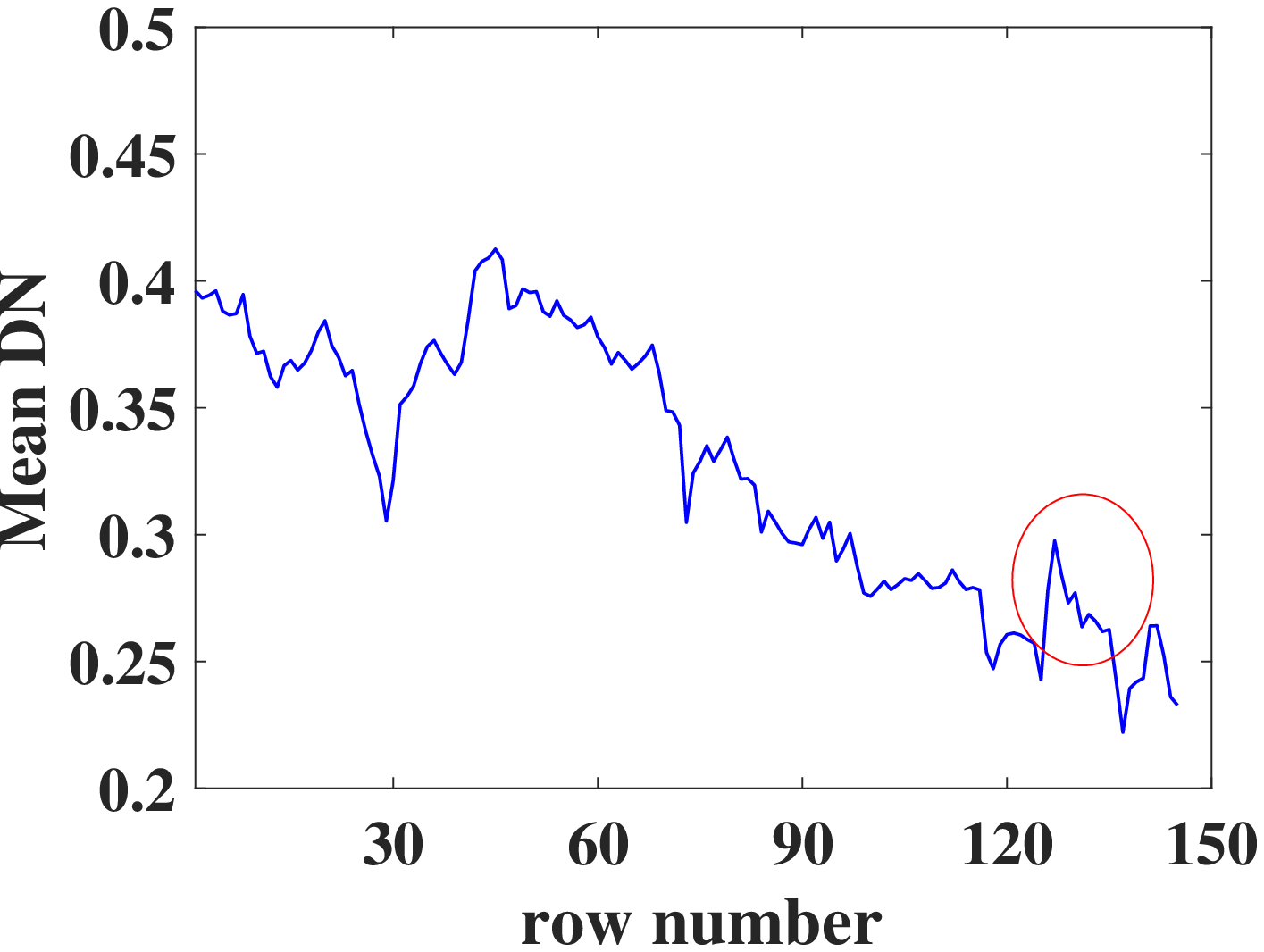}
	}\hspace{0.1mm}
	\subfloat[LRTDTV] {\includegraphics[width=0.35\columnwidth]{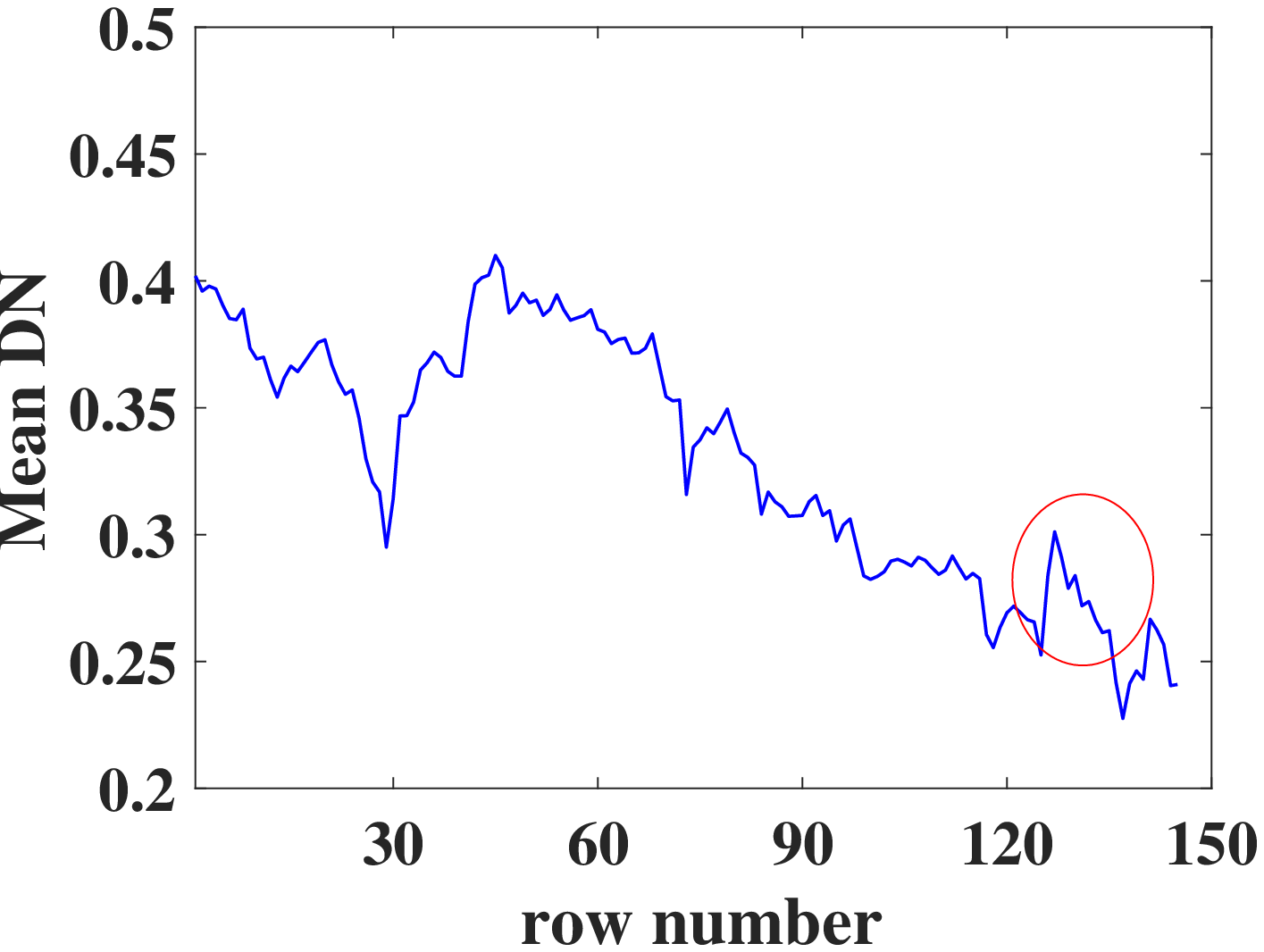}
	}\hspace{0.1mm}
	\subfloat[SSTV] {\includegraphics[width=0.35\columnwidth]{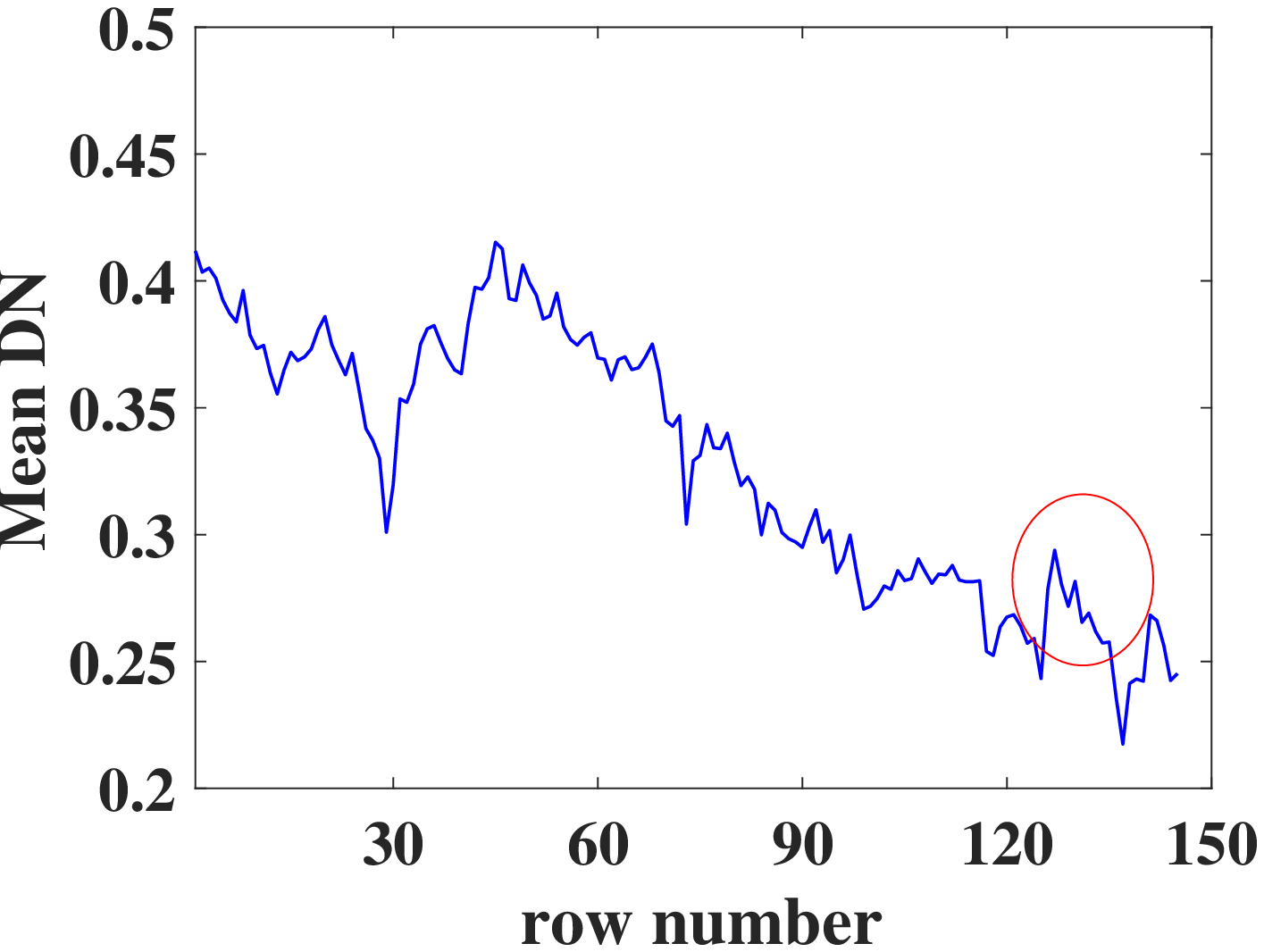}
	}\hspace{0.1mm}
	\subfloat[Our] {\includegraphics[width=0.35\columnwidth]{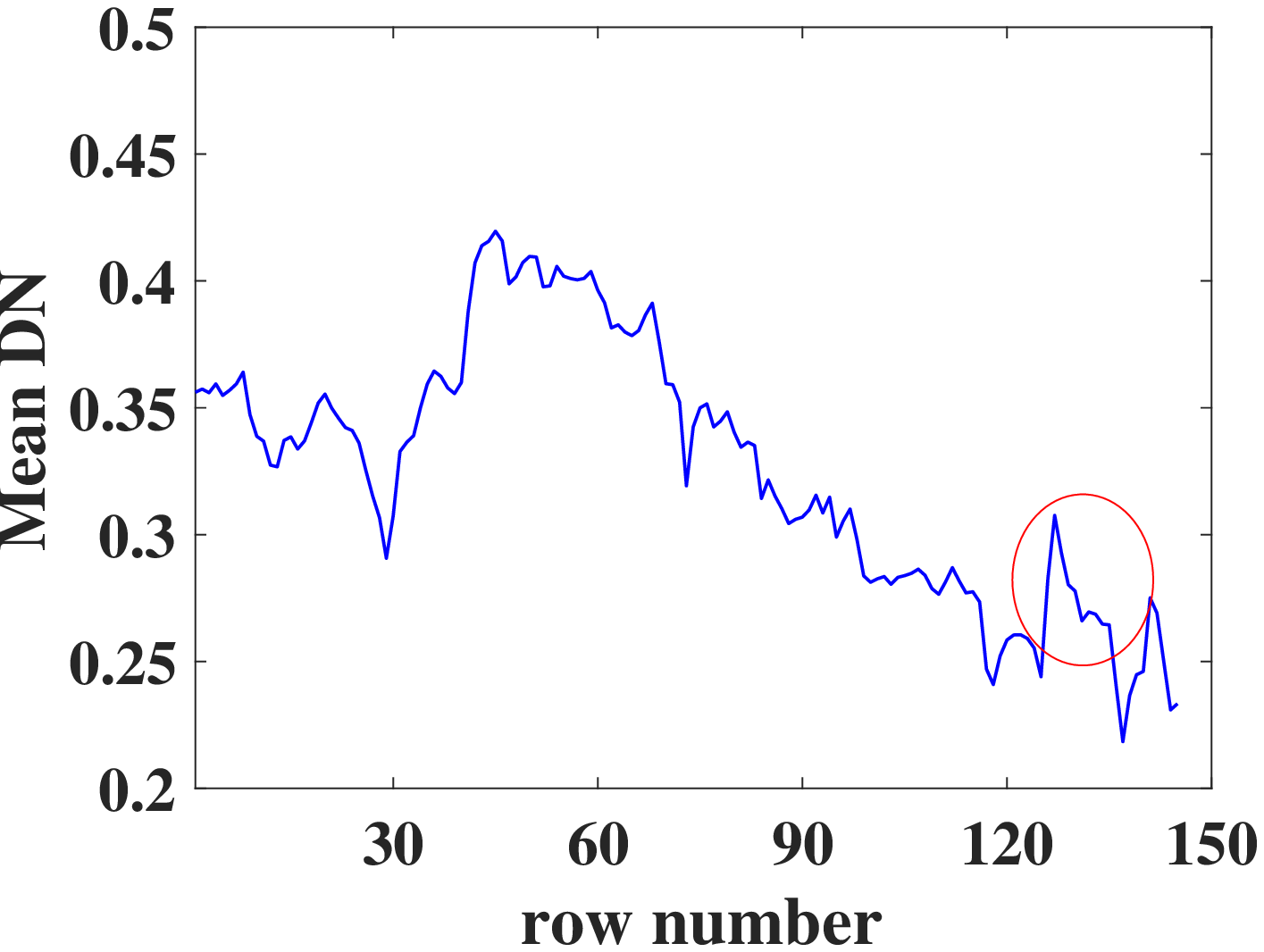}
	}\hspace{0.1mm}
	\caption{Spectral signatures curve comparison on the AVIRIS Indian Pines data. the real noisy image in (a) is band 218. }
	\label{real_imshowIndianDN}
\end{figure*}

	Lastly, we apply the proposed TDLRSTV to real data, i.e., HYDICE urban\footnote{https://engineering.purdue.edu/~biehl/MultiSpec/hyperspectral.html} and the AVIRIS Indian Pines \cite{LRTDTV} to verify the robustness of the algorithms.	
	Fig. \ref{real_imshowUrban} and Fig. \ref{real_imshowIndian} show the recovered two kinds of HSI by using different
	tested algorithms. 
	From the figures, one can see that TDLRSTV can provide better results than LRTDTV and SSTV, and again with more details be preserved.
	The horizontal mean profiles of band 109 of HYDICE Urban and band 218 of AVIRIS Indian Pine before and after restoration is shown in Fig. \ref{real_imshowUrbanDN} and Fig. \ref{real_imshowIndianDN}, where one can see that TDLRSTV provides results with most stable horizontal mean profiles curves.

\subsection{Discussion}
\label{discussion}

\begin{figure*}[htbp] \centering
	\subfloat[] {\includegraphics[width=0.24\columnwidth]{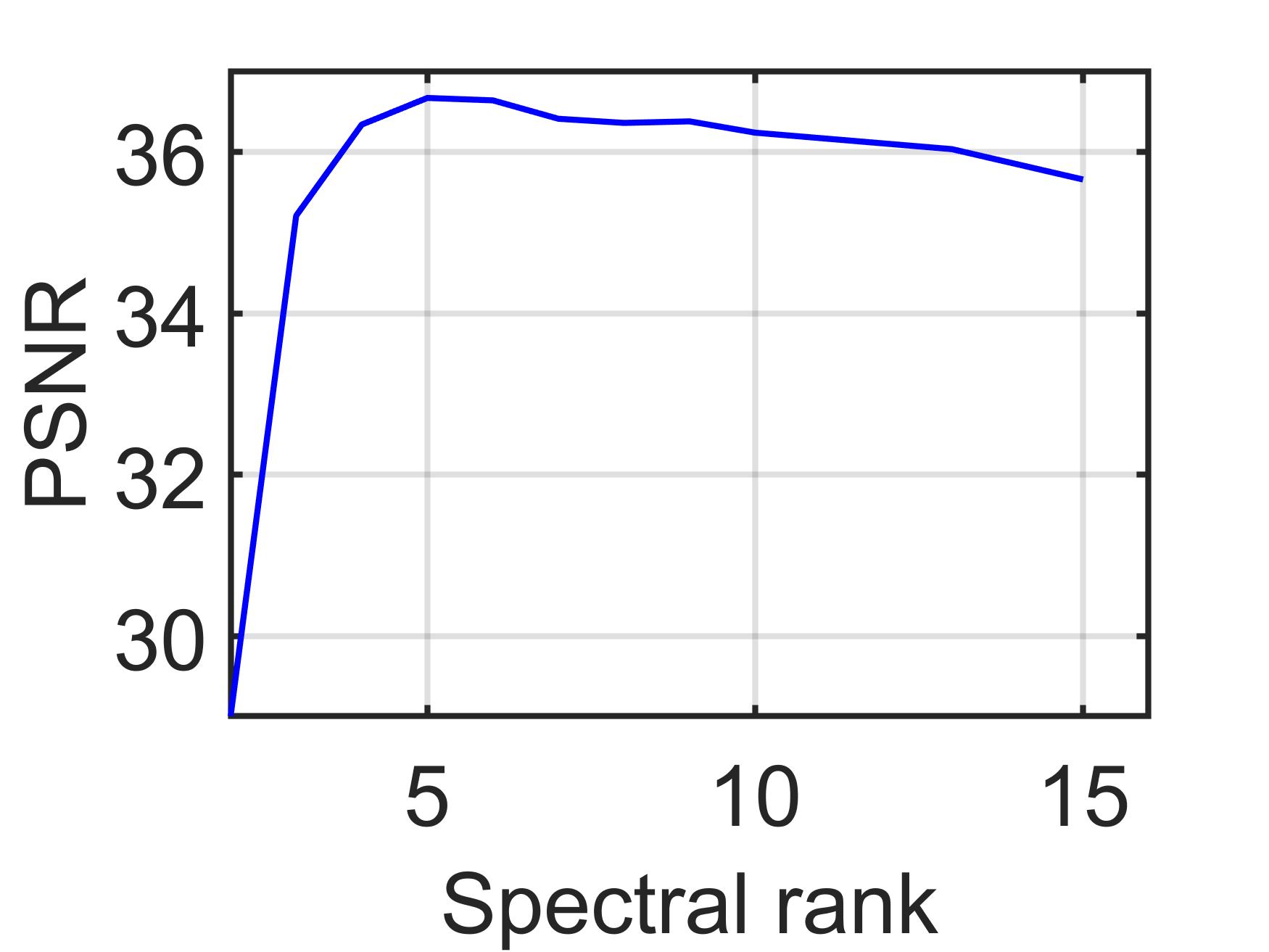}}
	\hspace{0.3mm}
	\subfloat[] {\includegraphics[width=0.24\columnwidth]{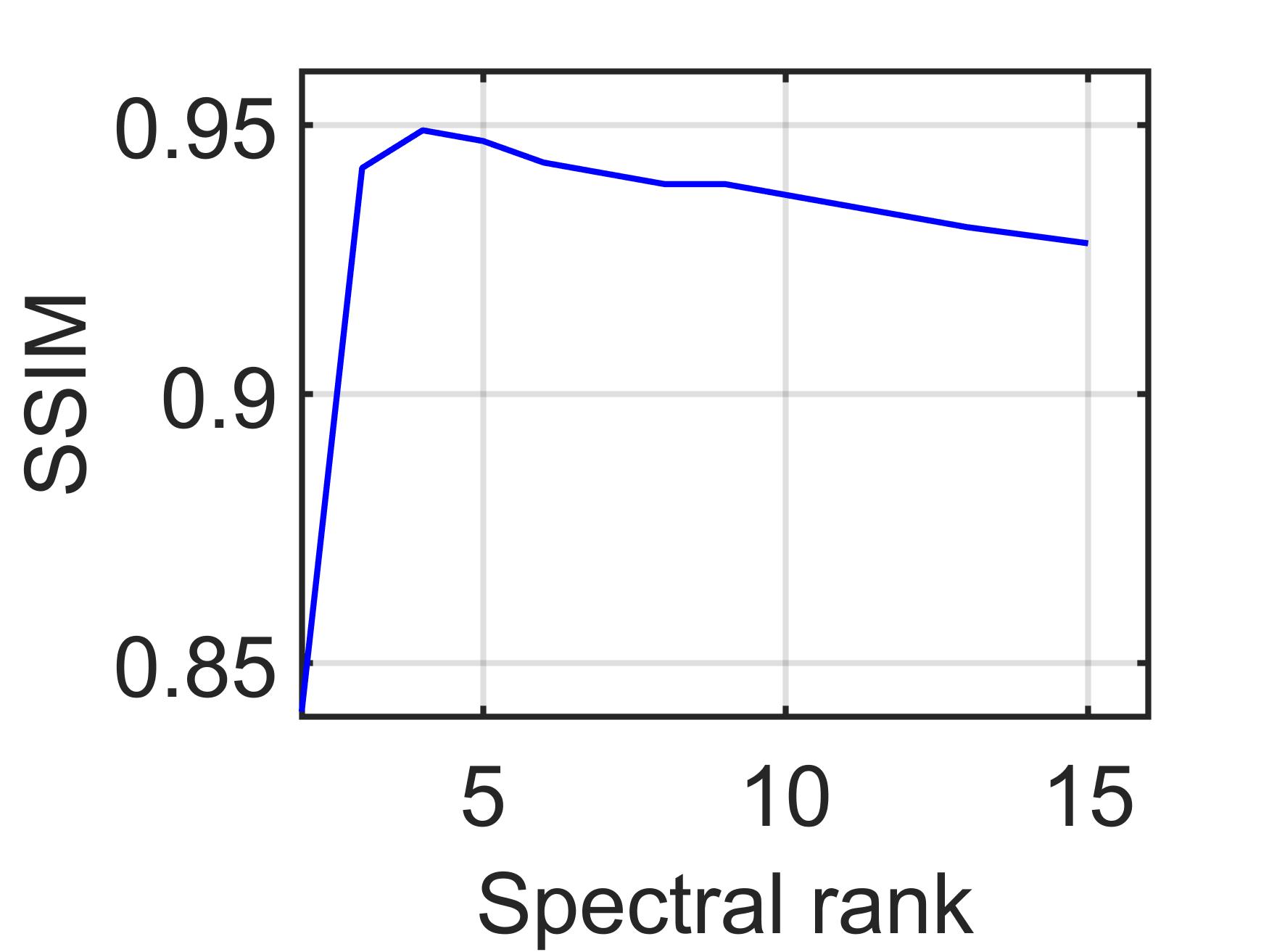}}
	\hspace{0.3mm}
	\subfloat[] {\includegraphics[width=0.24\columnwidth]{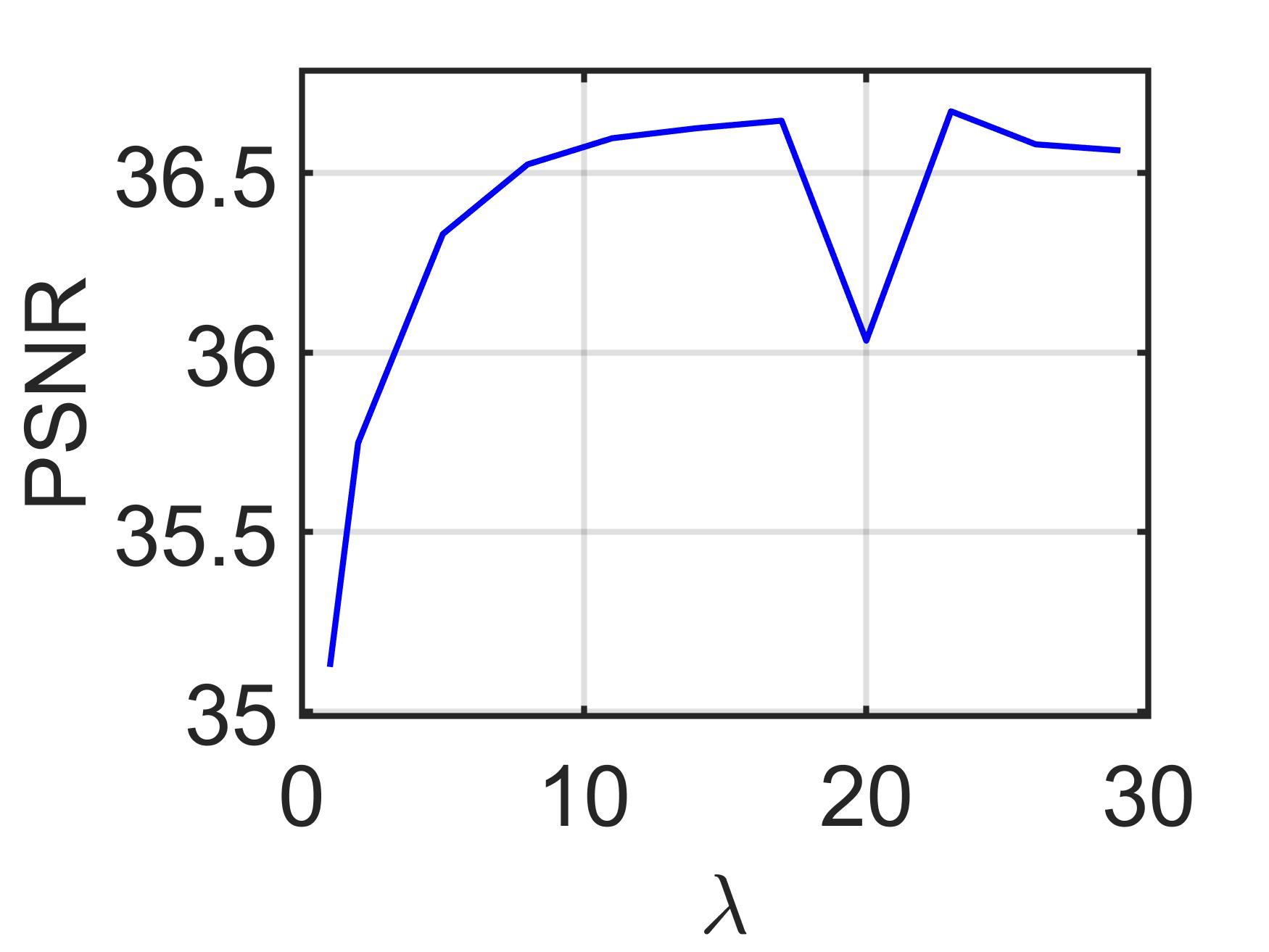}}
	\hspace{0.3mm}
	\subfloat[] {\includegraphics[width=0.24\columnwidth]{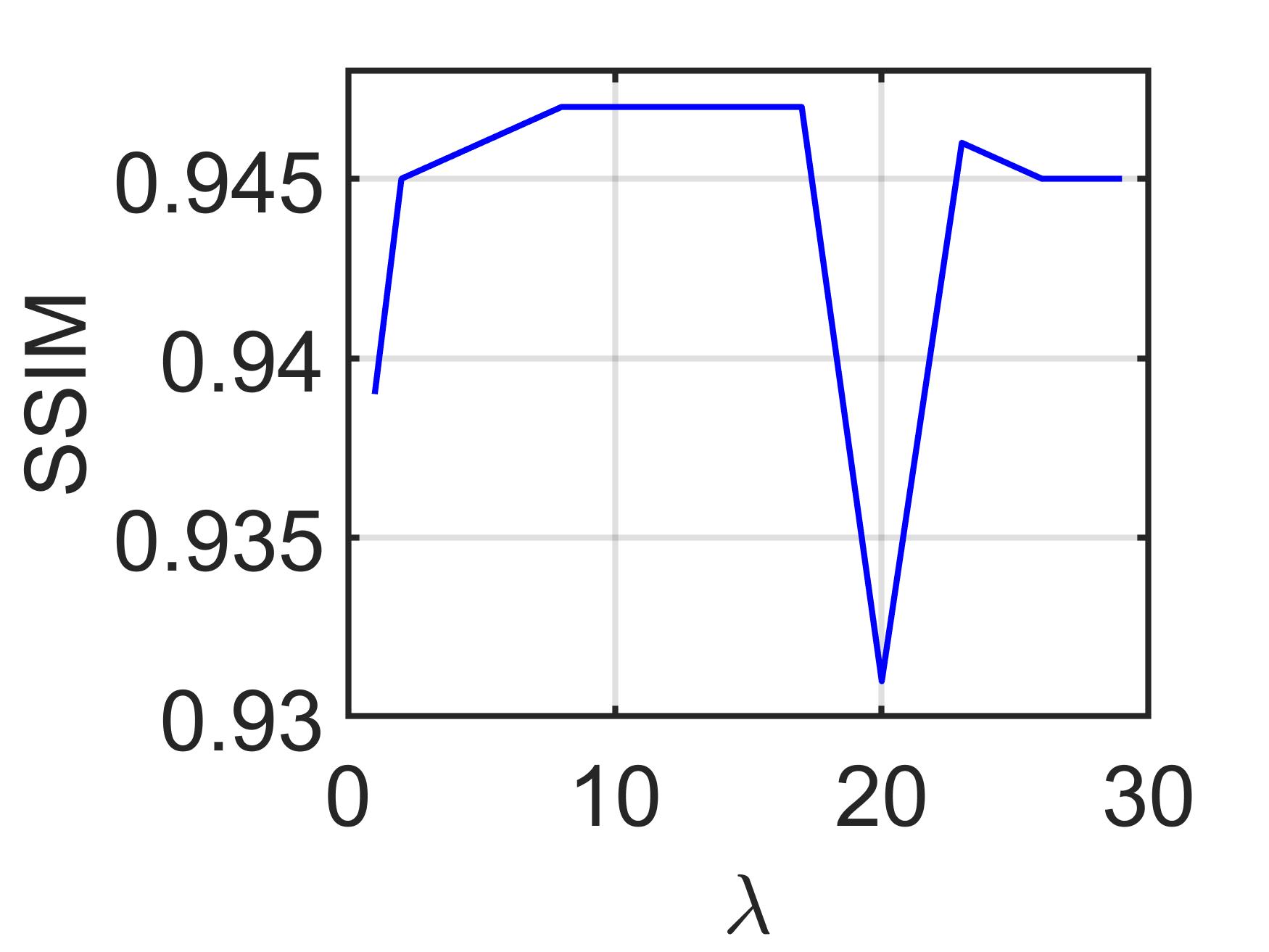}}
	\hspace{0.3mm}
	\subfloat[] {\includegraphics[width=0.24\columnwidth]{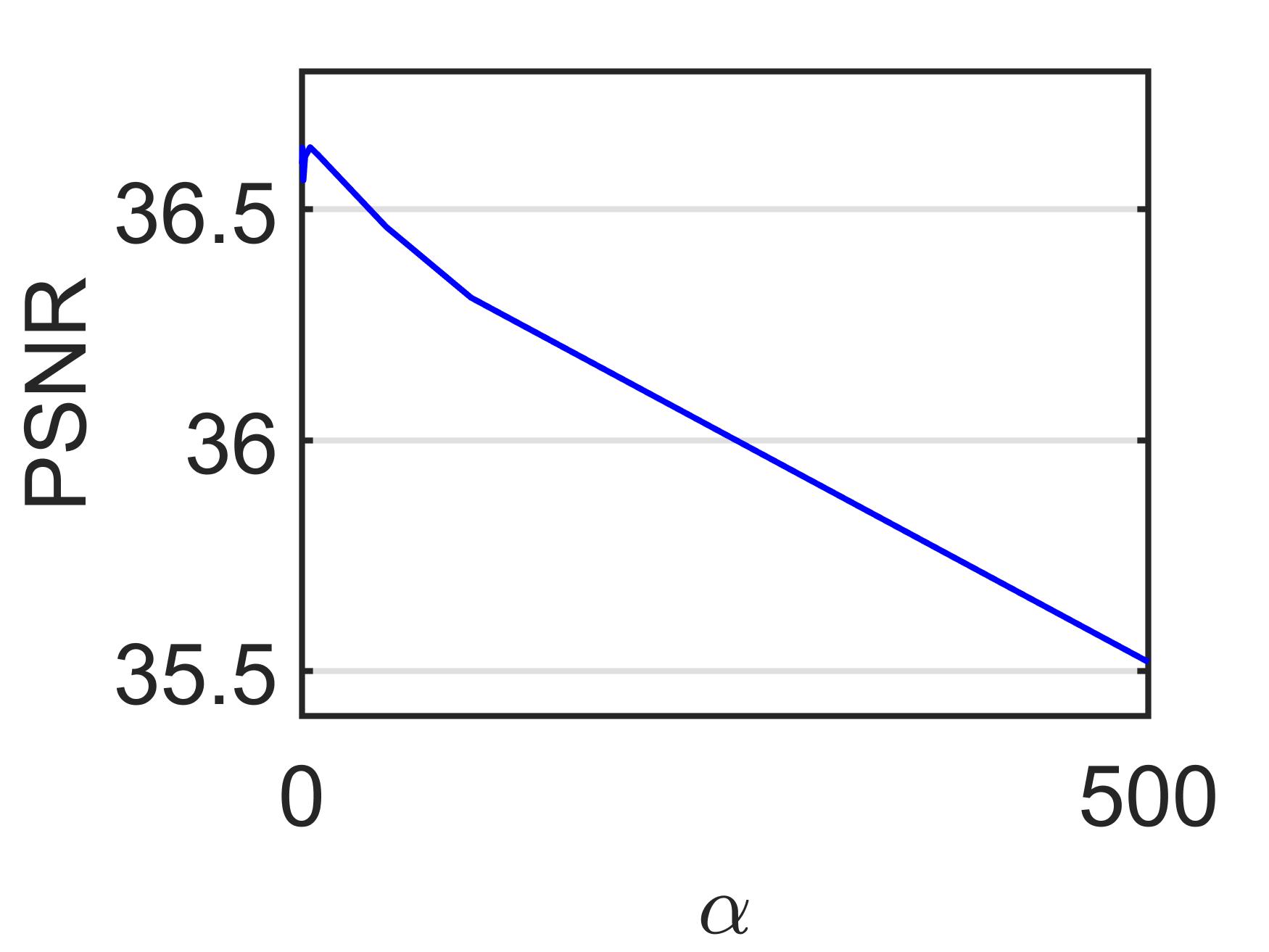}}
	\hspace{0.3mm}
	\subfloat[] {\includegraphics[width=0.24\columnwidth]{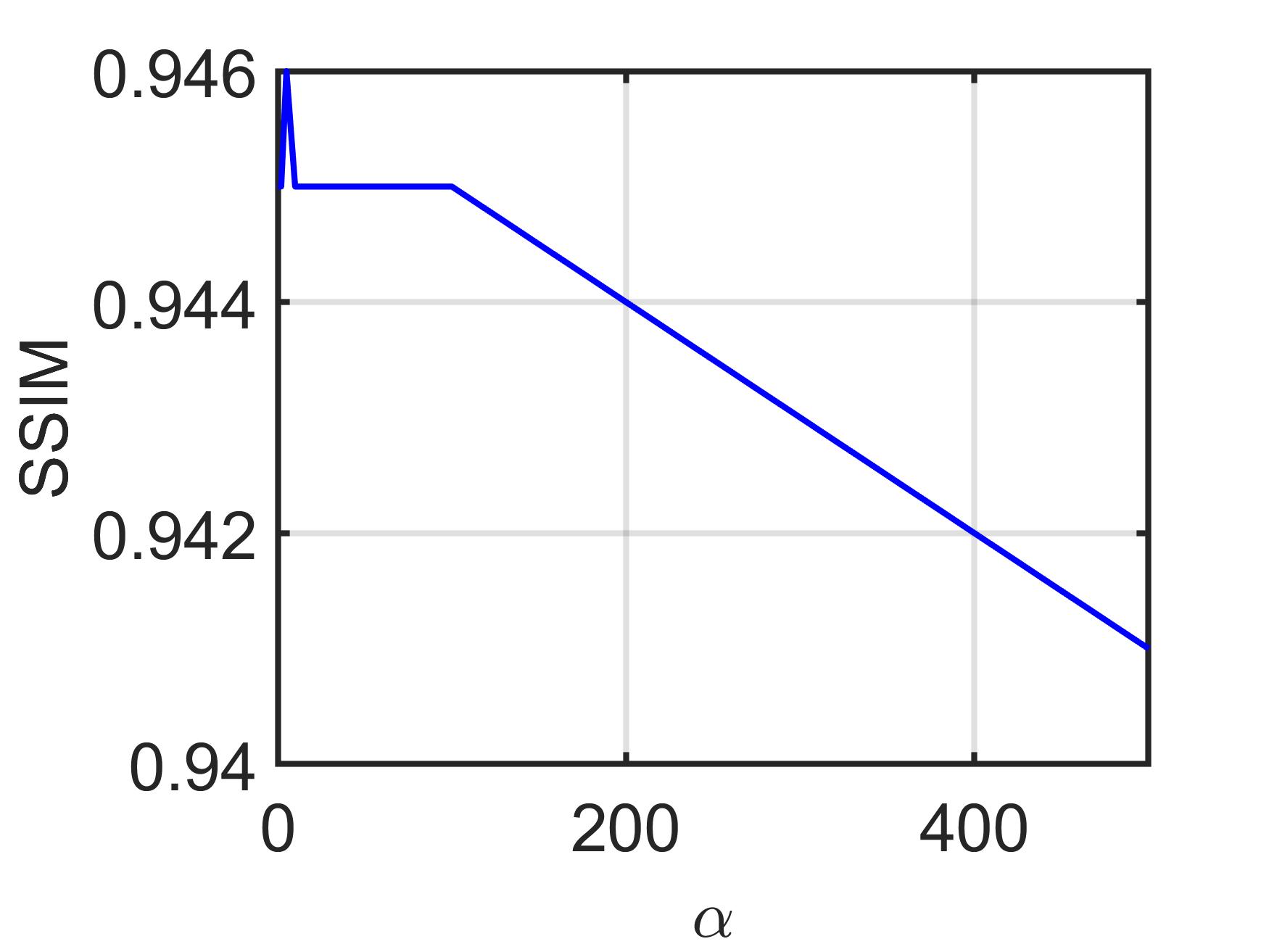}}
	\hspace{0.3mm}
	\subfloat[] {\includegraphics[width=0.24\columnwidth]{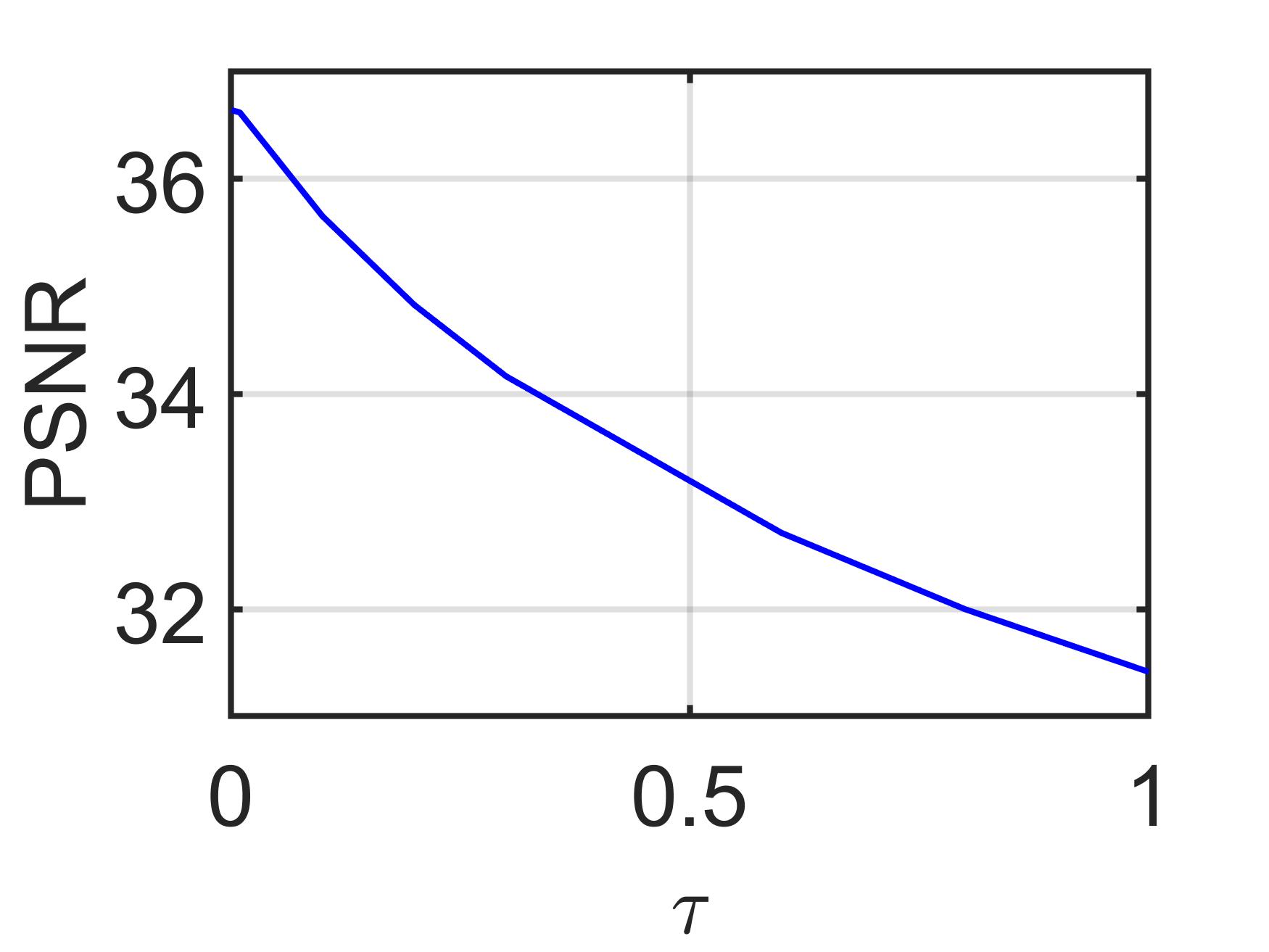}}
	\hspace{0.3mm}
	\subfloat[] {\includegraphics[width=0.24\columnwidth]{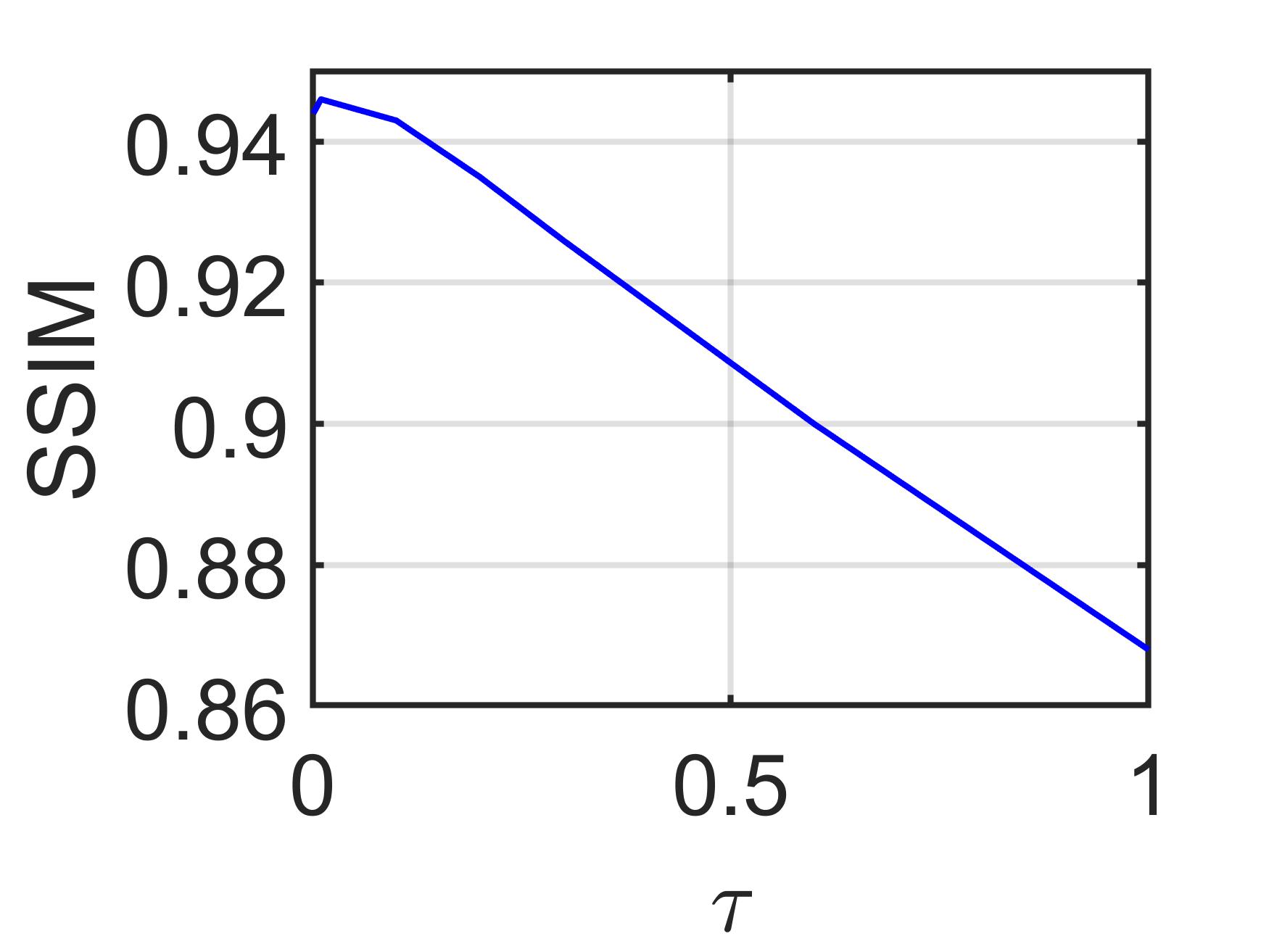}}
	\caption{Sensitivity Analysis of Parameter: spectral rank, $\lambda$ $\alpha $ and $\tau$. (a) PSNR value versus the spectral rank, (a) SSIM value versus the spectral rank, (c) PSNR value versus $\lambda$, (d) SSIM value versus $\lambda$, (e) PSNR value versus $\alpha$, (f) SSIM value versus $\alpha$, (g) PSNR value versus $\tau$, (h) SSIM value versus $\tau$.  }
	\label{para_analysing_TDLRSTV}
\end{figure*}

\emph{Sensitivity Analysis of Parameter:} there are three parameters used to trade-off the regularization terms in the proposed TDLRSTV, i.e., $\tau, \lambda $ and $\alpha$. 
$\lambda $ is used to trade-off the effect of sparsity term $\|\mathcal{S}\|_{1}$, it can be set as $\lambda=C / \sqrt{M N}$ ($M, N$ is the spatial size of underlying HSI) \cite{lu2019tensor}, which is a constant that needs to be tuned manually, Fig. \ref{para_analysing_TDLRSTV} (c-d) show the sensitivity of $\lambda$. 
 $\alpha$ is used to trade-off the low-rank term in LRSTV, $\tau$ is the parameter used to control the effect of sparsity of gradient maps, Fig. \ref{para_analysing_TDLRSTV} (e-f) and (g-h) show the sensitivity of $\alpha$ and $\tau$, respectively. 
Besides, the Tucker rank $(r_1, r_2, r_3)$ involved in the low-rank Tucker decomposition is also a sensitive parameter. We set the spatial Tuker rank as follows: $r_1$ is set to 0.8 times then vertical spatial size and $r_2$ to 0.8 times horizontal spatial size \cite{LRTDTV}, and tune $r_3$ manually. 
Fig. \ref{para_analysing_TDLRSTV} (a-b) show the change of PSNR and SSIM values versus the change of spectral rank $r_3$. 
Here, we set the spectral rank as the one corresponding to the best PSNR or SSIM.

%
%
%
%

\begin{figure}[htbp] \centering
	\subfloat[Error] {\includegraphics[width=0.32\columnwidth]{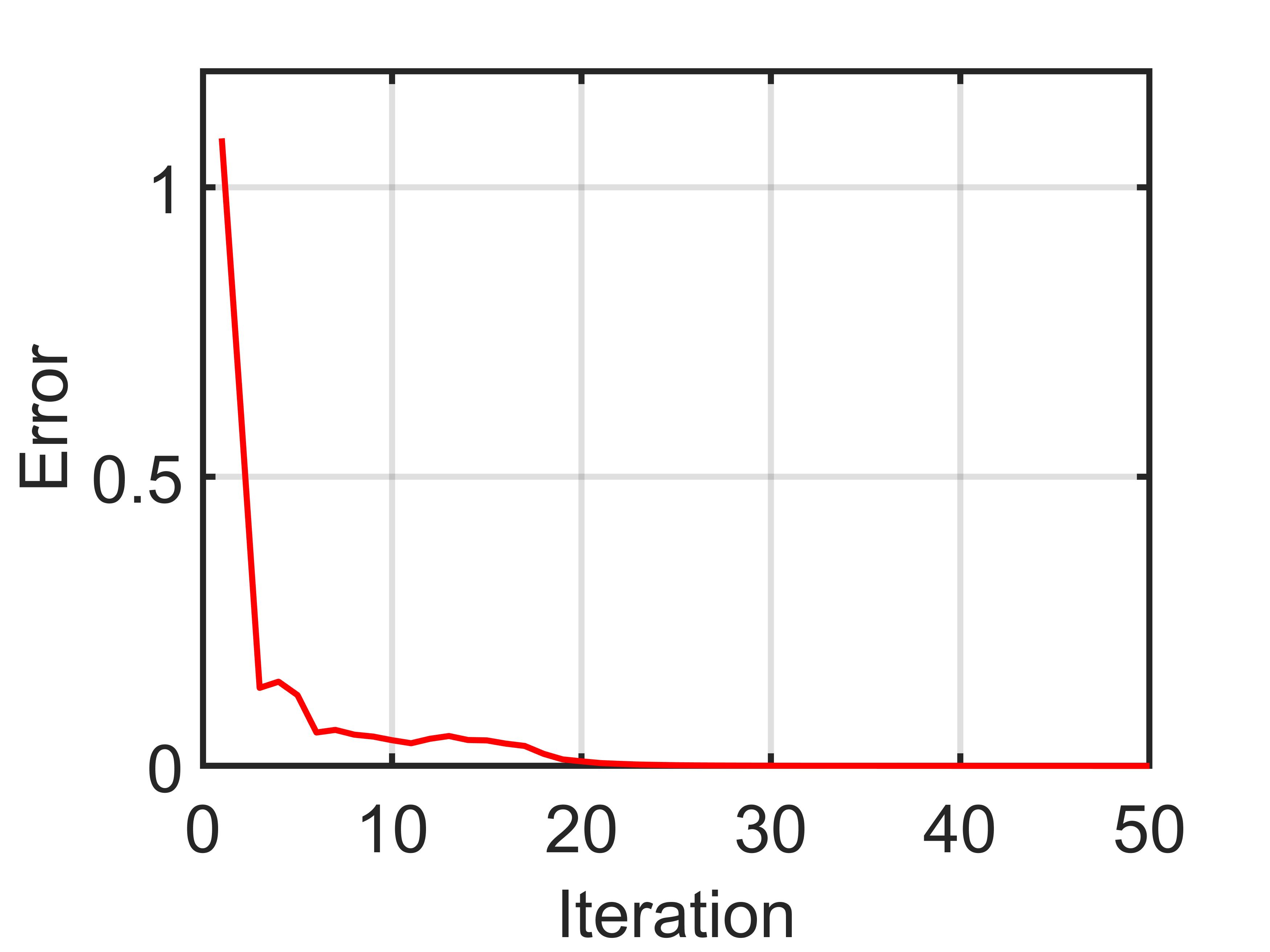}}
	\hspace{0.3mm}
	\subfloat[SSIM] {\includegraphics[width=0.32\columnwidth]{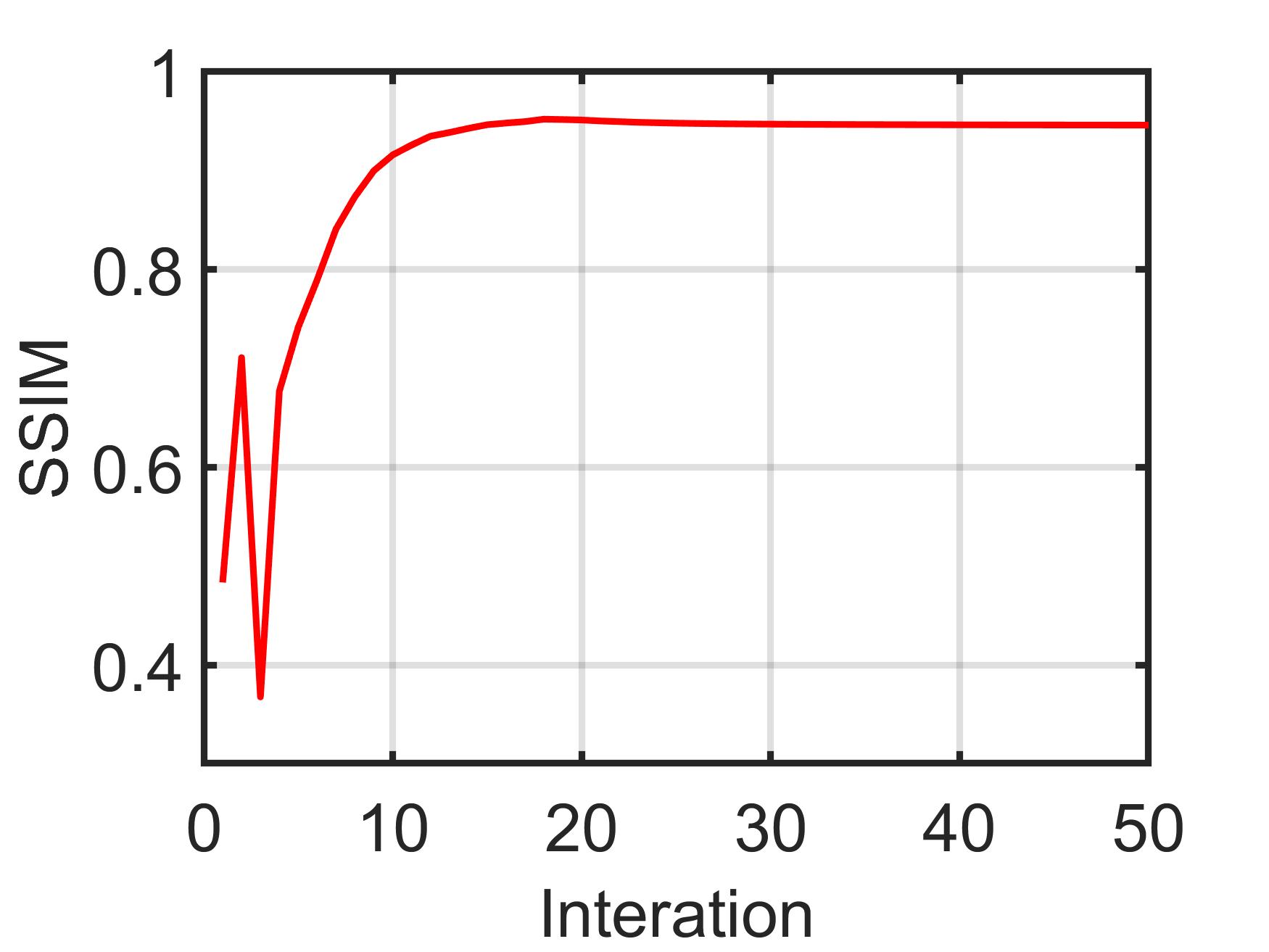}}
	\hspace{0.3mm}
	\subfloat[PSNR] {\includegraphics[width=0.32\columnwidth]{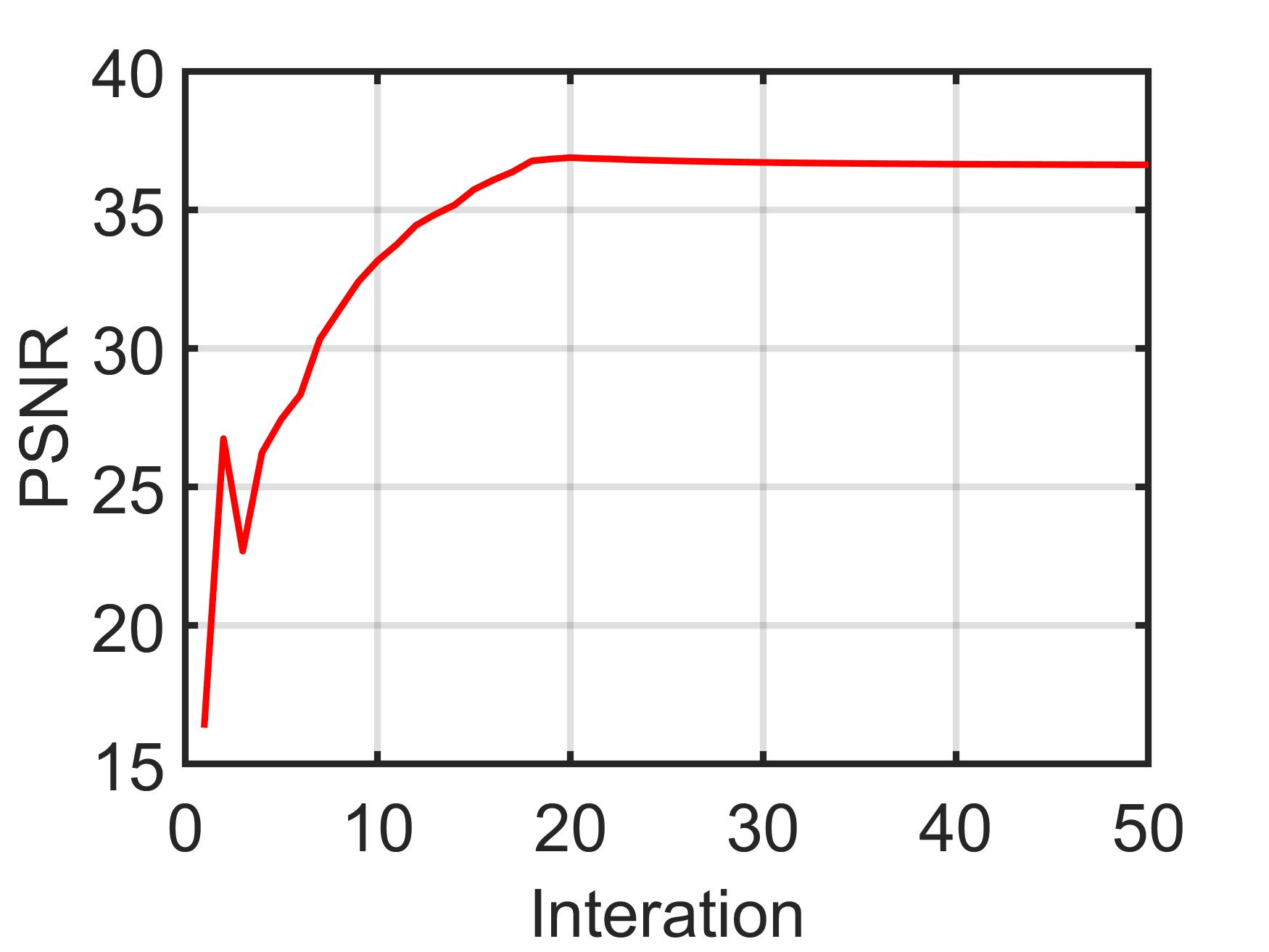}}
	\caption{Relative error, MSSIM and MPSNR value versus the iteration number of TDLRSTV.
		(a) Change in the relative error value, (b) Change in the MSSIM value, (c) Change in the MPSNR value.}
	\label{convergence_TDLRSTV}
\end{figure}

\emph{Convergence of the TDLRSTV based algorithm:} 
To show the convergence of the proposed TDLRSTV, Fig. \ref{convergence_TDLRSTV} shows the change of relative error, MSSIM and MPSNR as iteration increases from 1 to 50. 
In Fig. \ref{convergence_TDLRSTV}, one can see that the relative error, SSIM and PSNR all converge to fixed values, which means that the proposed algorithm has well robustness to be further used in real scene.

\section{conclusion} \label{conclusion}
In this paper, we propose a novel TV regularization technique for HS image restoration, called LRSTV. 
The proposed LRSTV uses the tensor $L_1$-norm and the convex envelope of the average rank to explicitly represent the low-rank and sparsity of the gradient map in the transform domain. 
We overcome the obstacles of traditional SSTV, which is not able
to retain the correlation of the spatial spectral structure of
the gradient map and which cannot cope with tasks with high noise
intensity.
On the one hand, we revealed the fact that the gradient map of HSI not only has sparseness, but also has a strong low rank in the Fourier domain. Finally, we simply embed the proposed LRSTV into the model designed for HSI denoising, and excellent results on both simulation and real datasets.
One deeper work in the future is to design a better low-rank regularization term or train a denoising network and combine it with the proposed LRSTV to improve the performance of HSI restoration model.

\ifCLASSOPTIONcaptionsoff
  \newpage
\fi

\small
\bibliographystyle{IEEEtran}
\bibliography{IEEEabrv,mybibfile}

\end{document}